\documentclass[11pt]{article}
\usepackage[final]{acl}
\usepackage{times}
\usepackage{latexsym}
\usepackage[T1]{fontenc}
\usepackage[utf8]{inputenc}
\usepackage{microtype}
\usepackage{inconsolata}
\usepackage{cuted}
\usepackage{balance}
\usepackage{caption}
\usepackage{amsmath, amssymb, amsfonts}
\usepackage{booktabs}
\usepackage{array}
\usepackage{colortbl}
\usepackage{enumitem}
\usepackage{xspace}
\usepackage{graphicx}
\graphicspath{{./}}
\usepackage[ruled, linesnumbered, vlined]{algorithm2e}
\usepackage{tabularx}
\newcolumntype{R}{>{\raggedleft\arraybackslash}X}
\usepackage{placeins}
\usepackage[most]{tcolorbox}
\usepackage{pgfplots}
\pgfplotsset{compat=1.18}
\usepgfplotslibrary{groupplots}
\usetikzlibrary{patterns, calc}
\usepackage{float}
\definecolor{sepoTint}{HTML}{EAF4F1}
\definecolor{tableHead}{HTML}{F3F6F8}
\definecolor{tableMean}{HTML}{FFF3D9}
\definecolor{tableRule}{HTML}{5B7F79}
\definecolor{llamaBlock}{HTML}{DCE8F6}
\definecolor{llamaBest}{HTML}{D7E6FA}
\definecolor{qwenBlock}{HTML}{DFF3E2}
\definecolor{qwenBest}{HTML}{D9F0DD}
\definecolor{diffadd}{HTML}{E6FFED}
\definecolor{diffdel}{HTML}{FFECEC}
\definecolor{warnfg}{HTML}{B00020}
\definecolor{tableRule}{HTML}{4A5568}
\definecolor{evgreen}{HTML}{22863A}
\definecolor{tableMean}{HTML}{F1F8FF}
\definecolor{promptBg}{HTML}{EEF3F8}     
\definecolor{promptFrame}{HTML}{2C5282}  
\definecolor{editblue}{HTML}{6058E0} 
\definecolor{hlrefine}{HTML}{C2D5F2}     
\providecommand{\sepotabscore}[2]{#1$_{\pm #2}$}
\providecommand{\sepotabbest}[2]{\textbf{#1}$_{\pm #2}$}
\providecommand{\sepotabllamabest}[2]{\cellcolor{llamaBest}\sepotabbest{#1}{#2}}
\providecommand{\sepotabqwenbest}[2]{\cellcolor{qwenBest}\sepotabbest{#1}{#2}}

\newcommand{\sepo}{\textsc{Sepo}\xspace}
\DeclareMathOperator*{\argmax}{arg\,max}

\newtcolorbox{sepoexample}[1]{
  enhanced, breakable,
  colback=promptBg,
  colframe=promptFrame,
  coltitle=white,
  colbacktitle=promptFrame,
  fonttitle=\bfseries\small,
  title={#1},
  arc=2pt,
  boxrule=0.5pt,
  left=8pt,right=8pt,top=4pt,bottom=4pt,
  before skip=8pt, after skip=8pt
}

\title{SEPO: Evidence-Grounded Prompt Optimization via\\
       Structural Editing}

\author{
\textbf{
Xiaoyu Ma\textsuperscript{$\spadesuit\dagger$}\quad
Haoyue Liu\textsuperscript{$\spadesuit\dagger$}\quad
Yiwen Li\textsuperscript{$\spadesuit$}\quad
Jionghao Zhu\textsuperscript{$\spadesuit$}
}\\[3pt]
\textbf{
Zhichao Wang\textsuperscript{$\spadesuit$}\quad
Ye Chen\textsuperscript{$\heartsuit$}\quad
Xiaoying Tang\textsuperscript{$\spadesuit\clubsuit\diamondsuit*$}
}\\[5pt]
{\textsuperscript{$\spadesuit$}School of Science and Engineering, The Chinese University of Hong Kong, Shenzhen}\\[-1pt]
{\textsuperscript{$\clubsuit$}Shenzhen Future Network of Intelligence Institute (FNiI-Shenzhen)}\\[-1pt]
{\textsuperscript{$\diamondsuit$}Guangdong Provincial Key Laboratory of Future Networks of Intelligence, CUHK(SZ)}\\[-1pt]
{\textsuperscript{$\heartsuit$}Xi'an Jiaotong University}\\[2pt]
{\textbf{Email:} \texttt{xiaoyuma@link.cuhk.edu.cn, tangxiaoying@cuhk.edu.cn}}\\[2pt]
{\small\textsuperscript{$\dagger$}Equal contribution.\quad
\textsuperscript{*}Corresponding author.}
}

\begin{document}
\maketitle

\begin{abstract}
Existing API-only prompt optimisers are often described as
\textbf{interpretable}, but in practice, this usually means only post-hoc
inspectability: each iteration still rewrites the prompt as one opaque
string, leaving a trace of full-prompt diffs rather than localisable,
machine-readable edits.  This paper introduces \sepo
(Structural, Evidence-grounded Prompt Optimization), a multi-trajectory
prompt optimiser centred on \emph{edit-effect lineage feedback}. Rather
than treating each iteration as an isolated whole-prompt rewrite, \sepo
locally edits stable, typed units in a two-layer prompt schema, links the
target and realised structural operations of each edit to the examples it
newly fixes or breaks, and carries this edit--effect record forward to
guide later architect calls on the same search branch. This makes prompt
optimisation \emph{addressable}, \emph{attributable}, and
\emph{actionable}. Across a 14-task held-out suite, \sepo improves over
the strongest baseline, \textsc{GEPA}, by 3.1\,pp on
Llama-3.1-8B-Instruct and 2.2\,pp on Qwen3-8B, reaching
61.9\% and 73.3\% macro accuracy.  \sepo also lies on both the
optimisation-time and test-time Pareto frontiers, spending 2.9M
optimisation tokens versus 4.1M for \textsc{GEPA} and producing prompts
over 5$\times$ shorter.
\end{abstract}

\begin{figure}[t]
\centering
\includegraphics[width=\columnwidth]{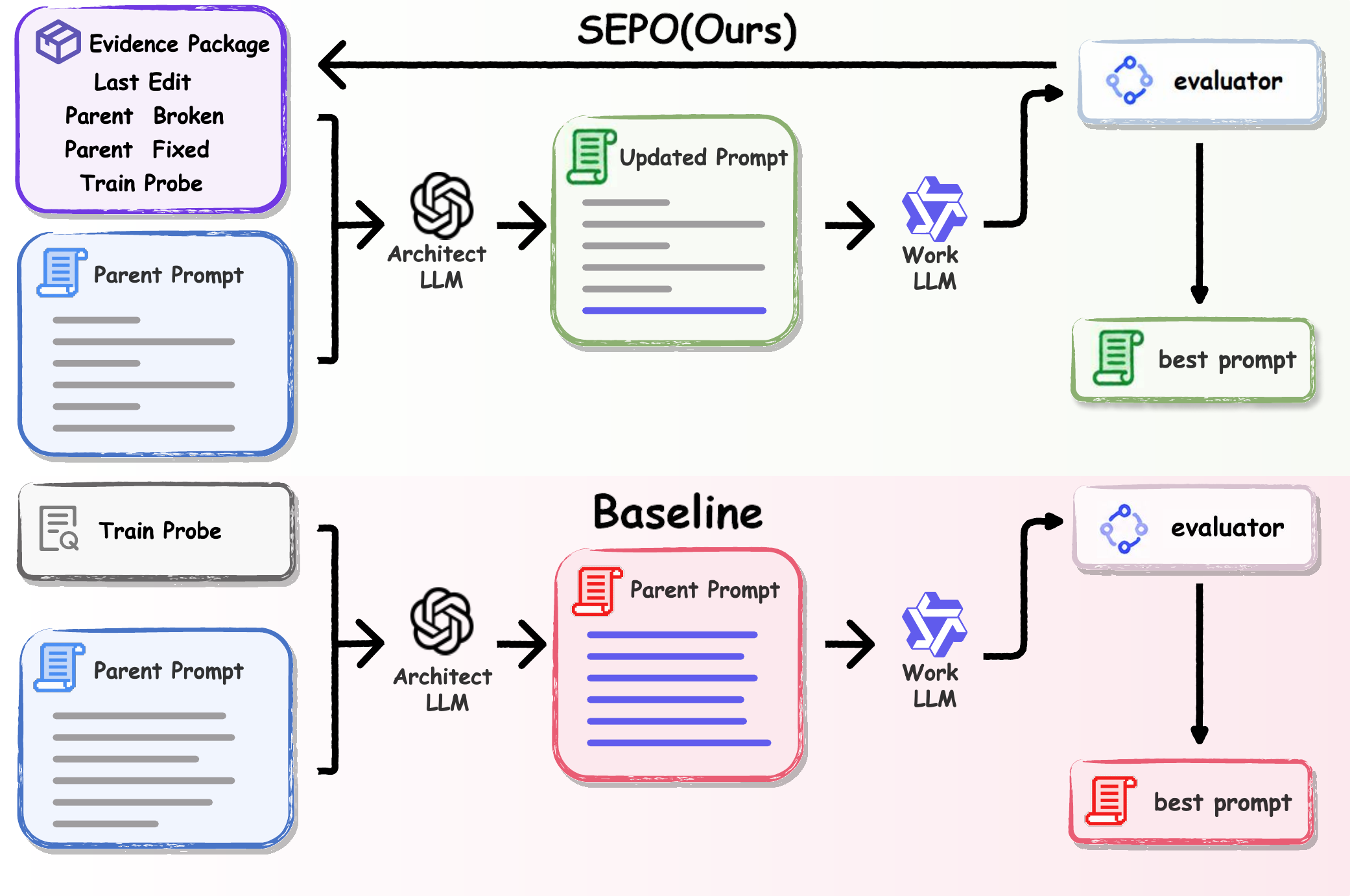}
\caption{\textbf{\sepo at a glance.} \sepo makes localised,
evidence-grounded edits (\textcolor{editblue}{blue}) to a structured prompt (top),
each traceable to examples it fixed or broke, whereas monolithic rewriters rewrite the
whole prompt (\textcolor{editblue}{blue}), opaquely (bottom).}
\label{fig:teaser}
\end{figure}
\section{Introduction}

Prompt optimisation has become a de-facto recipe for adapting frozen LLMs
to new tasks without weight updates: an architect LLM proposes a rewrite,
the candidate is scored on a training set, and the cycle
repeats~\citep{ramnath2025systematic, yang2024opro, pryzant2023protegi,
agrawal2025gepa, yuksekgonul2024textgrad}.  Despite being commonly described as
\emph{interpretable}, most API-only prompt optimisers still expose the
same underlying interface: each iteration rewrites the prompt as one
opaque string.  The resulting trace is therefore a sequence of
full-prompt diffs: it shows \emph{that} the prompt changed, but not
\emph{where} the relevant change occurred, \emph{why} it was made, or
\emph{which} examples it helped or harmed.  This limitation persists
whether the rewriter is driven by text-gradient feedback or by
LLM-as-optimiser sampling: the unit of edit remains the entire prompt.

A handful of recent structured optimisers begin to challenge this
single-string interface.  Some expose local edit targets, such as
factor-, section-, token-, or tree-level units
\citep{schnabel2024sammo,liu2026apsf,sharma2026modular,jain2025local,kumar2025sculpt};
others attribute failures to semantic units or agent contributions
\citep{hapo2026,xia2026hivemind,zhang2025agentracer}; and reflective or self-evolving
optimisers improve search with archives, critique, and synthesis
\citep{agrawal2025gepa,agarwal2025promptwizard}.  Existing prompt
optimisers typically update a prompt based on failures surfaced in the
current round.  However, rules added to address those failures can also
change the worker's behaviour on examples that it handled correctly under
the parent, thereby introducing new regressions.  Without this
parent-to-child transition, later iterations may conflate persistent
failures with regressions introduced by the preceding edit, leading to
repeated rule accretion or another broad prompt rewrite.  This can cause
prompt bloat and make effective improvements harder to preserve across
iterations.

In this paper, we propose \sepo, a multi-trajectory prompt optimiser that
addresses this problem through \emph{edit-effect lineage feedback}. Each
edit targets a stable typed unit in a two-layer prompt schema; after the
candidate is evaluated, \sepo links the target and realised operations of
the edit to the examples it newly fixes or breaks and reuses this record in
later architect calls on the same search branch. To preserve trajectory
diversity, \sepo combines example-level admission with
Lexicase-based parent selection~\citep{spector2010assessment}, keeping
specialists alive for rare training examples.

We evaluate \sepo on a 14-task held-out suite using
Llama-3.1-8B-Instruct and Qwen3-8B as worker models.  We cap
optimisation at $2000$ \emph{metric calls} per task and seed---one
metric call evaluates $\mu$ on one training example, consuming one
Worker LLM forward pass.  Under this matched budget, \sepo reaches
61.9/73.3\% macro accuracy on Llama/Qwen---$+3.1/+2.2$\,pp over the
best of six API-only baselines (\textsc{GEPA})---while landing on
both optimisation- and test-time Pareto frontiers.

\medskip
\noindent\textbf{Our contributions are four-fold.}

\textbf{(i)} We introduce \sepo and its core mechanism, \emph{edit-effect
lineage feedback}. This mechanism links the target and realised operations
of each local edit to the examples it newly fixes or breaks and reuses this
record in subsequent edits on the same search branch, making prompt
optimisation \emph{addressable}, \emph{attributable}, and
\emph{actionable}.

\textbf{(ii)} We provide a controlled API-only comparison on a 14-task
held-out suite covering reasoning (BBH, GSM-Hard), coding (MBPP+),
multi-hop question answering (HotpotQA), and multidomain tasks from
MMLU-Pro.  Under the matched $2000$-metric-call budget, \sepo
consistently matches or exceeds six API-only baselines across both
worker models, winning at least 11/14 tasks against every baseline.

\textbf{(iii)} We show that the gains from structured, lineage-aware
editing are strongest on procedural, multi-step tasks---most clearly
Word Sorting ($+18.9$\,pp over \textsc{GEPA} on Llama) and MBPP+
($+8.7$\,pp on Qwen).

\textbf{(iv)} We show that \sepo sits on both the optimisation-time and
test-time Pareto frontiers, spending 2.9M optimisation tokens and 203k
deployment tokens per cell---compared with 4.1M and 424k for
\textsc{GEPA}---with prompts over 5$\times$ shorter than
\textsc{GEPA}'s and remaining 6.9\,pp above \textsc{APSF}, the only
other method on both frontiers.


\section{Related Work}
\label{sec:related}

\paragraph{Monolithic whole-prompt rewriting.}
A first family rewrites the prompt as an opaque string under
a scalar validation reward~\citep{ye2024pe2}: APE~\citep{zhou2023ape}
samples and re-scores instructions, OPRO~\citep{yang2024opro} casts the
LM as a numerical-feedback optimiser, GrIPS~\citep{prasad2023grips}
searches with gradient-free operators, and
ProTeGi~\citep{pryzant2023protegi} introduces ``textual gradient'' beam
search, extended by TextGrad~\citep{yuksekgonul2024textgrad} to
text-level backpropagation. Lacking component-level addresses, a
rewrite that breaks one example yields no localisable change to
attribute, and its trace collapses to untyped full-prompt diffs.

\begin{figure*}[!t]
\centering
\includegraphics[width=0.95\textwidth]{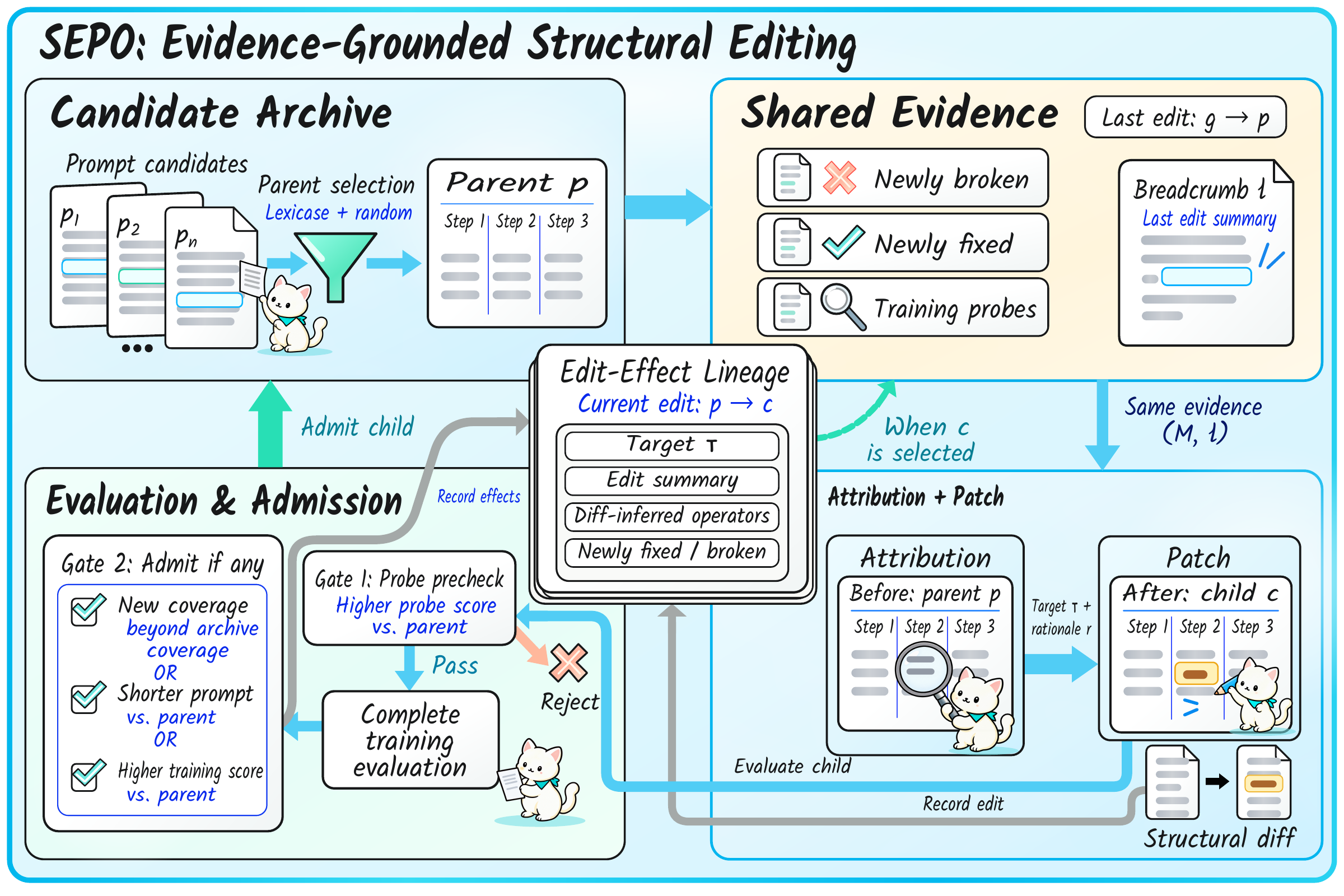}
\caption{\textbf{Per-iteration overview of \sepo.}
A parent is selected from the archive $\mathcal{A}$ using Lexicase or
random sampling (\S\ref{sec:loop}). The attribution and patch architects
share evidence from its preceding edit: newly broken and newly fixed
examples, training probes, and a breadcrumb summarising the edit
(\S\ref{sec:editing}). The resulting child must outperform its parent
on the probes (Gate~1) before completing training evaluation.
It enters the archive if it adds coverage beyond the archive,
is shorter than its parent, or achieves a higher training score
than its parent (Gate~2; \S\ref{sec:loop}).
The child's edit target, diff-inferred operators, and example-level
effects are recorded to guide subsequent edits when it is selected
as a parent.
}
\end{figure*}

\label{fig:overview}

\paragraph{Structure-aware factorised optimisers.}
A second family exposes components as first-class units:
\emph{scaffold-based} methods optimise per-module instructions and
demonstrations over a user-specified program
(DSPy+MIPROv2~\citep{khattab2024dspy, opsahlong2024miprov2}); 
\emph{fixed-template} methods refine
each section of a cross-task schema (Modular Prompt
Optimisation~\citep{sharma2026modular}); \emph{type-aware} methods
allocate repair operators by failure type
(TARA~\citep{liu2026rewritefixalltypeaware}); \emph{decomposition-based}
methods use a predefined taxonomy
(Task-Facet~\citep{juneja2025taskfacet}) or self-discovered factors
(aPSF~\citep{liu2026apsf}); and \emph{principle-aware} variants refine
via multi-aspect critique (CriSPO~\citep{he2024crispo}) or
generalisable criteria (ZERA~\citep{yi2025zera}). Although these
factorisations expose local editable units, they typically do not
explicitly link each realised edit to the examples it newly fixes or
breaks, or carry this edit--effect relation forward along the search
branch.

\paragraph{Reflective and evolutionary search with diversity.}
A third family uses natural-language reflection~\citep{shinn2023reflexion},
population diversity, or prompt-aware evaluation
scheduling~\citep{ma2026selectsmartermorepromptaware}.
Self-Refine~\citep{madaan2023selfrefine} and
ProMST~\citep{chen2024promst} show that iterative feedback can
substitute for an explicit scalar score, and population-based methods
add diversity: EvoPrompt~\citep{guo2024evoprompt} ports Genetic
Algorithm and Differential Evolution operators onto a prompt
population, Promptbreeder~\citep{fernando2023promptbreeder} co-evolves
mutation prompts alongside the task prompts they edit,
GEPA~\citep{agrawal2025gepa} (Genetic-Pareto) couples per-module
reflective mutation with a Pareto-frontier archive, and
PromptAgent~\citep{wang2024promptagent} runs Monte-Carlo tree search
with reflective error feedback. Outside prompt optimisation,
verifier-grounded planning has likewise paired typed structural patches
with deterministic constraint checking~\citep{wang2026planscheckverifiergroundedlearning}.
Yet even when archives or search trees
preserve multiple candidates, reflection often remains centred on the
current candidate's traces, without linking an edit's target and realised
operations to the examples newly fixed or broken in the parent-to-child
transition, or carrying this record forward to subsequent edits on the
same search branch.


\section{Method}
\label{sec:method}

\sepo is a multi-trajectory prompt optimiser: each iteration applies
one local edit to a structured prompt and records the edit together with
its example-level outcomes as an edit-effect lineage state, making the
optimisation process auditable and providing reusable feedback for later
edits on the same search branch. \sepo rests on three principles.
\textbf{(P1)}~\emph{Localised edits} over stable, addressable prompt
units give every edit an explicit target (\S\ref{sec:schema}).
\textbf{(P2)}~\emph{Edit-effect lineage feedback} uses two architect
calls, \emph{attribution} and \emph{patch}. Both calls share an evidence
packet comprising the parent's preceding edit, the examples that edit
newly fixed or broke, and the associated breadcrumb
(\S\ref{sec:editing}).
\textbf{(P3)}~\emph{Trajectory-diverse archiving} uses example-level
admission and Lexicase-based parent selection to determine which
candidate branches merit further search (\S\ref{sec:loop}).
Figure~\ref{fig:overview} and Algorithm~\ref{alg:sepo} detail one iteration.

\subsection{Problem Setup}
\label{sec:setup}

We assume a \emph{frozen} worker LLM $W$ for downstream inference
and a \emph{frozen} architect LLM $A$ used in the two roles described
in \S\ref{sec:editing}. Given a task $\mathcal{T}$ with a training set
$\mathcal{D}_{\mathrm{tr}}$ used during optimisation and a disjoint held-out
test split, we optimise a schema $s$ according to
\begin{equation}
s^{\star} \;=\; \argmax_{s} \; \mathbb{E}_{(x,y) \sim \mathcal{T}}
    \!\bigl[\,\mu(W(s,x),\, y)\bigr],
\label{eq:objective}
\end{equation}
where $\mu$ is a binary task metric and $W(s,x)$ is the worker's output
under the prompt rendered from $s$, with the expectation estimated by
the empirical mean over $\mathcal{D}_{\mathrm{tr}}$. For each $s$ we record
its score vector $\mathbf{c}(s)\!\in\!\{0,1\}^{|\mathcal{D}_{\mathrm{tr}}|}$
and correct set $C(s)=\{i:\mathbf{c}(s)_i=1\}$ under a budget
of $B$ metric calls.

\subsection{Prompt representation}
\label{sec:schema}

Unlike whole-prompt rewriters without local failure addresses, we
represent each candidate with a task-agnostic two-layer schema:
\begin{equation}
s \;=\; \bigl(\,\alpha,\; \{\rho_j\}_{j=1}^{m}\,\bigr),
\label{eq:schema}
\end{equation}
A top-level \emph{outline} $\alpha=(\alpha_1,\dots,\alpha_K)$ preserves
global intent, while each rule $\rho_j$ is attached to one step and
constrains local edits without fragmenting context. The attribution architect targets
one step or the whole outline, and the patch architect revises it
(\S\ref{sec:editing}). Linking each addressable edit to its example-level
effects provides the interface for edit-effect lineage feedback; other
representations can provide the same interface. The schema below
instantiates Eq.~\ref{eq:schema} for BBH Word Sorting with
$K{=}3$ outline steps and step-local rules.

\paragraph{Operator label space.}
After each patch call, a deterministic parser extracts the child's
outline steps and step-local rules; a structural diff against the
parent, together with the attributed target, then labels each changed
step or rule with one operator from $\mathcal{O}$,
\begin{equation}
o(c) \;\triangleq\; \textsc{InferOp}\bigl(\textsc{Diff}(p, c),\,\tau(c)\bigr),
\label{eq:operator}
\end{equation}
where $\textsc{InferOp}$ is deterministic and $o(c)$ collects one
label per changed location; $\tau(c){=}\tau$ is the attributed target
assigned to the child. These labels are recorded on the child
for use in the next iteration's breadcrumb. The vocabulary
$\mathcal{O}$, the structural diff $\textsc{Diff}(p, c)$
(Algorithm~\ref{alg:diff-schemas}), and the per-location classification
rule are in Appendix~\ref{app:operator-space}.

\begin{sepoexample}{Example schema (BBH Word Sorting, seed 45, test acc.\ 58\%)}
\footnotesize
\textbf{Step 1.} Understand the problem --- read carefully, parse all
entities, and decompose into sub-questions if needed.

\textbf{Step 2.} Reason step by step.
\begin{itemize}[label={-},leftmargin=14pt,topsep=1pt,itemsep=0pt]
  \item Copy the words EXACTLY as they appear after the marker
        ``\texttt{List:}''; do not substitute, invent, or import
        any words from training data or memory.
  \item Count the input words once, then sort in strict A--Z order
        in a single pass.
  \item CRITICAL: after sorting, do \emph{not} re-process the list
        (no re-reading or side-by-side checks)---both trigger
        hallucination loops on this worker; output immediately.
\end{itemize}
\textbf{Step 3.} Output the sorted list.
\begin{itemize}[label={-},leftmargin=14pt,topsep=1pt,itemsep=0pt]
  \item Output on the final line as
        ``\texttt{Answer: word1 word2 \ldots}'' --- exactly the
        extracted words, only reordered, separated by single
        spaces with no commas, newlines, numbering, or bullets.
\end{itemize}
\end{sepoexample}

\subsection{Lineage-guided reflective edit generation}
\label{sec:editing}

Each iteration of \sepo proceeds in five steps. We
(i)~draw a parent $p$ from the archive $\mathcal{A}$
(\S\ref{sec:loop});
(ii)~build an evidence packet from $p$ and its grandparent $g$,
comprising training examples that $p$'s last edit broke or newly
fixed, train probes, and a summary
$\ell$ of what the last edit changed;
(iii)~run an \emph{attribution} architect call returning a typed
target $\tau$ (Eq.~\ref{eq:attribution})---identifying either an edit
within one outline step or a structural change to the outline---
together with the failure reason;
(iv)~run a \emph{patch} architect call returning a full revised
schema $\widetilde{s}$, which is parsed into the child $c$;
(v)~infer the realised operators from the structural diff. After the
child is evaluated (\S\ref{sec:loop}), this completes its edit-effect
lineage tuple $(\tau,\,\ell,\,o(c),\,\mathrm{effect})$ for later
architect calls on the same search branch.

\paragraph{Balanced evidence packet.}
Both architect calls consume the same minibatch
$M = (M_{\mathrm{focus}}, M_{\mathrm{anchor}}, M_{\mathrm{train}})$,
where for $p$'s last edit
\begin{align*}
\textsc{newly\_broken}(p) &\triangleq \{i : \mathbf{c}(g)_i{=}1,\, \mathbf{c}(p)_i{=}0\}, \\
\textsc{newly\_fixed}(p)  &\triangleq \{i : \mathbf{c}(g)_i{=}0,\, \mathbf{c}(p)_i{=}1\}, \\
a_i &\triangleq |\mathcal{A}|^{-1} \!\sum_{s \in \mathcal{A}}\! \mathbf{c}(s)_i \in [0,1].
\end{align*}
$M_{\mathrm{focus}}$ samples $\textsc{newly\_broken}(p)$ with weights
$w_i \propto a_i$ (the archive agreement on example $i$), over-sampling failures on examples the archive
shows the worker can solve rather than failures likely caused by
capability limits. The architect uses these newly broken examples
together with train probes from $M_{\mathrm{train}}$ to infer why
the breakages occurred and to steer the next edit away from the same
failure mode; recovering in-capability failures directly supports
generalisation. $M_{\mathrm{anchor}}$ draws from $\textsc{newly\_fixed}(p)$,
preserving these recent gains. Slot sizes and
expressiveness arguments are in Appendix~\ref{app:hyperparams}.

\paragraph{Fallback without regressions.}
When \textsc{newly\_broken} is empty---at cold start
($g{=}\emptyset$) or when the parent's last edit introduced no
regression on $\mathcal{D}_{\mathrm{tr}}$---the iteration drops
$M_{\mathrm{focus}}$ and uses $M_{\mathrm{anchor}}$ together with
$M_{\mathrm{train}}$. In this case, no training regression is
available to explain, so the iteration continues improving on the
train probes while preserving one anchor.

\paragraph{Attribution and patch.}
The attribution architect consumes $(p, M, \ell)$ and diagnoses where
each error arises in the schema and why it occurred, returning a typed
target $\tau$ and failure rationale $r$,
\begin{equation}
(\tau,r) \;=\; A_{\mathrm{attr}}\!\bigl(p,\, M,\,
   \ell\bigr),
\label{eq:attribution}
\end{equation}
where $\tau$ takes one of five typed forms:
\begin{itemize}[leftmargin=*,itemsep=0pt,topsep=2pt]
\item \texttt{step <N>}: edit confined to step $N$ (the patch
architect may add, drop, refine, or merge rules inside that step,
or rewrite the step's top-level text);
\item \texttt{ADD step at <pos>}: insert a missing reasoning stage
at outline position \texttt{<pos>};
\item \texttt{SPLIT step <N>}: split two conflated responsibilities
inside step $N$;
\item \texttt{DROP step <N>}: remove step $N$;
\item \texttt{backbone}: rewrite the entire outline.
\end{itemize}
The attribution call returns only the edit location $\tau$ and failure
rationale $r$, not a new schema. The patch architect
takes $(p, \tau, r, M, \ell)$ and emits a full revised schema text
\begin{equation}
\begin{aligned}
\widetilde{s} \;&=\; A_{\mathrm{patch}}\!\bigl(p,\, \tau,\, r,\, M,\,
   \ell\bigr), \\
c \;&=\; \textsc{ParseSchema}(\widetilde{s}),
\end{aligned}
\label{eq:patch}
\end{equation}

\paragraph{Carry-forward lineage record.}
After producing and evaluating child $c$, we record its attributed target
$\tau$, diff-inferred realised operators $o(c)$, breadcrumb $\ell$, and
newly-fixed/newly-broken examples as the edit-effect lineage tuple
$(\tau,\ell,o(c),\mathrm{effect})$; $\mathrm{effect}$ stores the example
sets, while reporting uses only their counts. If $c$ is later selected as
a parent, $\ell$ and these transition examples jointly condition the next
attribution and patch calls. The \textsc{$-$Lineage} and
\textsc{$-$Breadcrumb} ablations (\S\ref{sec:q2-mechanism}) remove the
lineage evidence and the breadcrumb summary $\ell$ respectively.

\subsection{Example admission and Lexicase sampling}
\label{sec:loop}

To control evaluation cost, \sepo first evaluates each child on the
$M_{\mathrm{train}}$ probe slot and discards candidates that do not
strictly improve over the parent on those probes. A child that passes
this precheck is evaluated on the remaining training examples, reusing
the probe results to complete its full-training score vector, and is
admitted into the archive when at least one of the following holds:
\begin{equation}
\begin{cases}
\text{(a)}\;\; C(c) \!\setminus\! \textstyle\bigcup_{s \in \mathcal{A}} C(s) \neq \emptyset \\[2pt]
\text{(b)}\;\; |c| < |p| \\[2pt]
\text{(c)}\;\; \textstyle\sum \mathbf{c}(c) > \sum \mathbf{c}(p)
\end{cases}
\label{eq:admission}
\end{equation}
That is, the child adds unique coverage, is shorter than the parent, or
achieves a higher score than the parent. The unique-coverage criterion
preserves specialists that solve rare training examples.

We select the next parent from the archive using Lexicase
selection~\citep{spector2010assessment} with probability $0.8$ and
uniform random sampling with probability $0.2$. Lexicase filters
candidates using a randomly ordered sequence of training examples,
preserving specialists for difficult cases, while random sampling
retains additional exploration. Full budget accounting and
implementation details are given in Appendix~\ref{app:algorithm}.

\section{Experiments}
\label{sec:experiments}
\begin{table*}[t]
\centering
\footnotesize
\setlength{\tabcolsep}{3.2pt}
\renewcommand{\arraystretch}{1.4}
\arrayrulecolor{tableRule}

\resizebox{\textwidth}{!}{%
\begin{tabular}{@{}lrrrrrrrr|r|rrr|rr|r@{}}
\toprule
\multicolumn{1}{@{}l}{\textbf{Method}} &
\multicolumn{8}{c|}{\textbf{BBH}} &
\multicolumn{1}{c|}{\textbf{Code}} &
\multicolumn{3}{c|}{\textbf{MMLU-Pro}} &
\multicolumn{2}{c|}{\textbf{Reasoning}} &
\multicolumn{1}{c@{}}{\textbf{Avg.}} \\
\cmidrule(lr){2-9}\cmidrule(lr){10-10}\cmidrule(lr){11-13}\cmidrule(lr){14-15}\cmidrule(l){16-16}
& \multicolumn{1}{c}{Caus.} & \multicolumn{1}{c}{Date} & \multicolumn{1}{c}{Disamb.} & \multicolumn{1}{c}{Fall.} & \multicolumn{1}{c}{Deduct.} & \multicolumn{1}{c}{Obj.} & \multicolumn{1}{c}{Snarks} & \multicolumn{1}{c|}{Word.} & \multicolumn{1}{c|}{MBPP+} & \multicolumn{1}{c}{CS} & \multicolumn{1}{c}{Econ.} & \multicolumn{1}{c|}{Psych.} & \multicolumn{1}{c}{GSM-H} & \multicolumn{1}{c|}{HQA} & \\
\midrule

\rowcolor{llamaBlock}
\multicolumn{16}{c}{\textit{Worker LLM: Llama-3.1-8B-Instruct}} \\
Seed & \sepotabscore{37.5}{0.0} & \sepotabscore{66.7}{0.0} & \sepotabscore{61.7}{0.0} & \sepotabscore{44.5}{0.0} & \sepotabscore{45.0}{0.0} & \sepotabscore{82.5}{0.0} & \sepotabscore{72.4}{0.0} & \sepotabscore{0.0}{0.0} & \sepotabscore{57.3}{0.0} & \sepotabscore{44.7}{0.0} & \sepotabscore{57.5}{0.0} & \sepotabscore{56.7}{0.0} & \sepotabscore{35.8}{0.0} & \sepotabscore{25.0}{0.0} & \sepotabscore{49.1}{0.0} \\
\textsc{OPRO} & \sepotabscore{59.9}{3.2} & \sepotabscore{65.2}{2.3} & \sepotabscore{65.8}{4.1} & \sepotabscore{42.8}{3.8} & \sepotabscore{45.7}{5.2} & \sepotabscore{81.8}{9.0} & \sepotabllamabest{82.3}{3.2} & \sepotabscore{5.2}{0.6} & \sepotabscore{60.2}{1.4} & \sepotabscore{41.0}{2.6} & \sepotabscore{52.2}{2.6} & \sepotabscore{59.8}{2.6} & \sepotabscore{30.2}{4.9} & \sepotabscore{69.6}{8.7} & \sepotabscore{54.4}{3.9} \\
\textsc{APSF} & \sepotabscore{52.6}{8.3} & \sepotabscore{68.2}{1.2} & \sepotabscore{62.8}{2.4} & \sepotabscore{45.8}{7.3} & \sepotabscore{45.2}{4.3} & \sepotabscore{83.5}{1.3} & \sepotabscore{76.8}{4.0} & \sepotabscore{25.2}{12.0} & \sepotabscore{60.7}{2.3} & \sepotabllamabest{48.7}{4.5} & \sepotabscore{57.8}{2.8} & \sepotabllamabest{60.3}{2.3} & \sepotabscore{30.1}{2.5} & \sepotabscore{51.9}{11.2} & \sepotabscore{55.0}{4.7} \\
\textsc{MIPROv2} & \sepotabllamabest{63.3}{2.9} & \sepotabscore{67.8}{2.3} & \sepotabscore{65.2}{4.7} & \sepotabscore{49.3}{8.6} & \sepotabscore{44.5}{2.2} & \sepotabscore{79.0}{1.7} & \sepotabscore{76.6}{1.4} & \sepotabscore{3.3}{0.3} & \sepotabscore{56.0}{2.2} & \sepotabscore{47.3}{3.0} & \sepotabscore{52.7}{1.9} & \sepotabscore{59.3}{2.6} & \sepotabscore{37.2}{2.1} & \sepotabscore{62.5}{1.1} & \sepotabscore{54.6}{2.6} \\
\textsc{MPO} & \sepotabscore{52.6}{3.3} & \sepotabscore{73.2}{6.9} & \sepotabscore{67.0}{2.2} & \sepotabscore{44.8}{4.0} & \sepotabllamabest{47.7}{4.4} & \sepotabscore{83.7}{5.1} & \sepotabscore{75.0}{2.1} & \sepotabscore{56.5}{2.2} & \sepotabscore{59.0}{1.7} & \sepotabscore{47.7}{1.4} & \sepotabscore{55.5}{4.6} & \sepotabscore{56.0}{0.9} & \sepotabscore{33.8}{2.0} & \sepotabscore{67.6}{0.9} & \sepotabscore{58.6}{3.0} \\
\textsc{GEPA} & \sepotabscore{59.9}{3.8} & \sepotabscore{75.3}{6.1} & \sepotabscore{64.5}{6.6} & \sepotabscore{47.5}{2.2} & \sepotabscore{47.0}{3.9} & \sepotabscore{86.0}{3.1} & \sepotabscore{77.3}{2.1} & \sepotabscore{39.3}{18.3} & \sepotabscore{60.0}{1.3} & \sepotabscore{46.5}{3.0} & \sepotabscore{54.5}{1.3} & \sepotabscore{59.3}{4.2} & \sepotabscore{37.2}{2.1} & \sepotabscore{68.4}{1.9} & \sepotabscore{58.8}{4.3} \\
\sepo & \sepotabscore{61.1}{4.6} & \sepotabllamabest{76.5}{4.3} & \sepotabllamabest{68.8}{4.6} & \sepotabllamabest{50.8}{3.6} & \sepotabscore{47.3}{2.1} & \sepotabllamabest{87.5}{0.9} & \sepotabllamabest{82.3}{1.6} & \sepotabllamabest{58.2}{1.3} & \sepotabllamabest{62.6}{1.8} & \sepotabscore{46.7}{1.8} & \sepotabllamabest{58.8}{2.8} & \sepotabscore{58.7}{1.8} & \sepotabllamabest{37.6}{1.8} & \sepotabllamabest{70.2}{3.8} & \sepotabllamabest{61.9}{2.6} \\

\addlinespace[2pt]
\rowcolor{qwenBlock}
\multicolumn{16}{c}{\textit{Worker LLM: Qwen3-8B (HQA $=$ HotpotQA Hard)}} \\
Seed & \sepotabscore{56.9}{0.0} & \sepotabscore{83.8}{0.0} & \sepotabscore{44.3}{0.0} & \sepotabscore{46.0}{0.0} & \sepotabscore{84.7}{0.0} & \sepotabscore{94.5}{0.0} & \sepotabscore{76.0}{0.0} & \sepotabscore{5.3}{0.0} & \sepotabscore{15.7}{0.0} & \sepotabscore{68.7}{0.0} & \sepotabscore{73.8}{0.0} & \sepotabscore{66.2}{0.0} & \sepotabscore{48.6}{0.0} & \sepotabscore{35.5}{0.0} & \sepotabscore{57.1}{0.0} \\
\textsc{OPRO} & \sepotabscore{60.6}{3.2} & \sepotabscore{78.7}{1.8} & \sepotabscore{66.0}{7.0} & \sepotabscore{47.3}{3.6} & \sepotabscore{86.7}{4.2} & \sepotabscore{97.0}{2.2} & \sepotabscore{82.3}{1.6} & \sepotabscore{7.7}{1.3} & \sepotabscore{66.7}{1.0} & \sepotabscore{69.3}{1.2} & \sepotabscore{75.0}{2.2} & \sepotabscore{67.7}{3.0} & \sepotabscore{69.0}{0.9} & \sepotabscore{32.8}{2.3} & \sepotabscore{64.8}{2.5} \\
\textsc{APSF} & \sepotabscore{55.7}{4.5} & \sepotabscore{84.3}{2.3} & \sepotabscore{65.0}{1.3} & \sepotabscore{64.2}{2.0} & \sepotabscore{79.5}{1.3} & \sepotabscore{94.3}{2.3} & \sepotabscore{77.6}{5.5} & \sepotabscore{31.5}{2.6} & \sepotabscore{68.7}{1.2} & \sepotabscore{66.7}{2.8} & \sepotabscore{75.2}{2.8} & \sepotabscore{66.7}{3.1} & \sepotabscore{53.8}{7.5} & \sepotabscore{38.9}{1.1} & \sepotabscore{65.9}{2.9} \\
\textsc{MIPROv2} & \sepotabscore{57.7}{1.3} & \sepotabscore{86.0}{2.2} & \sepotabscore{77.0}{2.3} & \sepotabscore{56.7}{14.4} & \sepotabscore{83.5}{3.0} & \sepotabscore{97.3}{0.8} & \sepotabscore{81.2}{2.8} & \sepotabscore{6.7}{6.7} & \sepotabscore{68.3}{1.2} & \sepotabscore{69.5}{2.2} & \sepotabscore{76.0}{0.9} & \sepotabqwenbest{68.3}{1.2} & \sepotabscore{62.8}{0.8} & \sepotabscore{32.7}{1.1} & \sepotabscore{66.0}{2.9} \\
\textsc{MPO} & \sepotabscore{56.7}{3.0} & \sepotabscore{86.0}{4.3} & \sepotabscore{70.5}{7.2} & \sepotabscore{67.7}{9.0} & \sepotabscore{80.5}{3.1} & \sepotabscore{97.7}{0.8} & \sepotabscore{78.1}{3.6} & \sepotabscore{50.5}{0.9} & \sepotabscore{67.5}{1.7} & \sepotabscore{65.0}{1.7} & \sepotabqwenbest{77.5}{1.3} & \sepotabscore{64.0}{2.3} & \sepotabscore{62.6}{0.9} & \sepotabscore{35.3}{2.9} & \sepotabscore{68.5}{3.1} \\
\textsc{GEPA} & \sepotabqwenbest{61.1}{1.5} & \sepotabscore{86.3}{1.2} & \sepotabscore{74.5}{2.6} & \sepotabqwenbest{78.5}{7.1} & \sepotabscore{88.7}{2.5} & \sepotabscore{98.5}{1.3} & \sepotabscore{78.6}{3.9} & \sepotabscore{51.7}{5.3} & \sepotabscore{64.0}{1.3} & \sepotabscore{69.0}{2.2} & \sepotabscore{73.3}{1.2} & \sepotabscore{67.5}{1.3} & \sepotabscore{67.5}{1.1} & \sepotabscore{36.5}{0.7} & \sepotabscore{71.1}{2.4} \\
\sepo & \sepotabscore{58.9}{3.7} & \sepotabqwenbest{90.0}{1.7} & \sepotabqwenbest{77.7}{1.4} & \sepotabscore{75.7}{4.9} & \sepotabqwenbest{92.3}{3.0} & \sepotabqwenbest{99.0}{0.9} & \sepotabqwenbest{82.6}{3.0} & \sepotabqwenbest{54.7}{3.9} & \sepotabqwenbest{72.7}{1.1} & \sepotabqwenbest{70.8}{1.3} & \sepotabscore{76.2}{0.8} & \sepotabscore{66.7}{2.0} & \sepotabqwenbest{70.0}{8.0} & \sepotabqwenbest{39.1}{0.6} & \sepotabqwenbest{73.3}{2.6} \\

\bottomrule
\end{tabular}%
}

\arrayrulecolor{black}
\caption{\textbf{Held-out test accuracy (\%)} on the 14-task suite;
mean over three seeds (subscript: std). Tinted bold cells mark the best
per Worker LLM block. HotpotQA: Medium (Llama), Hard/HQA (Qwen).}
\label{tab:sepo-per-task}
\end{table*}

\paragraph{Experimental questions.}
We evaluate \sepo through four questions:
\begin{itemize}[leftmargin=*,itemsep=1pt,topsep=2pt]
\item \textbf{Q1: Held-out effectiveness.} Under a matched API-only
optimisation and deployment protocol, does \sepo produce stronger final
prompts than strong prompt optimisers?
\item \textbf{Q2: Edit-diagnostic mechanism.} Are \sepo's gains driven
by its edit-aware diagnostic loop, rather than by generic reflection,
static factorisation, or unlocalised prompt rewriting?
\item \textbf{Q3: Behavioral scope and auditability.} Where does
\sepo help or fail, and do its edit traces make these outcomes
inspectable at evidence, target, operator, and
example-level effect?
\item \textbf{Q4: Cost-effectiveness.} What optimisation-time and
deployment-time cost do \sepo's gains require, and does \sepo remain
competitive on accuracy--cost trade-offs?
\end{itemize}

\subsection{Experimental Setup and Protocol}
\label{sec:exp-setup}

\paragraph{Benchmarks.}
Our primary suite contains 14 held-out tasks spanning four families:
\begin{itemize}[leftmargin=*,itemsep=1pt,topsep=2pt]
\item \textbf{BBH (8 tasks)}~\citep{suzgun2023bbh,srivastava2023beyond}: causal judgement,
date understanding,
disambiguation QA, formal fallacies, logical deduction over seven
objects, reasoning about colored objects, snarks, and word sorting,
chosen to cover BBH's distinct capability types (world-knowledge
causal inference, temporal reasoning, linguistic disambiguation,
formal logic, multi-step deduction, symbolic manipulation, sarcasm
understanding, and algorithmic sorting).
\item \textbf{MMLU-Pro (3 tasks)}~\citep{wang2024mmlupro}: computer science, economics, and
psychology, spanning a technical, an economic--quantitative, and a
behavioural--social domain.
\item \textbf{Code:} MBPP+~\citep{liu2023evalplus}, scored by pass@1~\citep{chen2021evaluating}.
\item \textbf{Reasoning:} GSM-Hard~\citep{gao2023pal} and
HotpotQA~\citep{yang2018hotpotqa} (Medium split under
Llama, Hard split under Qwen).
\end{itemize}

\paragraph{Models.}
We run our pipeline with two frozen Workers of different
architecture and capability to test cross-family generalisation:
\begin{itemize}[leftmargin=*,itemsep=1pt,topsep=2pt]
\item \textbf{Worker LLMs:} Llama-3.1-8B-Instruct~\citep{grattafiori2024llama}
and Qwen3-8B~\citep{yang2025qwen3}, temperature $0.0$ for deterministic metric
evaluation.
\item \textbf{Architect LLM:} Qwen3.5-397B-A17B~\citep{qwen2026qwen35modelcard},
temperature $0.7$ to
encourage diverse candidate edits.
\end{itemize}

\paragraph{Baselines and protocol.}
We compare \sepo against six API-only baselines---an unoptimised
\textsc{Seed} prompt, \textsc{APSF}, \textsc{OPRO}, \textsc{MIPROv2},
\textsc{MPO}, and \textsc{GEPA}---under a shared budget of $2000$
\emph{metric calls} per task and seed (architect LLM calls are
\emph{not} counted toward this budget). This matches DSPy / MIPROv2's
$\texttt{max\_metric\_calls}{=}2000$~\citep{khattab2024dspy, opsahlong2024miprov2}.
All methods start from the same task-agnostic
understand--reason--answer seed on every benchmark.
The cross-worker HotpotQA column is not comparable: Llama uses Medium
because limited worker headroom compresses performance on Hard, whereas
Qwen uses Hard (Appendix~\ref{app:hqa-hard-sensitivity}).  The minibatch's three slots
$(M_{\mathrm{focus}}, M_{\mathrm{anchor}}, M_{\mathrm{train}})$ have sizes
$(n_f, n_a, n_{tr}) = (2, 1, 3)$.  We report macro accuracy (unweighted 14-task mean), per-baseline win rate, optimisation/deployment
tokens, and final prompt length; full configs in
Appendix~\ref{app:model-config}.

\FloatBarrier
\subsection{Q1: \sepo Beats All Baselines on Both Workers in Macro Accuracy}
\label{sec:q1-main-results}

\begin{figure}[!htbp]
\centering
\includegraphics[width=0.88\columnwidth]{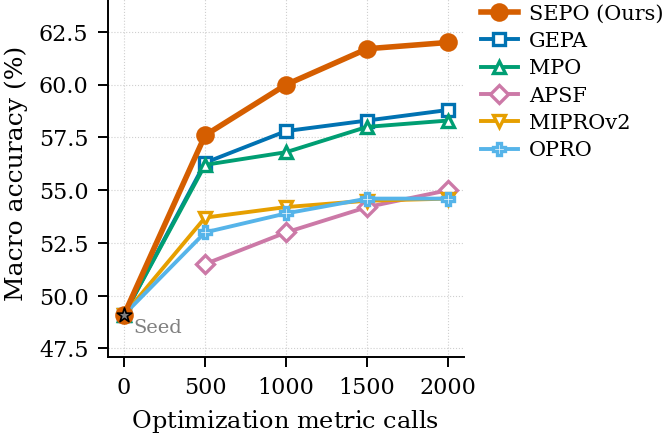}
\caption{\textbf{Budget-matched accuracy curves} (Llama-3.1-8B-Instruct).
Test accuracy at metric-call budgets
$0/500/1000/1500/2000$.}
\label{fig:rollout-budget-curve}
\end{figure}

We evaluate whether structured, lineage-aware editing
yields stronger held-out prompts than competing optimisers under a
matched API-only protocol; per-task results are in
Table~\ref{tab:sepo-per-task}.

\begin{enumerate}[leftmargin=*,itemsep=1pt,topsep=2pt]
\item \sepo reaches $61.9\%$ on Llama ($+3.1$\,pp over \textsc{GEPA};
$\geq 3.3$\,pp over the rest) and $73.3\%$ on Qwen ($+2.2$\,pp over
\textsc{GEPA}; $\geq 4.8$\,pp over the rest), beating \textsc{GEPA} on
13/14 (Llama) and 11/14 (Qwen), and \mbox{$\geq 11/14$} against every
other baseline under both workers.
\item The gains span procedural and reasoning families---\textbf{BBH}
($+4.5/+1.6$\,pp), \textbf{Code} (MBPP+, $+2.6/+8.7$\,pp), and
\textbf{Reasoning} (HotpotQA $+$ GSM-Hard, $+1.1/+2.5$\,pp)---ruling
out a single-domain artefact, and concentrate on procedural tasks
where one failed local sub-step can invalidate the final answer.
Word Sorting is the clearest case: \sepo is best under both workers
(Llama $58.2\%$ / Qwen $54.7\%$, $+18.9$\,pp over \textsc{GEPA} on Llama; the
seed collapses to $0.0\%$ on Llama), and MBPP+ shows the same pattern
($62.6\%$ / $72.7\%$).  On the MMLU-Pro multidomain tasks, \sepo
remains within the leading cluster but with thinner per-task margins,
illustrating that the magnitude of prompt-optimization gains varies
across task families.

\item \sepo also reaches a given accuracy with fewer metric calls: at
every metric-call budget in Figure~\ref{fig:rollout-budget-curve} it
stays at or above every baseline, so its lead holds across the budget
axis and is not an artefact of the $2000$-metric-call cap.
\end{enumerate}


\subsection{Q2: Each Principle (P1/P2/P3) of \sepo is Load-Bearing}
\label{sec:q2-mechanism}

We ablate three principles to test whether
\sepo's gains are driven by its edit-aware diagnostic loop, not by generic reflection or unlocalised rewriting.  We isolate four
single-removal variants---\textsc{$-$Attribution} (drops P1),
\textsc{$-$Breadcrumb} and \textsc{$-$Lineage} (drop P2 at two
strengths), and \textsc{$-$Multitraj} (drops P3)---and analyse their
accuracy, length, and acceptance signatures.

We ablate on five tasks spanning code, math,
temporal, linguistic, and multidomain tasks (MBPP+, GSM-Hard,
BBH Date, BBH Snarks, and MMLU-Pro CS).  Every single-removal
variant costs $1.8$--$4.7$\,pp on the five-benchmark diagnostic
average, with the largest damage on procedural multi-sub-skill tasks.
Matched proposal-level controls further isolate these roles: random
targets reduce net gain from $+2.5$ to $-1.7$ examples per $100$
proposals, while replacing newly broken cases with random current errors
reduces it to $+0.8$ (Appendix~\ref{app:proposal-controls}).

\begin{table}[!htbp]
\centering
\footnotesize
\setlength{\tabcolsep}{6pt}
\renewcommand{\arraystretch}{1.15}
\arrayrulecolor{tableRule}
\begin{tabular}{lcrrrr}
\toprule
\textbf{Variant} & \textbf{P} & \textbf{Acc.} & $\boldsymbol{\sigma}$ & $\boldsymbol{\Delta}$\textbf{Acc} & \textbf{Len.} \\
\midrule
\rowcolor{llamaBlock}
Full \sepo            & ---  & \textbf{61.1} & 2.2 & ---    & 228 \\
\textsc{$-$Attribution} & P1 & 58.0 & \textbf{2.5} & $-3.1$ & \textbf{450} \\
\textsc{$-$Lineage}    & P2 & 58.2 & 1.5 & $-2.9$ & 292 \\
\textsc{$-$Breadcrumb} & P2 & 59.3 & 2.0 & $-1.8$ & 235 \\
\textsc{$-$Multitraj}     & P3 & 56.4 & 2.3 & $-4.7$ & \emph{198} \\
\bottomrule
\end{tabular}
\arrayrulecolor{black}
\caption{\textbf{Aggregate ablation} on the five-benchmark diagnostic
subset (Llama-3.1-8B-Instruct; 3 seeds; $\sigma$: std).  Acc.\ \%, Len.\ tokens;
\textbf{bold}/\emph{italic}: cells discussed in P1/P3.  Breakdowns:
Appendices~\ref{app:ablation-design-map},~\ref{app:ablation-per-task}.}
\label{tab:sepo-mechanism-ablations}
\end{table}

\begin{figure}[!htbp]
\centering
\includegraphics[width=\columnwidth]{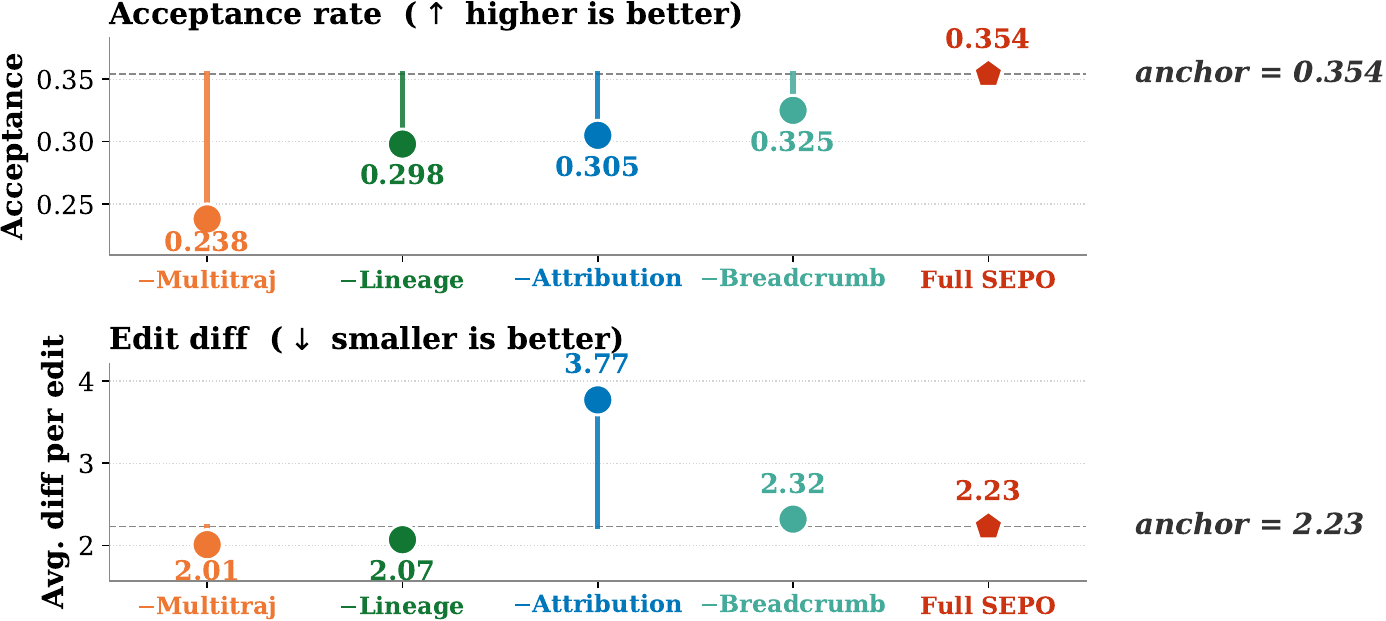}
\caption{\textbf{Edit-quality signature.}  Stem length $=$ each
variant's drift from the Full \sepo anchor (dashed) in acceptance rate
(top) and average edit size (bottom), sorted by acceptance; colour
$=$ principle (P1/P2/P3).}
\label{fig:ablation-acc-len}
\end{figure}

Each variant leaves a distinct signature in
Table~\ref{tab:sepo-mechanism-ablations} and Figure~\ref{fig:ablation-acc-len}; P1--P3 below.

\paragraph{P1 --- typed targeting localises edits.}
Removing the attribution architect (\textsc{$-$Attribution})
inflates prompt length $1.97\times$ ($228{\to}450$ tokens), drops
accuracy $3.1$\,pp (Table~\ref{tab:sepo-mechanism-ablations}), grows
each edit $\sim$$1.7\times$ in diff size ($2.23{\to}3.77$),
and lowers acceptance from $0.354$ to $0.305$ at matched
compute ($318$ vs $320$ candidate edits;
Figure~\ref{fig:ablation-acc-len}).  Both signatures
share one mechanism: \sepo edits one step per iteration and
accumulates, with $\tau$ enforcing the per-iteration target.
Without it, the patch architect lacks a single target, so it
touches multiple steps at once --- larger edits inflate length
and clear admission less often.  The damage concentrates on
\textbf{linguistic-reasoning tasks}
(\S\ref{sec:q1-main-results}), whose compositional rule
scaffolds are most fragile under multi-region rewrites
(BBH-Snarks $-8.5$\,pp,
Table~\ref{tab:sepo-mechanism-pertask}).

\begin{table}[!htbp]
\centering
\footnotesize
\renewcommand{\arraystretch}{1.15}
\arrayrulecolor{tableRule}
\setlength{\tabcolsep}{6pt}
\begin{tabular}{@{}lcccc@{}}
\toprule
& \textbf{P1} & \multicolumn{2}{c}{\textbf{P2}} & \textbf{P3} \\
\cmidrule(lr){2-2}\cmidrule(lr){3-4}\cmidrule(l){5-5}
\textbf{Task} & \textsc{$-$Attr} & \textsc{$-$Lin} & \textsc{$-$Brd} & \textsc{$-$MT} \\
\midrule
BBH Date    & $-2.5$           & \textbf{$-4.3$}  & $-0.5$           & $-4.7$           \\
BBH Snarks  & \textbf{$-8.5$}  & $-3.8$           & \textbf{$-5.7$}  & \textbf{$-7.3$}  \\
MBPP+       & $-1.7$           & $-4.1$           & $-2.8$           & $-7.0$           \\
GSM-Hard    & $-1.2$           & $-2.6$           & $-0.6$           & $-2.5$           \\
MMLU-Pro CS & \emph{$-1.7$}    & \emph{$+0.1$}    & \emph{$+0.3$}    & \emph{$-2.2$}    \\
\bottomrule
\end{tabular}
\arrayrulecolor{black}
\caption{\textbf{Per-task $\Delta$Acc} (five benchmarks,
Llama-3.1-8B-Instruct).  \textbf{Bold}: largest drop per variant; italic row:
scope-negative case.  Full heatmap: Appendix~\ref{app:ablation-per-task}.}
\label{tab:sepo-mechanism-pertask}
\end{table}
\FloatBarrier

\paragraph{P2 --- edit-effect lineage feedback supports net-positive edits.}
\textsc{$-$Breadcrumb} (removes the prior-edit trace $\ell$) costs
$1.8$\,pp, while \textsc{$-$Lineage} (removes $\ell$ and the
newly-fixed/newly-broken transition information) costs $2.9$\,pp
(Table~\ref{tab:sepo-mechanism-pertask}). These nested ablations indicate
complementary roles: $\ell$ summarises what changed, while the transition
examples identify which cases were newly fixed or broken; together they
guide the next edit. The
effect is most pronounced on procedural, multi-step tasks that require
sequencing local rules, whereas MMLU-Pro CS is roughly indifferent,
illustrating that the benefit of lineage evidence varies with task
structure.

\paragraph{P3 --- archive escapes local optima of greedy editing.}
\textsc{$-$Multitraj} disables the archive and Lexicase selection
(best-only, greedy single-trajectory editing) and drops $4.7$\,pp at a
\emph{shorter} prompt ($\times 0.87$, $228\to198$ tokens) with
acceptance falling from $0.354$ to $0.238$
(Table~\ref{tab:sepo-mechanism-ablations},
Figure~\ref{fig:ablation-acc-len}); the result is consistent with a reduced
exploration radius rather than prompt bloat.  \sepo's archive mitigates
this by keeping
diverse candidates alive: \textbf{Admission} (Eq.~\ref{eq:admission})
retains unique-coverage, shorter, or higher-scoring children rather
than ranking on a single scalar, and \textbf{Lexicase} parent
selection keeps rare-example solvers as future parents
(Appendix~\ref{app:lexicase-vs-freq}).  The auxiliary controls show that
retaining and revisiting multiple candidate branches provides the dominant
benefit, while Lexicase selection and unique-coverage admission each add a
modest average gain.
\FloatBarrier

\subsection{Q3: Behavioral Scope and Auditability}
\label{sec:q3-scope-audit}

\paragraph{Auditability.}
\sepo logs a typed tuple $(\tau,\ell,o(c),\mathrm{effect})$ for every
child (\S\ref{sec:editing}); Figure~\ref{fig:audit-card} reads one
packet end to end.

\begin{center}
\begin{tcolorbox}[
  enhanced, colback=promptBg, colframe=promptFrame,
  coltitle=white, colbacktitle=promptFrame, fonttitle=\bfseries\footnotesize,
  title={Audit packet for one \sepo\ edit (BBH Word Sorting)},
  arc=2pt, boxrule=0.5pt, left=5pt,right=5pt,top=2pt,bottom=2pt,
  width=\columnwidth
]
\scriptsize
\textbf{Local target:} $\tau=\texttt{step\,3}$, the \textbf{output formatting rule}.\\[-1pt]
\textbf{Parent breadcrumb $\ell$:} previous edit added count/sort rules under step\,2:
\colorbox{diffdel}{\parbox{0.88\linewidth}{\textbf{Count the input words once, then sort in strict A--Z order in a single pass.}}}\\[0pt]
\textbf{\textcolor{warnfg}{Observed side effect:}} $+15$ newly fixed on sorting, but with sorting now reliable the worker \textcolor{warnfg}{\textbf{renders output freely without format constraints}}, \textcolor{warnfg}{\textbf{newly breaking 4}} correctly-sorted cases on format.
\begin{itemize}[label={--},leftmargin=11pt,topsep=0pt,itemsep=0pt,parsep=0pt]
  \item \textbf{\textcolor{warnfg}{Newly broken (focus):}} \texttt{Ex\,\#13} is \textcolor{tableRule}{\textbf{sorted correctly}} but renders with \textcolor{warnfg}{\textbf{commas}}; \texttt{Ex\,\#16} breaks a word onto a \textcolor{warnfg}{\textbf{new line}} --- both fail the single-line answer.
  \item \textbf{\textcolor{evgreen}{Anchor kept:}} \texttt{Ex\,\#17} --- already \textcolor{evgreen}{\textbf{correct sort and correct format}}, preserved by adding rather than overwriting.
  \item \textbf{\textcolor{evgreen}{Train probe:}} an independent comma-joined case is \textcolor{evgreen}{\textbf{likewise fixed}}, so the rule is \textcolor{evgreen}{\textbf{not confined}} to focus examples.
\end{itemize}
\textbf{Patch on Step 3:} one operation per failure mode.\\[0pt]
\colorbox{hlrefine}{\parbox{0.88\linewidth}{$\circlearrowright$ \textsc{RewriteStep}: rename step\,3 to \textbf{``Output the sorted list.''}}}\\[0pt]
\colorbox{diffadd}{\parbox{0.88\linewidth}{$+$ \textsc{AddRule}: output on the final line as \textbf{``Answer: word1 word2 \ldots''} --- exactly the extracted words, only reordered, \textbf{separated by single spaces} with no commas, newlines, numbering, or bullets (closes \texttt{Ex\#13}/\texttt{Ex\#16}).}}\\[0pt]
\textbf{Operator:} $o(c){=}\{\textsc{RewriteStep},\textsc{AddRule}\}$ \hfill
\textbf{Effect:} \textcolor{evgreen}{\textbf{$+17$ fixed, $-6$ broken}}, $20{\to}31$, \textcolor{evgreen}{kept}.
\end{tcolorbox}
\captionof{figure}{\textbf{A local audit packet} (representative; full
edit tree in Figure~\ref{fig:edit-tree}, Appendix~\ref{app:qualitative}).}
\label{fig:audit-card}
\end{center}

The breadcrumb summarises the parent's preceding edit, the focus examples
show which cases broke, the patch addresses those cases, and the anchor plus
train probe test whether the benefit extends beyond the focus cases.
Together, these signals make the edit packet both retrospectively
inspectable and useful for guiding subsequent edits. The
walkthrough, the edit tree (Figure~\ref{fig:edit-tree}), and
example final prompts are in Appendix~\ref{app:qualitative}.

\begin{table}[H]
\centering
\footnotesize
\setlength{\tabcolsep}{5pt}\renewcommand{\arraystretch}{1.15}
\arrayrulecolor{tableRule}
\resizebox{\columnwidth}{!}{%
\begin{tabular}{@{}lrr@{\hspace{1.4em}}lrr@{}}
\toprule
\textbf{Rule-level op} & \textbf{\#} & \textbf{share} &
\textbf{Structural op} & \textbf{\#} & \textbf{share} \\
\midrule
\textsc{AddRule}    & 423 & 75.4\% & \textsc{RewriteStep} & 34 & 7.3\% \\
\textsc{DropRule}   & 147 & 31.7\% & \textsc{AddStep}     & 14 & 3.0\% \\
\textsc{RefineRule} & 126 & 37.2\% & \textsc{SplitStep}   & 18 & 4.0\% \\
\textsc{MergeRules} &   3 &  0.6\% & \textsc{Backbone}    &  1 & 0.3\% \\
\bottomrule
\end{tabular}%
}
\arrayrulecolor{black}
\caption{\textbf{Typed edit operations} (14-task suite, 3 seeds;
$N\!\approx\!338$ edits).  \emph{\#} $=$ total occurrences;
\emph{share} $=$ \% of edits containing the op (one edit may use
several ops and repeat one, so shares sum to $>$$100\%$).}
\label{tab:edit-distribution}
\end{table}

Aggregated over the suite, the same log also characterises \emph{how}
\sepo edits (Table~\ref{tab:edit-distribution}).
{\textbf{Edits stay overwhelmingly step-local}}:
structural rewrites that add, split, or replace outline steps remain
rare, so localised editing is \sepo's dominant editing mode, in contrast
to monolithic whole-prompt rewriting.  Each edit is also composite---most add a rule while many
also drop or refine one---so \sepo actively prunes and revises rather
than merely accreting.

\begin{table}[!b]
\centering
\footnotesize
\setlength{\tabcolsep}{2.6pt}
\renewcommand{\arraystretch}{1.15}
\arrayrulecolor{tableRule}
\resizebox{\columnwidth}{!}{%
\begin{tabular}{lrrrrrrrrr}
\toprule
& & \multicolumn{3}{c}{\textbf{Opt.\ (M tok)}} & \textbf{Dep.} & \textbf{Len.} & $\boldsymbol{\rho}$ & \textbf{\$/cell} & $\boldsymbol{\rho_{\$}}$ \\
\cmidrule(lr){3-5}
\textbf{Method} & \textbf{Acc.} & \textbf{Arch.} & \textbf{Work.} & \textbf{Total} & (k) & (tok) & & (USD) & (pp/\$) \\
\midrule
\textsc{OPRO}    & 54.4 & 6.84 & 3.11 & 9.95 & 244.0 & 358 & 0.53 & 1.66 & 3.2 \\
\textsc{MIPROv2} & 54.6 & 3.52 & 2.60 & 6.12 & 208.4 & 230 & 0.90 & 0.89 & 6.2 \\
\textsc{APSF}    & 55.0 & 0.69 & 2.13 & \textbf{2.82} & \textbf{157.8} & \textbf{65} & 2.09 & 0.19 & 31.1 \\
\textsc{MPO}     & 58.6 & 0.19 & 3.40 & 3.59 & 261.0 & 416 & 2.65 & \textbf{0.15} & 63.3 \\
\textsc{GEPA}    & 58.8 & 0.23 & 3.88 & 4.11 & 423.7 & 1{,}324 & 2.36 & 0.18 & 53.9 \\
\rowcolor{llamaBlock}
\sepo            & \textbf{61.9} & 0.34 & 2.59 & 2.93 & 203.2 & 236 & \textbf{4.37} & 0.17 & \textbf{75.3} \\
\bottomrule
\end{tabular}}
\arrayrulecolor{black}
\caption{\textbf{Efficiency decomposition} (Llama-3.1-8B-Instruct,
14-task macro). $\Delta\text{Acc}$ is each method's gain over the Seed
prompt ($49.1\%$). $\rho{=}\Delta\text{Acc}/\text{Total Opt.}$ (pp/M tok);
$\rho_{\$}{=}\Delta\text{Acc}/(\text{\$/cell})$ (pp/USD); list prices,
exchange rate, and per-benchmark breakdowns in
Appendix~\ref{app:cost-extra}.}
\label{tab:sepo-efficiency}
\end{table}
\subsection{Q4: \sepo Dominates Cost-Effectiveness across Architect and Worker Channels}
\label{sec:q4-cost-effectiveness}
We compare \sepo's optimisation- and deployment-time costs in
Table~\ref{tab:sepo-efficiency}.
\paragraph{\textbf{\sepo achieves the best $\rho$ under a shared budget
by trimming both the architect and worker channels, while every
baseline overspends on at least one.}}
Table~\ref{tab:sepo-efficiency} decomposes the cost drivers and reports
the cost-effectiveness ratio $\rho$.  \sepo reaches $\rho{=}4.37$ ---
$1.6\times$ the next-best \textsc{MPO} ($2.65$), $1.9\times$
\textsc{GEPA} ($2.36$), $2.1\times$ \textsc{APSF} ($2.09$; lowest
total opt.\ but $6.9$\,pp weaker), and $5$--$8\times$
\textsc{MIPROv2} ($0.90$) / \textsc{OPRO} ($0.53$).  \sepo trims
\emph{both} channels at once: localised edits hold the architect
side to $0.34$M tokens per cell ($10$--$20\times$ less than
\textsc{OPRO} and \textsc{MIPROv2}), and a $236$-token prompt holds
worker optimisation to $2.59$M (vs.\ \textsc{GEPA} $3.88$M,
\textsc{MPO} $3.40$M) and deployment to $203$k per cell (half of
\textsc{GEPA}'s $424$k).  Each baseline overspends on one channel ---
\textsc{OPRO}/\textsc{MIPROv2} on the architect, \textsc{GEPA}/\textsc{MPO}
on worker length.  At list prices, \sepo costs \$0.17/cell
($\rho_{\$}{=}75.3$\,pp/USD) --- $\sim$10$\times$ cheaper than
\textsc{OPRO} (\$1.66), and within \$0.02 of \textsc{MPO}/\textsc{GEPA}
(\$0.15/\$0.18) while $+3.1$\,pp more accurate.

\section{Conclusion}
\label{sec:conclusion}
We presented \sepo and its core mechanism, \emph{edit-effect lineage
feedback}, which links each local edit's target and realised operations
to the examples it newly fixes or breaks and reuses this record in later
architect calls on the same search branch. A two-layer prompt schema
provides a stable, addressable implementation. Across 14 tasks and two
workers, \sepo beats six API-only baselines ($+3.1/+2.2$\,pp over the
strongest, \textsc{GEPA}) and lies on both the optimisation- and
test-time Pareto frontiers, yielding a stronger accuracy--cost trade-off.

\section*{Limitations}
\label{sec:limitations}

\textbf{Instruction-space scalability.} Our evaluation covers prompt
lengths and instruction spaces representative of existing
prompt-optimization work.  Whether \sepo's efficiency advantage extends
to substantially longer prompts or more compositional agentic skills
involving multiple steps, tool calls, or state tracking remains open.
\textbf{Task-tailored schemas.} By design, \sepo tailors a schema per task, yielding a task-specialised
artefact rather than a universal prompt; cross-task transfer is left to
future work.
\textbf{Evaluation setting.} We evaluate \sepo only on single-turn,
single-call, text-based tasks; its structured, attributable edits may
suit broader domains---agentic pipelines and multimodal
prompts---which we leave to future work.

\section*{Acknowledgments}
This work is supported in part by the Guangdong Basic and Applied Basic
Research Foundation under Grant No.~2025A1515012968, in part by the
Shenzhen Science and Technology Program under Grant
No.~JCYJ20240813113502004, in part by the National Natural Science
Foundation of China under Grant No.~62001412, in part by Shenzhen
Stability Science Program 2023, in part by the Guangdong Provincial Key
Laboratory of Future Networks of Intelligence (Grant
No.~2022B1212010001), and in part by the Shenzhen Key Lab of Crowd
Intelligence Empowered Low-Carbon Energy Network (Grant
No.~ZDSYS20220606100601002).

\bibliography{references}

@misc{liu2026apsf,
title={Adaptive Prompt Structure Factorization: A Framework for Self-Discovering and Optimizing Compositional Prompt Programs},
author={Haoyue Liu and Zhichao Wang and Yongxin Guo and Haoran Shou and Xiaoying Tang},
year={2026},
eprint={2604.06699},
archivePrefix={arXiv},
primaryClass={cs.CL},
url={https://arxiv.org/abs/2604.06699}
}

@inproceedings{juneja2025taskfacet,
    title = "Task Facet Learning: A Structured Approach To Prompt Optimization",
    author = "Juneja, Gurusha  and
      Jajoo, Gautam  and
      Li, Hua  and
      Jiao, Jian  and
      Natarajan, Nagarajan  and
      Sharma, Amit",
    booktitle = "Findings of the Association for Computational Linguistics: ACL 2025",
    month = jul,
    year = "2025",
    address = "Vienna, Austria",
    publisher = "Association for Computational Linguistics",
    url = "https://aclanthology.org/2025.findings-acl.1206/",
    doi = "10.18653/v1/2025.findings-acl.1206",
    pages = "23473--23496",
}

@inproceedings{agrawal2025gepa,
 author = {Agrawal, Lakshya A and Tan, Shangyin and Soylu, Dilara and Ziems, Noah and Khare, Rishi and Opsahl-Ong, Krista and Singhvi, Arnav and Shandilya, Herumb and Ryan, Michael J and Jiang, Meng and Potts, Christopher and Sen, Koushik and Dimakis, Alex and Stoica, Ion and Klein, Dan and Zaharia, Matei and Khattab, Omar},
 booktitle = {International Conference on Learning Representations},
 pages = {8479--8565},
 title = {{GEPA}: Reflective Prompt Evolution Can Outperform Reinforcement Learning},
 url = {https://proceedings.iclr.cc/paper_files/paper/2026/file/0e9e708b6f48e14fd0ac29e167413f76-Paper-Conference.pdf},
 volume = {2026},
 year = {2026}
}

@inproceedings{schnabel2024sammo,
    title = "Symbolic Prompt Program Search: A Structure-Aware Approach to Efficient Compile-Time Prompt Optimization",
    author = "Schnabel, Tobias  and
      Neville, Jennifer",
    booktitle = "Findings of the Association for Computational Linguistics: EMNLP 2024",
    month = nov,
    year = "2024",
    address = "Miami, Florida, USA",
    publisher = "Association for Computational Linguistics",
    url = "https://aclanthology.org/2024.findings-emnlp.37/",
    doi = "10.18653/v1/2024.findings-emnlp.37",
    pages = "670--686",
}

@inproceedings{pryzant2023protegi,
    title = "Automatic Prompt Optimization with ``Gradient Descent'' and Beam Search",
    author = "Pryzant, Reid  and
      Iter, Dan  and
      Li, Jerry  and
      Lee, Yin  and
      Zhu, Chenguang  and
      Zeng, Michael",
    booktitle = "Proceedings of the 2023 Conference on Empirical Methods in Natural Language Processing",
    month = dec,
    year = "2023",
    address = "Singapore",
    publisher = "Association for Computational Linguistics",
    url = "https://aclanthology.org/2023.emnlp-main.494/",
    doi = "10.18653/v1/2023.emnlp-main.494",
    pages = "7957--7968",
}

@inproceedings{yang2024opro,
 author = {Yang, Chengrun and Wang, Xuezhi and Lu, Yifeng and Liu, Hanxiao and Le, Quoc V and Zhou, Denny and Chen, Xinyun},
 booktitle = {International Conference on Learning Representations},
 pages = {12028--12068},
 title = {Large Language Models as Optimizers},
 url = {https://proceedings.iclr.cc/paper_files/paper/2024/file/3339f19c5fcee3ad74502947a32be9e6-Paper-Conference.pdf},
 volume = {2024},
 year = {2024}
}

@article{yuksekgonul2024textgrad,
  title   = {Optimizing generative {AI} by backpropagating language model feedback},
  author  = {Yuksekgonul, Mert and Bianchi, Federico and Boen, Joseph and Liu, Sheng and Lu, Pan and Huang, Zhi and Guestrin, Carlos and Zou, James},
  journal = {Nature},
  volume  = {639},
  number  = {8055},
  pages   = {609--616},
  year    = {2025},
  month   = mar,
  doi     = {10.1038/s41586-025-08661-4},
  url     = {https://doi.org/10.1038/s41586-025-08661-4}
}

@misc{sharma2026modular,
  author        = {Sharma, Prith and Henley, Austin Z.},
  title         = {Modular Prompt Optimization: Optimizing Structured Prompts with Section-Local Textual Gradients},
  year          = {2026},
  eprint        = {2601.04055},
  archivePrefix = {arXiv},
  primaryClass  = {cs.CL},
  doi           = {10.48550/arXiv.2601.04055},
  url           = {https://arxiv.org/abs/2601.04055}
}

@misc{hapo2026,
      title={Learning from Prompt itself: the Hierarchical Attribution Prompt Optimization}, 
      author={Dongyu Chen and Jian Ma and Xianpeng Zhang and Lei Zhang and Haonan Lu and Chen Chen and Chuangchuang Wang and Kai Tang},
      year={2026},
      eprint={2601.02683},
      archivePrefix={arXiv},
      primaryClass={cs.AI},
      url={https://arxiv.org/abs/2601.02683}, 
}

@inproceedings{zhou2023ape,
title={Large Language Models are Human-Level Prompt Engineers},
author={Yongchao Zhou and Andrei Ioan Muresanu and Ziwen Han and Keiran Paster and Silviu Pitis and Harris Chan and Jimmy Ba},
booktitle={The Eleventh International Conference on Learning Representations },
year={2023},
url={https://openreview.net/forum?id=92gvk82DE-}
}

@inproceedings{khattab2024dspy,
 author = {Khattab, Omar and Singhvi, Arnav and Maheshwari, Paridhi and Zhang, Zhiyuan and Santhanam, Keshav and A, Sri Vardhamanan and Haq, Saiful and Sharma, Ashutosh and Joshi, Thomas and Moazam, Hanna and Miller, Heather and Zaharia, Matei and Potts, Christopher},
 booktitle = {International Conference on Learning Representations},
 pages = {54928--54958},
 title = {{DSPy}: Compiling Declarative Language Model Calls into State-of-the-Art Pipelines},
 url = {https://proceedings.iclr.cc/paper_files/paper/2024/file/f1cf02ce09757f57c3b93c0db83181e0-Paper-Conference.pdf},
 volume = {2024},
 year = {2024}
}

@inproceedings{opsahlong2024miprov2,
    title = "Optimizing Instructions and Demonstrations for Multi-Stage Language Model Programs",
    author = "Opsahl-Ong, Krista  and
      Ryan, Michael J  and
      Purtell, Josh  and
      Broman, David  and
      Potts, Christopher  and
      Zaharia, Matei  and
      Khattab, Omar",
    booktitle = "Proceedings of the 2024 Conference on Empirical Methods in Natural Language Processing",
    month = nov,
    year = "2024",
    address = "Miami, Florida, USA",
    publisher = "Association for Computational Linguistics",
    url = "https://aclanthology.org/2024.emnlp-main.525/",
    doi = "10.18653/v1/2024.emnlp-main.525",
    pages = "9340--9366",
}

@inproceedings{guo2024evoprompt,
 author = {Guo, Qingyan and Wang, Rui and Guo, Junliang and Li, Bei and Song, Kaitao and Tan, Xu and Liu, Guoqing and Bian, Jiang and Yang, Yujiu},
 booktitle = {International Conference on Learning Representations},
 pages = {34133--34156},
 title = {Connecting Large Language Models with Evolutionary Algorithms Yields Powerful Prompt Optimizers},
 url = {https://proceedings.iclr.cc/paper_files/paper/2024/file/9156b0f6dfa9bbd18c79cc459ef5d61c-Paper-Conference.pdf},
 volume = {2024},
 year = {2024}
}

@inproceedings{ye2024pe2,
    title = "Prompt Engineering a Prompt Engineer",
    author = "Ye, Qinyuan  and
      Ahmed, Mohamed  and
      Pryzant, Reid  and
      Khani, Fereshte",
    booktitle = "Findings of the Association for Computational Linguistics: ACL 2024",
    month = aug,
    year = "2024",
    address = "Bangkok, Thailand",
    publisher = "Association for Computational Linguistics",
    url = "https://aclanthology.org/2024.findings-acl.21/",
    doi = "10.18653/v1/2024.findings-acl.21",
    pages = "355--385",
}

@inproceedings{prasad2023grips,
    title = "{G}r{IPS}: Gradient-free, Edit-based Instruction Search for Prompting Large Language Models",
    author = "Prasad, Archiki  and
      Hase, Peter  and
      Zhou, Xiang  and
      Bansal, Mohit",
    booktitle = "Proceedings of the 17th Conference of the European Chapter of the Association for Computational Linguistics",
    month = may,
    year = "2023",
    address = "Dubrovnik, Croatia",
    publisher = "Association for Computational Linguistics",
    url = "https://aclanthology.org/2023.eacl-main.277/",
    doi = "10.18653/v1/2023.eacl-main.277",
    pages = "3845--3864",
}

@inproceedings{madaan2023selfrefine,
 author = {Madaan, Aman and Tandon, Niket and Gupta, Prakhar and Hallinan, Skyler and Gao, Luyu and Wiegreffe, Sarah and Alon, Uri and Dziri, Nouha and Prabhumoye, Shrimai and Yang, Yiming and Gupta, Shashank and Majumder, Bodhisattwa Prasad and Hermann, Katherine and Welleck, Sean and Yazdanbakhsh, Amir and Clark, Peter},
 booktitle = {Advances in Neural Information Processing Systems},
 doi = {10.52202/075280-2019},
 pages = {46534--46594},
 publisher = {Curran Associates, Inc.},
 title = {{Self-Refine}: Iterative Refinement with Self-Feedback},
 url = {https://proceedings.neurips.cc/paper_files/paper/2023/file/91edff07232fb1b55a505a9e9f6c0ff3-Paper-Conference.pdf},
 volume = {36},
 year = {2023}
}

@InProceedings{fernando2023promptbreeder,
  title = 	 {Promptbreeder: Self-Referential Self-Improvement via Prompt Evolution},
  author =       {Fernando, Chrisantha and Banarse, Dylan Sunil and Michalewski, Henryk and Osindero, Simon and Rockt\"{a}schel, Tim},
  booktitle = 	 {Proceedings of the 41st International Conference on Machine Learning},
  pages = 	 {13481--13544},
  year = 	 {2024},
  volume = 	 {235},
  series = 	 {Proceedings of Machine Learning Research},
  month = 	 {21--27 Jul},
  publisher =    {PMLR},
  url = 	 {https://proceedings.mlr.press/v235/fernando24a.html},
}

@inproceedings{he2024crispo,
  title={{CriSPO}: Multi-aspect critique-suggestion-guided automatic prompt optimization for text generation},
  author={He, Han and Liu, Qianchu and Xu, Lei and Shivade, Chaitanya and Zhang, Yi and Srinivasan, Sundararajan and Kirchhoff, Katrin},
  booktitle={Proceedings of the AAAI Conference on Artificial Intelligence},
  volume={39},
  pages={24014--24022},
  doi = {10.1609/aaai.v39i22.34575},
  year={2025}
}

@inproceedings{yi2025zera,
    title = "{ZERA}: Zero-init Instruction Evolving Refinement Agent {--} From Zero Instructions to Structured Prompts via Principle-based Optimization",
    author = "Yi, Seungyoun  and
      Khang, Minsoo  and
      Park, Sungrae",
    booktitle = "Proceedings of the 2025 Conference on Empirical Methods in Natural Language Processing",
    month = nov,
    year = "2025",
    address = "Suzhou, China",
    publisher = "Association for Computational Linguistics",
    url = "https://aclanthology.org/2025.emnlp-main.1190/",
    doi = "10.18653/v1/2025.emnlp-main.1190",
    pages = "23323--23337",
}

@inproceedings{spector2010assessment,
  title={Assessment of problem modality by differential performance of lexicase selection in genetic programming: a preliminary report},
  author={Spector, Lee},
  booktitle={Proceedings of the 14th annual conference companion on Genetic and evolutionary computation},
  pages={401--408},
  doi = {10.1145/2330784.2330846},
  year={2012}
}

@misc{grattafiori2024llama,
      title={The {Llama} 3 Herd of Models}, 
      author={Aaron Grattafiori and Abhimanyu Dubey and Abhinav Jauhri and Abhinav Pandey and Abhishek Kadian and Ahmad Al-Dahle and Aiesha Letman and Akhil Mathur and Alan Schelten and Alex Vaughan and Amy Yang and Angela Fan and Anirudh Goyal and Anthony Hartshorn and Aobo Yang and Archi Mitra and Archie Sravankumar and Artem Korenev and Arthur Hinsvark and Arun Rao and Aston Zhang and Aurelien Rodriguez and Austen Gregerson and Ava Spataru and Baptiste Roziere and Bethany Biron and Binh Tang and Bobbie Chern and Charlotte Caucheteux and Chaya Nayak and Chloe Bi and Chris Marra and Chris McConnell and Christian Keller and Christophe Touret and Chunyang Wu and Corinne Wong and Cristian Canton Ferrer and Cyrus Nikolaidis and Damien Allonsius and Daniel Song and Danielle Pintz and Danny Livshits and Danny Wyatt and David Esiobu and Dhruv Choudhary and Dhruv Mahajan and Diego Garcia-Olano and Diego Perino and Dieuwke Hupkes and Egor Lakomkin and Ehab AlBadawy and Elina Lobanova and Emily Dinan and Eric Michael Smith and Filip Radenovic and Francisco Guzmán and Frank Zhang and Gabriel Synnaeve and Gabrielle Lee and Georgia Lewis Anderson and Govind Thattai and Graeme Nail and Gregoire Mialon and Guan Pang and Guillem Cucurell and Hailey Nguyen and Hannah Korevaar and Hu Xu and Hugo Touvron and Iliyan Zarov and Imanol Arrieta Ibarra and Isabel Kloumann and Ishan Misra and Ivan Evtimov and Jack Zhang and Jade Copet and Jaewon Lee and Jan Geffert and Jana Vranes and Jason Park and Jay Mahadeokar and Jeet Shah and Jelmer van der Linde and Jennifer Billock and Jenny Hong and Jenya Lee and Jeremy Fu and Jianfeng Chi and Jianyu Huang and Jiawen Liu and Jie Wang and Jiecao Yu and Joanna Bitton and Joe Spisak and Jongsoo Park and Joseph Rocca and Joshua Johnstun and Joshua Saxe and Junteng Jia and Kalyan Vasuden Alwala and Karthik Prasad and Kartikeya Upasani and Kate Plawiak and Ke Li and Kenneth Heafield and Kevin Stone and Khalid El-Arini and Krithika Iyer and Kshitiz Malik and Kuenley Chiu and Kunal Bhalla and Kushal Lakhotia and Lauren Rantala-Yeary and Laurens van der Maaten and Lawrence Chen and Liang Tan and Liz Jenkins and Louis Martin and Lovish Madaan and Lubo Malo and Lukas Blecher and Lukas Landzaat and Luke de Oliveira and Madeline Muzzi and Mahesh Pasupuleti and Mannat Singh and Manohar Paluri and Marcin Kardas and Maria Tsimpoukelli and Mathew Oldham and Mathieu Rita and Maya Pavlova and Melanie Kambadur and Mike Lewis and Min Si and Mitesh Kumar Singh and Mona Hassan and Naman Goyal and Narjes Torabi and Nikolay Bashlykov and Nikolay Bogoychev and Niladri Chatterji and Ning Zhang and Olivier Duchenne and Onur Çelebi and Patrick Alrassy and Pengchuan Zhang and Pengwei Li and Petar Vasic and Peter Weng and Prajjwal Bhargava and Pratik Dubal and Praveen Krishnan and Punit Singh Koura and Puxin Xu and Qing He and Qingxiao Dong and Ragavan Srinivasan and Raj Ganapathy and Ramon Calderer and Ricardo Silveira Cabral and Robert Stojnic and Roberta Raileanu and Rohan Maheswari and Rohit Girdhar and Rohit Patel and Romain Sauvestre and Ronnie Polidoro and Roshan Sumbaly and Ross Taylor and Ruan Silva and Rui Hou and Rui Wang and Saghar Hosseini and Sahana Chennabasappa and Sanjay Singh and Sean Bell and Seohyun Sonia Kim and Sergey Edunov and Shaoliang Nie and Sharan Narang and Sharath Raparthy and Sheng Shen and Shengye Wan and Shruti Bhosale and Shun Zhang and Simon Vandenhende and Soumya Batra and Spencer Whitman and Sten Sootla and Stephane Collot and Suchin Gururangan and Sydney Borodinsky and Tamar Herman and Tara Fowler and Tarek Sheasha and Thomas Georgiou and Thomas Scialom and Tobias Speckbacher and Todor Mihaylov and Tong Xiao and Ujjwal Karn and Vedanuj Goswami and Vibhor Gupta and Vignesh Ramanathan and Viktor Kerkez and Vincent Gonguet and Virginie Do and Vish Vogeti and Vítor Albiero and Vladan Petrovic and Weiwei Chu and Wenhan Xiong and Wenyin Fu and Whitney Meers and Xavier Martinet and Xiaodong Wang and Xiaofang Wang and Xiaoqing Ellen Tan and Xide Xia and Xinfeng Xie and Xuchao Jia and Xuewei Wang and Yaelle Goldschlag and Yashesh Gaur and Yasmine Babaei and Yi Wen and Yiwen Song and Yuchen Zhang and Yue Li and Yuning Mao and Zacharie Delpierre Coudert and Zheng Yan and Zhengxing Chen and Zoe Papakipos and Aaditya Singh and Aayushi Srivastava and Abha Jain and Adam Kelsey and Adam Shajnfeld and Adithya Gangidi and Adolfo Victoria and Ahuva Goldstand and Ajay Menon and Ajay Sharma and Alex Boesenberg and Alexei Baevski and Allie Feinstein and Amanda Kallet and Amit Sangani and Amos Teo and Anam Yunus and Andrei Lupu and Andres Alvarado and Andrew Caples and Andrew Gu and Andrew Ho and Andrew Poulton and Andrew Ryan and Ankit Ramchandani and Annie Dong and Annie Franco and Anuj Goyal and Aparajita Saraf and Arkabandhu Chowdhury and Ashley Gabriel and Ashwin Bharambe and Assaf Eisenman and Azadeh Yazdan and Beau James and Ben Maurer and Benjamin Leonhardi and Bernie Huang and Beth Loyd and Beto De Paola and Bhargavi Paranjape and Bing Liu and Bo Wu and Boyu Ni and Braden Hancock and Bram Wasti and Brandon Spence and Brani Stojkovic and Brian Gamido and Britt Montalvo and Carl Parker and Carly Burton and Catalina Mejia and Ce Liu and Changhan Wang and Changkyu Kim and Chao Zhou and Chester Hu and Ching-Hsiang Chu and Chris Cai and Chris Tindal and Christoph Feichtenhofer and Cynthia Gao and Damon Civin and Dana Beaty and Daniel Kreymer and Daniel Li and David Adkins and David Xu and Davide Testuggine and Delia David and Devi Parikh and Diana Liskovich and Didem Foss and Dingkang Wang and Duc Le and Dustin Holland and Edward Dowling and Eissa Jamil and Elaine Montgomery and Eleonora Presani and Emily Hahn and Emily Wood and Eric-Tuan Le and Erik Brinkman and Esteban Arcaute and Evan Dunbar and Evan Smothers and Fei Sun and Felix Kreuk and Feng Tian and Filippos Kokkinos and Firat Ozgenel and Francesco Caggioni and Frank Kanayet and Frank Seide and Gabriela Medina Florez and Gabriella Schwarz and Gada Badeer and Georgia Swee and Gil Halpern and Grant Herman and Grigory Sizov and Guangyi and Zhang and Guna Lakshminarayanan and Hakan Inan and Hamid Shojanazeri and Han Zou and Hannah Wang and Hanwen Zha and Haroun Habeeb and Harrison Rudolph and Helen Suk and Henry Aspegren and Hunter Goldman and Hongyuan Zhan and Ibrahim Damlaj and Igor Molybog and Igor Tufanov and Ilias Leontiadis and Irina-Elena Veliche and Itai Gat and Jake Weissman and James Geboski and James Kohli and Janice Lam and Japhet Asher and Jean-Baptiste Gaya and Jeff Marcus and Jeff Tang and Jennifer Chan and Jenny Zhen and Jeremy Reizenstein and Jeremy Teboul and Jessica Zhong and Jian Jin and Jingyi Yang and Joe Cummings and Jon Carvill and Jon Shepard and Jonathan McPhie and Jonathan Torres and Josh Ginsburg and Junjie Wang and Kai Wu and Kam Hou U and Karan Saxena and Kartikay Khandelwal and Katayoun Zand and Kathy Matosich and Kaushik Veeraraghavan and Kelly Michelena and Keqian Li and Kiran Jagadeesh and Kun Huang and Kunal Chawla and Kyle Huang and Lailin Chen and Lakshya Garg and Lavender A and Leandro Silva and Lee Bell and Lei Zhang and Liangpeng Guo and Licheng Yu and Liron Moshkovich and Luca Wehrstedt and Madian Khabsa and Manav Avalani and Manish Bhatt and Martynas Mankus and Matan Hasson and Matthew Lennie and Matthias Reso and Maxim Groshev and Maxim Naumov and Maya Lathi and Meghan Keneally and Miao Liu and Michael L. Seltzer and Michal Valko and Michelle Restrepo and Mihir Patel and Mik Vyatskov and Mikayel Samvelyan and Mike Clark and Mike Macey and Mike Wang and Miquel Jubert Hermoso and Mo Metanat and Mohammad Rastegari and Munish Bansal and Nandhini Santhanam and Natascha Parks and Natasha White and Navyata Bawa and Nayan Singhal and Nick Egebo and Nicolas Usunier and Nikhil Mehta and Nikolay Pavlovich Laptev and Ning Dong and Norman Cheng and Oleg Chernoguz and Olivia Hart and Omkar Salpekar and Ozlem Kalinli and Parkin Kent and Parth Parekh and Paul Saab and Pavan Balaji and Pedro Rittner and Philip Bontrager and Pierre Roux and Piotr Dollar and Polina Zvyagina and Prashant Ratanchandani and Pritish Yuvraj and Qian Liang and Rachad Alao and Rachel Rodriguez and Rafi Ayub and Raghotham Murthy and Raghu Nayani and Rahul Mitra and Rangaprabhu Parthasarathy and Raymond Li and Rebekkah Hogan and Robin Battey and Rocky Wang and Russ Howes and Ruty Rinott and Sachin Mehta and Sachin Siby and Sai Jayesh Bondu and Samyak Datta and Sara Chugh and Sara Hunt and Sargun Dhillon and Sasha Sidorov and Satadru Pan and Saurabh Mahajan and Saurabh Verma and Seiji Yamamoto and Sharadh Ramaswamy and Shaun Lindsay and Shaun Lindsay and Sheng Feng and Shenghao Lin and Shengxin Cindy Zha and Shishir Patil and Shiva Shankar and Shuqiang Zhang and Shuqiang Zhang and Sinong Wang and Sneha Agarwal and Soji Sajuyigbe and Soumith Chintala and Stephanie Max and Stephen Chen and Steve Kehoe and Steve Satterfield and Sudarshan Govindaprasad and Sumit Gupta and Summer Deng and Sungmin Cho and Sunny Virk and Suraj Subramanian and Sy Choudhury and Sydney Goldman and Tal Remez and Tamar Glaser and Tamara Best and Thilo Koehler and Thomas Robinson and Tianhe Li and Tianjun Zhang and Tim Matthews and Timothy Chou and Tzook Shaked and Varun Vontimitta and Victoria Ajayi and Victoria Montanez and Vijai Mohan and Vinay Satish Kumar and Vishal Mangla and Vlad Ionescu and Vlad Poenaru and Vlad Tiberiu Mihailescu and Vladimir Ivanov and Wei Li and Wenchen Wang and Wenwen Jiang and Wes Bouaziz and Will Constable and Xiaocheng Tang and Xiaojian Wu and Xiaolan Wang and Xilun Wu and Xinbo Gao and Yaniv Kleinman and Yanjun Chen and Ye Hu and Ye Jia and Ye Qi and Yenda Li and Yilin Zhang and Ying Zhang and Yossi Adi and Youngjin Nam and Yu and Wang and Yu Zhao and Yuchen Hao and Yundi Qian and Yunlu Li and Yuzi He and Zach Rait and Zachary DeVito and Zef Rosnbrick and Zhaoduo Wen and Zhenyu Yang and Zhiwei Zhao and Zhiyu Ma},
      year={2024},
      eprint={2407.21783},
      archivePrefix={arXiv},
      primaryClass={cs.AI},
      url={https://arxiv.org/abs/2407.21783}, 
}

@misc{yang2025qwen3,
      title={Qwen3 Technical Report}, 
      author={An Yang and Anfeng Li and Baosong Yang and Beichen Zhang and Binyuan Hui and Bo Zheng and Bowen Yu and Chang Gao and Chengen Huang and Chenxu Lv and Chujie Zheng and Dayiheng Liu and Fan Zhou and Fei Huang and Feng Hu and Hao Ge and Haoran Wei and Huan Lin and Jialong Tang and Jian Yang and Jianhong Tu and Jianwei Zhang and Jianxin Yang and Jiaxi Yang and Jing Zhou and Jingren Zhou and Junyang Lin and Kai Dang and Keqin Bao and Kexin Yang and Le Yu and Lianghao Deng and Mei Li and Mingfeng Xue and Mingze Li and Pei Zhang and Peng Wang and Qin Zhu and Rui Men and Ruize Gao and Shixuan Liu and Shuang Luo and Tianhao Li and Tianyi Tang and Wenbiao Yin and Xingzhang Ren and Xinyu Wang and Xinyu Zhang and Xuancheng Ren and Yang Fan and Yang Su and Yichang Zhang and Yinger Zhang and Yu Wan and Yuqiong Liu and Zekun Wang and Zeyu Cui and Zhenru Zhang and Zhipeng Zhou and Zihan Qiu},
      year={2025},
      eprint={2505.09388},
      archivePrefix={arXiv},
      primaryClass={cs.CL},
      url={https://arxiv.org/abs/2505.09388}, 
}

@misc{qwen2026qwen35modelcard,
    title  = {{Qwen3.5}: Towards Native Multimodal Agents},
    author = {{Qwen Team}},
    month  = {February},
    year   = {2026},
    url    = {https://qwen.ai/blog?id=qwen3.5}
}

@article{efraimidis2006weighted,
    title = {Weighted random sampling with a reservoir},
    journal = {Information Processing Letters},
    volume = {97},
    number = {5},
    pages = {181-185},
    year = {2006},
    issn = {0020-0190},
    doi = {https://doi.org/10.1016/j.ipl.2005.11.003},
    url = {https://www.sciencedirect.com/science/article/pii/S002001900500298X},
    author = {Pavlos S. Efraimidis and Paul G. Spirakis}
}

@inproceedings{suzgun2023bbh,
    title = "Challenging {{BIG}-Bench} Tasks and Whether Chain-of-Thought Can Solve Them",
    author = {Suzgun, Mirac  and
      Scales, Nathan  and
      Sch{\"a}rli, Nathanael  and
      Gehrmann, Sebastian  and
      Tay, Yi  and
      Chung, Hyung Won  and
      Chowdhery, Aakanksha  and
      Le, Quoc  and
      Chi, Ed  and
      Zhou, Denny  and
      Wei, Jason},
    booktitle = "Findings of the Association for Computational Linguistics: ACL 2023",
    month = jul,
    year = "2023",
    address = "Toronto, Canada",
    publisher = "Association for Computational Linguistics",
    url = "https://aclanthology.org/2023.findings-acl.824/",
    doi = "10.18653/v1/2023.findings-acl.824",
    pages = "13003--13051",
}

@inproceedings{wang2024mmlupro,
 author = {Wang, Yubo and Ma, Xueguang and Zhang, Ge and Ni, Yuansheng and Chandra, Abhranil and Guo, Shiguang and Ren, Weiming and Arulraj, Aaran and He, Xuan and Jiang, Ziyan and Li, Tianle and Ku, Max and Wang, Kai and Zhuang, Alex and Fan, Rongqi and Yue, Xiang and Chen, Wenhu},
 booktitle = {Advances in Neural Information Processing Systems},
 doi = {10.52202/079017-3018},
 pages = {95266--95290},
 publisher = {Curran Associates, Inc.},
 title = {{MMLU-Pro}: A More Robust and Challenging Multi-Task Language Understanding Benchmark},
 url = {https://proceedings.neurips.cc/paper_files/paper/2024/file/ad236edc564f3e3156e1b2feafb99a24-Paper-Datasets_and_Benchmarks_Track.pdf},
 volume = {37},
 year = {2024}
}

@inproceedings{liu2023evalplus,
 author = {Liu, Jiawei and Xia, Chunqiu Steven and Wang, Yuyao and Zhang, Lingming},
 booktitle = {Advances in Neural Information Processing Systems},
 doi = {10.52202/075280-0943},
 pages = {21558--21572},
 publisher = {Curran Associates, Inc.},
 title = {Is Your Code Generated by {ChatGPT} Really Correct? Rigorous Evaluation of Large Language Models for Code Generation},
 volume = {36},
 year = {2023}
}

@InProceedings{gao2023pal,
  title = 	 {{PAL}: Program-aided Language Models},
  author =       {Gao, Luyu and Madaan, Aman and Zhou, Shuyan and Alon, Uri and Liu, Pengfei and Yang, Yiming and Callan, Jamie and Neubig, Graham},
  booktitle = 	 {Proceedings of the 40th International Conference on Machine Learning},
  pages = 	 {10764--10799},
  year = 	 {2023},
  volume = 	 {202},
  series = 	 {Proceedings of Machine Learning Research},
  month = 	 {23--29 Jul},
  publisher =    {PMLR},
  url = 	 {https://proceedings.mlr.press/v202/gao23f.html},
}

@inproceedings{yang2018hotpotqa,
    title = "{H}otpot{QA}: A Dataset for Diverse, Explainable Multi-hop Question Answering",
    author = "Yang, Zhilin  and
      Qi, Peng  and
      Zhang, Saizheng  and
      Bengio, Yoshua  and
      Cohen, William  and
      Salakhutdinov, Ruslan  and
      Manning, Christopher D.",
    booktitle = "Proceedings of the 2018 Conference on Empirical Methods in Natural Language Processing",
    month = oct # "-" # nov,
    year = "2018",
    address = "Brussels, Belgium",
    publisher = "Association for Computational Linguistics",
    url = "https://aclanthology.org/D18-1259/",
    doi = "10.18653/v1/D18-1259",
    pages = "2369--2380",
}

@inproceedings{ramnath2025systematic,
  title = {A Systematic Survey of Automatic Prompt Optimization Techniques},
  author = {Ramnath, Kiran and Zhou, Kang and Guan, Sheng and
            Mishra, Soumya Smruti and Qi, Xuan and Shen, Zhengyuan and
            Wang, Shuai and Woo, Sangmin and Jeoung, Sullam and Wang, Yawei and
            Wang, Haozhu and Ding, Han and Lu, Yuzhe and Xu, Zhichao and
            Zhou, Yun and Srinivasan, Balasubramaniam and Yan, Qiaojing and
            Chen, Yueyan and Ding, Haibo and Xu, Panpan and Cheong, Lin Lee},
  booktitle = {Proceedings of the 2025 Conference on Empirical Methods in Natural Language Processing},
  pages = {33078--33110},
  year = {2025},
  url = {https://aclanthology.org/2025.emnlp-main.1681/},
  doi = {10.18653/v1/2025.emnlp-main.1681}
}

@inproceedings{jain2025local,
    title = "Local Prompt Optimization",
    author = "Jain, Yash  and
      Chowdhary, Vishal",
    booktitle = "Proceedings of the 2025 Conference of the Nations of the Americas Chapter of the Association for Computational Linguistics: Human Language Technologies (Volume 2: Short Papers)",
    month = apr,
    year = "2025",
    address = "Albuquerque, New Mexico",
    publisher = "Association for Computational Linguistics",
    url = "https://aclanthology.org/2025.naacl-short.7/",
    doi = "10.18653/v1/2025.naacl-short.7",
    pages = "75--81",
}

@inproceedings{kumar2025sculpt,
    title = "{SCULPT}: Systematic Tuning of Long Prompts",
    author = "Kumar, Shanu  and
      Venkata, Akhila Yesantarao  and
      Khandelwal, Shubhanshu  and
      Santra, Bishal  and
      Agrawal, Parag  and
      Gupta, Manish",
    booktitle = "Proceedings of the 63rd Annual Meeting of the Association for Computational Linguistics (Volume 1: Long Papers)",
    month = jul,
    year = "2025",
    address = "Vienna, Austria",
    publisher = "Association for Computational Linguistics",
    url = "https://aclanthology.org/2025.acl-long.730/",
    doi = "10.18653/v1/2025.acl-long.730",
    pages = "14996--15029",
}

@inproceedings{xia2026hivemind,
  title={{HiveMind}: Contribution-guided online prompt optimization of {LLM} multi-agent systems},
  author={Xia, Yihan and Wang, Taotao and Zhang, Shengli and Weng, Zhangyuhua and Cao, Bin and Liew, Soung Chang},
  booktitle={Proceedings of the AAAI Conference on Artificial Intelligence},
  volume={40},
  doi = "10.1609/aaai.v40i35.40222",
  pages={29767--29774},
  year={2026}
}

@inproceedings{agarwal2025promptwizard,
  title={{P}rompt{W}izard: Optimizing Prompts via Task-Aware, Feedback-Driven Self-Evolution},
  author={Agarwal, Eshaan and Magazine, Raghav and Singh, Joykirat and Dani, Vivek and Ganu, Tanuja and Nambi, Akshay},
  booktitle={Findings of the Association for Computational Linguistics: ACL 2025},
  url = "https://aclanthology.org/2025.findings-acl.1025/",
  doi = "10.18653/v1/2025.findings-acl.1025",
  pages={19974--20003},
  year={2025}
}

@inproceedings{shinn2023reflexion,
 author = {Shinn, Noah and Cassano, Federico and Gopinath, Ashwin and Narasimhan, Karthik and Yao, Shunyu},
 booktitle = {Advances in Neural Information Processing Systems},
 doi = {10.52202/075280-0377},
 pages = {8634--8652},
 publisher = {Curran Associates, Inc.},
 title = {Reflexion: language agents with verbal reinforcement learning},
 url = {https://proceedings.neurips.cc/paper_files/paper/2023/file/1b44b878bb782e6954cd888628510e90-Paper-Conference.pdf},
 volume = {36},
 year = {2023}
}

@inproceedings{chen2024promst,
    title = "{PR}ompt Optimization in Multi-Step Tasks ({PROMST}): Integrating Human Feedback and Heuristic-based Sampling",
    author = "Chen, Yongchao  and
      Arkin, Jacob  and
      Hao, Yilun  and
      Zhang, Yang  and
      Roy, Nicholas  and
      Fan, Chuchu",
    booktitle = "Proceedings of the 2024 Conference on Empirical Methods in Natural Language Processing",
    month = nov,
    year = "2024",
    address = "Miami, Florida, USA",
    publisher = "Association for Computational Linguistics",
    url = "https://aclanthology.org/2024.emnlp-main.226/",
    doi = "10.18653/v1/2024.emnlp-main.226",
    pages = "3859--3920",
}

@article{srivastava2023beyond,
  title     = {{Beyond the Imitation Game: Quantifying and Extrapolating the Capabilities of Language Models}},
  author    = {Srivastava, Aarohi and Rastogi, Abhinav and Rao, Abhishek and Shoeb, Abu Awal Md and Abid, Abubakar and Fisch, Adam and Brown, Adam R. and Santoro, Adam and Gupta, Aditya and Garriga-Alonso, Adrià and Kluska, Agnieszka and Lewkowycz, Aitor and Agarwal, Akshat and Power, Alethea and Ray, Alex and Warstadt, Alex and Kocurek, Alexander W. and Safaya, Ali and Tazarv, Ali and Xiang, Alice and Parrish, Alicia and Nie, Allen and Hussain, Aman and Askell, Amanda and Dsouza, Amanda and Slone, Ambrose and Rahane, Ameet and Iyer, Anantharaman S. and Andreassen, Anders Johan and Madotto, Andrea and Santilli, Andrea and Stuhlmüller, Andreas and Dai, Andrew M. and La, Andrew and Lampinen, Andrew Kyle and Zou, Andy and Jiang, Angela and Chen, Angelica and Vuong, Anh and Gupta, Animesh and Gottardi, Anna and Norelli, Antonio and Venkatesh, Anu and Gholamidavoodi, Arash and Tabassum, Arfa and Menezes, Arul and Kirubarajan, Arun and Mullokandov, Asher and Sabharwal, Ashish and Herrick, Austin and Efrat, Avia and Erdem, Aykut and Karakaş, Ayla and Roberts, B. Ryan and Loe, Bao Sheng and Zoph, Barret and Bojanowski, Bartłomiej and Özyurt, Batuhan and Hedayatnia, Behnam and Neyshabur, Behnam and Inden, Benjamin and Stein, Benno and Ekmekci, Berk and Lin, Bill Yuchen and Howald, Blake and Orinion, Bryan and Diao, Cameron and Dour, Cameron and Stinson, Catherine and Argueta, Cedrick and Ferri, Cesar and Singh, Chandan and Rathkopf, Charles and Meng, Chenlin and Baral, Chitta and Wu, Chiyu and Callison-Burch, Chris and Waites, Christopher and Voigt, Christian and Manning, Christopher D and Potts, Christopher and Ramirez, Cindy and Rivera, Clara E. and Siro, Clemencia and Raffel, Colin and Ashcraft, Courtney and Garbacea, Cristina and Sileo, Damien and Garrette, Dan and Hendrycks, Dan and Kilman, Dan and Roth, Dan and Freeman, C. Daniel and Khashabi, Daniel and Levy, Daniel and González, Daniel Moseguí and Perszyk, Danielle and Hernandez, Danny and Chen, Danqi and Ippolito, Daphne and Gilboa, Dar and Dohan, David and Drakard, David and Jurgens, David and Datta, Debajyoti and Ganguli, Deep and Emelin, Denis and Kleyko, Denis and Yuret, Deniz and Chen, Derek and Tam, Derek and Hupkes, Dieuwke and Misra, Diganta and Buzan, Dilyar and Mollo, Dimitri Coelho and Yang, Diyi and Lee, Dong-Ho and Schrader, Dylan and Shutova, Ekaterina and Cubuk, Ekin Dogus and Segal, Elad and Hagerman, Eleanor and Barnes, Elizabeth and Donoway, Elizabeth and Pavlick, Ellie and Rodolà, Emanuele and Lam, Emma and Chu, Eric and Tang, Eric and Erdem, Erkut and Chang, Ernie and Chi, Ethan A and Dyer, Ethan and Jerzak, Ethan and Kim, Ethan and Manyasi, Eunice Engefu and Zheltonozhskii, Evgenii and Xia, Fanyue and Siar, Fatemeh and Martínez-Plumed, Fernando and Happé, Francesca and Chollet, Francois and Rong, Frieda and Mishra, Gaurav and Winata, Genta Indra and de Melo, Gerard and Kruszewski, Germàn and Parascandolo, Giambattista and Mariani, Giorgio and Wang, Gloria Xinyue and Jaimovitch-Lopez, Gonzalo and Betz, Gregor and Gur-Ari, Guy and Galijasevic, Hana and Kim, Hannah and Rashkin, Hannah and Hajishirzi, Hannaneh and Mehta, Harsh and Bogar, Hayden and Shevlin, Henry Francis Anthony and Schuetze, Hinrich and Yakura, Hiromu and Zhang, Hongming and Wong, Hugh Mee and Ng, Ian and Noble, Isaac and Jumelet, Jaap and Geissinger, Jack and Kernion, Jackson and Hilton, Jacob and Lee, Jaehoon and Fisac, Jaime Fernández and Simon, James B and Koppel, James and Zheng, James and Zou, James and Kocon, Jan and Thompson, Jana and Wingfield, Janelle and Kaplan, Jared and Radom, Jarema and Sohl-Dickstein, Jascha and Phang, Jason and Wei, Jason and Yosinski, Jason and Novikova, Jekaterina and Bosscher, Jelle and Marsh, Jennifer and Kim, Jeremy and Taal, Jeroen and Engel, Jesse and Alabi, Jesujoba and Xu, Jiacheng and Song, Jiaming and Tang, Jillian and Waweru, Joan and Burden, John and Miller, John and Balis, John U. and Batchelder, Jonathan and Berant, Jonathan and Frohberg, Jörg and Rozen, Jos and Hernandez-Orallo, Jose and Boudeman, Joseph and Guerr, Joseph and Jones, Joseph and Tenenbaum, Joshua B. and Rule, Joshua S. and Chua, Joyce and Kanclerz, Kamil and Livescu, Karen and Krauth, Karl and Gopalakrishnan, Karthik and Ignatyeva, Katerina and Markert, Katja and Dhole, Kaustubh and Gimpel, Kevin and Omondi, Kevin and Mathewson, Kory Wallace and Chiafullo, Kristen and Shkaruta, Ksenia and Shridhar, Kumar and McDonell, Kyle and Richardson, Kyle and Reynolds, Laria and Gao, Leo and Zhang, Li and Dugan, Liam and Qin, Lianhui and Contreras-Ochando, Lidia and Morency, Louis-Philippe and Moschella, Luca and Lam, Lucas and Noble, Lucy and Schmidt, Ludwig and He, Luheng and Oliveros-Colón, Luis and Metz, Luke and Senel, Lütfi Kerem and Bosma, Maarten and Sap, Maarten and Ter Hoeve, Maartje and Farooqi, Maheen and Faruqui, Manaal and Mazeika, Mantas and Baturan, Marco and Marelli, Marco and Maru, Marco and Ramirez-Quintana, Maria Jose and Tolkiehn, Marie and Giulianelli, Mario and Lewis, Martha and Potthast, Martin and Leavitt, Matthew L and Hagen, Matthias and Schubert, Mátyás and Baitemirova, Medina Orduna and Arnaud, Melody and McElrath, Melvin and Yee, Michael Andrew and Cohen, Michael and Gu, Michael and Ivanitskiy, Michael and Starritt, Michael and Strube, Michael and Swędrowski, Michał and Bevilacqua, Michele and Yasunaga, Michihiro and Kale, Mihir and Cain, Mike and Xu, Mimee and Suzgun, Mirac and Walker, Mitch and Tiwari, Mo and Bansal, Mohit and Aminnaseri, Moin and Geva, Mor and Gheini, Mozhdeh and T, Mukund Varma and Peng, Nanyun and Chi, Nathan Andrew and Lee, Nayeon and Krakover, Neta Gur-Ari and Cameron, Nicholas and Roberts, Nicholas and Doiron, Nick and Martinez, Nicole and Nangia, Nikita and Deckers, Niklas and Muennighoff, Niklas and Keskar, Nitish Shirish and Iyer, Niveditha S. and Constant, Noah and Fiedel, Noah and Wen, Nuan and Zhang, Oliver and Agha, Omar and Elbaghdadi, Omar and Levy, Omer and Evans, Owain and Casares, Pablo Antonio Moreno and Doshi, Parth and Fung, Pascale and Liang, Paul Pu and Vicol, Paul and Alipoormolabashi, Pegah and Liao, Peiyuan and Liang, Percy and Chang, Peter W and Eckersley, Peter and Htut, Phu Mon and Hwang, Pinyu and Miłkowski, Piotr and Patil, Piyush and Pezeshkpour, Pouya and Oli, Priti and Mei, Qiaozhu and Lyu, Qing and Chen, Qinlang and Banjade, Rabin and Rudolph, Rachel Etta and Gabriel, Raefer and Habacker, Rahel and Risco, Ramon and Millière, Raphaël and Garg, Rhythm and Barnes, Richard and Saurous, Rif A. and Arakawa, Riku and Raymaekers, Robbe and Frank, Robert and Sikand, Rohan and Novak, Roman and Sitelew, Roman and Le Bras, Ronan and Liu, Rosanne and Jacobs, Rowan and Zhang, Rui and Salakhutdinov, Russ and Chi, Ryan Andrew and Lee, Seungjae Ryan and Stovall, Ryan and Teehan, Ryan and Yang, Rylan and Singh, Sahib and Mohammad, Saif M. and Anand, Sajant and Dillavou, Sam and Shleifer, Sam and Wiseman, Sam and Gruetter, Samuel and Bowman, Samuel R. and Schoenholz, Samuel Stern and Han, Sanghyun and Kwatra, Sanjeev and Rous, Sarah A. and Ghazarian, Sarik and Ghosh, Sayan and Casey, Sean and Bischoff, Sebastian and Gehrmann, Sebastian and Schuster, Sebastian and Sadeghi, Sepideh and Hamdan, Shadi and Zhou, Sharon and Srivastava, Shashank and Shi, Sherry and Singh, Shikhar and Asaadi, Shima and Gu, Shixiang Shane and Pachchigar, Shubh and Toshniwal, Shubham and Upadhyay, Shyam and Debnath, Shyamolima Shammie and Shakeri, Siamak and Thormeyer, Simon and Melzi, Simone and Reddy, Siva and Makini, Sneha Priscilla and Lee, Soo-Hwan and Torene, Spencer and Hatwar, Sriharsha and Dehaene, Stanislas and Divic, Stefan and Ermon, Stefano and Biderman, Stella and Lin, Stephanie and Prasad, Stephen and Piantadosi, Steven and Shieber, Stuart and Misherghi, Summer and Kiritchenko, Svetlana and Mishra, Swaroop and Linzen, Tal and Schuster, Tal and Li, Tao and Yu, Tao and Ali, Tariq and Hashimoto, Tatsunori and Wu, Te-Lin and Desbordes, Théo and Rothschild, Theodore and Phan, Thomas and Wang, Tianle and Nkinyili, Tiberius and Schick, Timo and Kornev, Timofei and Tunduny, Titus and Gerstenberg, Tobias and Chang, Trenton and Neeraj, Trishala and Khot, Tushar and Shultz, Tyler and Shaham, Uri and Misra, Vedant and Demberg, Vera and Nyamai, Victoria and Raunak, Vikas and Ramasesh, Vinay Venkatesh and Prabhu, Vinay Uday and Padmakumar, Vishakh and Srikumar, Vivek and Fedus, William and Saunders, William and Zhang, William and Vossen, Wout and Ren, Xiang and Tong, Xiaoyu and Zhao, Xinran and Wu, Xinyi and Shen, Xudong and Yaghoobzadeh, Yadollah and Lakretz, Yair and Song, Yangqiu and Bahri, Yasaman and Choi, Yejin and Yang, Yichi and Hao, Sophie and Chen, Yifu and Belinkov, Yonatan and Hou, Yu and Hou, Yufang and Bai, Yuntao and Seid, Zachary and Zhao, Zhuoye and Wang, Zijian and Wang, Zijie J. and Wang, Zirui and Wu, Ziyi},
  journal   = {Transactions on Machine Learning Research},
  year      = {2023},
  url       = {https://mlanthology.org/tmlr/2023/srivastava2023tmlr-beyond/}
}

@misc{chen2021evaluating,
      title={Evaluating Large Language Models Trained on Code}, 
      author={Mark Chen and Jerry Tworek and Heewoo Jun and Qiming Yuan and Henrique Ponde de Oliveira Pinto and Jared Kaplan and Harri Edwards and Yuri Burda and Nicholas Joseph and Greg Brockman and Alex Ray and Raul Puri and Gretchen Krueger and Michael Petrov and Heidy Khlaaf and Girish Sastry and Pamela Mishkin and Brooke Chan and Scott Gray and Nick Ryder and Mikhail Pavlov and Alethea Power and Lukasz Kaiser and Mohammad Bavarian and Clemens Winter and Philippe Tillet and Felipe Petroski Such and Dave Cummings and Matthias Plappert and Fotios Chantzis and Elizabeth Barnes and Ariel Herbert-Voss and William Hebgen Guss and Alex Nichol and Alex Paino and Nikolas Tezak and Jie Tang and Igor Babuschkin and Suchir Balaji and Shantanu Jain and William Saunders and Christopher Hesse and Andrew N. Carr and Jan Leike and Josh Achiam and Vedant Misra and Evan Morikawa and Alec Radford and Matthew Knight and Miles Brundage and Mira Murati and Katie Mayer and Peter Welinder and Bob McGrew and Dario Amodei and Sam McCandlish and Ilya Sutskever and Wojciech Zaremba},
      year={2021},
      eprint={2107.03374},
      archivePrefix={arXiv},
      primaryClass={cs.LG},
      url={https://arxiv.org/abs/2107.03374}, 
}

@inproceedings{wei2022chain,
 author = {Wei, Jason and Wang, Xuezhi and Schuurmans, Dale and Bosma, Maarten and Ichter, Brian and Xia, Fei and Chi, Ed and Le, Quoc V and Zhou, Denny},
 booktitle = {Advances in Neural Information Processing Systems},
 doi = {10.52202/068431-1800},
 pages = {24824--24837},
 publisher = {Curran Associates, Inc.},
 title = {Chain-of-Thought Prompting Elicits Reasoning in Large Language Models},
 url = {https://proceedings.neurips.cc/paper_files/paper/2022/file/9d5609613524ecf4f15af0f7b31abca4-Paper-Conference.pdf},
 volume = {35},
 year = {2022}
}

@inproceedings{zhang2025agentracer,
title={{AgenTracer}: Who Is Inducing Failure in the {LLM} Agentic Systems?},
author={Guibin Zhang and Junhao Wang and Junjie Chen and Wangchunshu Zhou and Kun Wang and Shuicheng YAN},
booktitle={The Fourteenth International Conference on Learning Representations},
year={2026},
url={https://openreview.net/forum?id=l05DseqvuD}
}

@inproceedings{wang2024promptagent,
 author = {Wang, Xinyuan and Li, Chenxi and Wang, Zhen and Bai, Fan and Luo, Haotian and Zhang, Jiayou and Jojic, Nebojsa and Xing, Eric  and Hu, Zhiting},
 booktitle = {International Conference on Learning Representations},
 pages = {23967--24001},
 title = {{PromptAgent}: Strategic Planning with Language Models Enables Expert-level Prompt Optimization},
 url = {https://proceedings.iclr.cc/paper_files/paper/2024/file/686a3f32067838c8dbb68da6e9e3cf69-Paper-Conference.pdf},
 volume = {2024},
 year = {2024}
}

@misc{liu2026rewritefixalltypeaware,
  title={One Rewrite to Fix Them All? Type-Aware Repair Allocation for Text-to-Image Prompt Optimization},
  author={Liu, Haoyue and Ma, Xiaoyu and Chen, Ye and Cui, Shuguang and Tang, Xiaoying},
  year={2026},
  eprint={2607.18724},
  archivePrefix={arXiv},
  primaryClass={cs.AI},
  url={https://arxiv.org/abs/2607.18724}
}

@misc{ma2026selectsmartermorepromptaware,
  title={Select Smarter, Not More: Prompt-Aware Evaluation Scheduling with Submodular Guarantees},
  author={Ma, Xiaoyu and Li, Yiwen and Liu, Haoyue and Wang, Zhichao and Chen, Ye and Guo, Yongxin and Tang, Xiaoying},
  year={2026},
  eprint={2604.11328},
  archivePrefix={arXiv},
  primaryClass={cs.AI},
  url={https://arxiv.org/abs/2604.11328}
}

@misc{wang2026planscheckverifiergroundedlearning,
      title={Plans You Can Check: Verifier-Grounded Learning of an Open-Weight Planner for Executable Video-Editing}, 
      author={Haoyu Wang and Cheng Feng and Liuyang Bian and Ruiyang Huang and Lei Wei and Yafei Wen and Xiaoxin Chen and Xiaoying Tang},
      year={2026},
      eprint={2608.25622},
      archivePrefix={arXiv},
      primaryClass={cs.CV},
      url={https://arxiv.org/abs/2608.25622}, 
}

\clearpage
\appendix

\section{Reproducibility Statement}
\label{app:reproducibility}

This appendix consolidates the information needed to reproduce every
number in \S\ref{sec:experiments}.  Sections~\ref{app:datasets}--%
\ref{app:reporting} cover data, models, compute, and reporting
conventions; Sections~\ref{app:prompts}--\ref{app:algorithm} cover the
\sepo loop and architect prompts; Section~\ref{app:baselines} covers
each baseline; the remaining appendices report extended quantitative
results, ablations, and qualitative analysis.

\subsection{Datasets, Splits, and Licences}
\label{app:datasets}

Table~\ref{tab:datasets} summarises the 14 held-out tasks used in
\S\ref{sec:experiments}, drawn from five public benchmarks:
\textbf{BBH} (8 reasoning sub-tasks),
\textbf{MMLU-Pro} (3 subject sub-tasks),
\textbf{MBPP+},
\textbf{GSM-Hard}, and
\textbf{HotpotQA}.  For each task we report
the source benchmark, the number of optimisation-time training
examples ($|\mathcal{D}_{\mathrm{tr}}|$), the size of the held-out test split,
the task metric, and the source licence.  All splits are fixed
across optimisers and seeds; the same indices are used for every
method.  No example in $\mathcal{D}_{\mathrm{tr}}$ overlaps with the held-out
test split.
\hypersetup{
  pdftitle={SEPO: Evidence-Grounded Prompt Optimization via Structural Editing},
  pdfauthor={Xiaoyu Ma, Haoyue Liu, Yiwen Li, Jionghao Zhu, Zhichao Wang, Ye Chen, Xiaoying Tang}
}
\paragraph{Evidence-packet probes.}
Every task uses a fixed training set $\mathcal{D}_{\mathrm{tr}}$ and a
disjoint held-out test split.  Evidence packets, including the
$M_{\mathrm{train}}$ slot in Appendix~\ref{app:hyperparams}, are
constructed from this training set after the focus and anchor
slots have been selected.  This convention is used uniformly across
BBH, MMLU-Pro, MBPP+, HotpotQA, and GSM-Hard
(Table~\ref{tab:datasets}).

\begin{table}[!ht]
\centering
\scriptsize
\setlength{\tabcolsep}{3pt}
\renewcommand{\arraystretch}{1.10}
\arrayrulecolor{tableRule}
\begin{tabular}{@{}lccll@{}}
\toprule
\textbf{Task} & $|\mathcal{D}_{\mathrm{tr}}|$ & \textbf{Test} & \textbf{Metric} & \textbf{Licence} \\
\midrule
BBH Causal Judgement      &  50 & 137 & exact match  & MIT \\
BBH Date Understanding    &  50 & 200 & exact match  & MIT \\
BBH Disambiguation QA     &  50 & 200 & exact match  & MIT \\
BBH Formal Fallacies      &  50 & 200 & exact match  & MIT \\
BBH Logical Deduction (7) &  50 & 200 & exact match  & MIT \\
BBH Colored Objects       &  50 & 200 & exact match  & MIT \\
BBH Snarks                &  50 & 128 & exact match  & MIT \\
BBH Word Sorting          &  50 & 200 & exact match  & MIT \\
MMLU-Pro CS               & 100 & 200 & multi-choice & MIT \\
MMLU-Pro Economics        & 100 & 200 & multi-choice & MIT \\
MMLU-Pro Psychology       & 100 & 200 & multi-choice & MIT \\
MBPP+                     & 100 & 228 & pass@1       & Apache-2.0 \\
GSM-Hard                  & 200 & 400 & exact match  & MIT \\
HotpotQA$^\dagger$        & 200 & 400 & EM (string)  & CC BY-SA \\
\bottomrule
\end{tabular}
\arrayrulecolor{black}
\caption{Dataset statistics and licences.  All BBH and MMLU-Pro
splits follow each benchmark's original release.  Every task uses
the same evidence-packet convention described above.
$^\dagger$\textbf{HotpotQA split swap.}  We use the
\emph{Medium} split under the Llama worker and the \emph{Hard}
split under the Qwen worker.  The two splits share the same
\texttt{distractor}-setting passages but differ in extraction
difficulty: the Hard split requires the worker to resolve a
distractor entity before extracting the answer, while the Medium
split has cleaner gold-passage alignment. Under the Llama worker,
limited answer accessibility on Hard compresses performance across
optimisers; a matched Hard-split sensitivity test preserves their
ranking (Appendix~\ref{app:hqa-hard-sensitivity}). We therefore use
Medium under Llama and Hard under Qwen. The implication is that the HotpotQA
column in Table~\ref{tab:sepo-per-task} (column ``HQA'') is not
directly comparable between the Llama and Qwen blocks; the gap
across blocks reflects both the worker change and the split swap.}
\label{tab:datasets}
\end{table}

All datasets are used in accordance with their licences and intended
research-evaluation purpose.  None contains personally identifying
information; HotpotQA passages are drawn from Wikipedia under the
CC BY-SA licence.

\subsection{Models and Decoding Configuration}
\label{app:model-config}

We keep the evaluation harness fixed across optimisers and random
seeds.  The frozen worker LLMs are
\textbf{Llama-3.1-8B-Instruct}
(Meta, $8.0$B parameters) and
\textbf{Qwen3-8B} (Alibaba, $7.6$B parameters);
each worker
is used for both optimisation-time scoring and held-out deployment.
Worker inference uses temperature $0.0$, top-$p$ $1.0$, and a
maximum generation length of $2{,}048$ tokens, so that metric calls
are deterministic under a fixed prompt and example.

\sepo uses \textbf{Qwen3.5-397B-A17B}
(a Mixture-of-Experts model
with $17$B active parameters out of a $397$B total) as the architect
LLM for the attribution and patch calls, with temperature $0.7$ and
a maximum generation length of $16{,}000$ tokens to encourage
diverse candidate edits.  Baselines are run through their
corresponding API-only optimiser interfaces while sharing the same
worker, data split, and metric-call budget; baseline-specific
architect choices are reported in Appendix~\ref{app:baselines}.

\subsection{Random Seeds and Reporting Conventions}
\label{app:reporting}

For each (method, benchmark) cell we run three independent
optimisation seeds; the deployment held-out evaluation reuses the
optimiser's selected prompt and is itself deterministic
(temperature $0$).  Tables report the seed mean with the standard
deviation as the subscript (e.g., $62.0_{\pm 2.6}$).  All accuracies are reported in
percent on the held-out test split, not the training split used
for optimiser feedback.

\subsection{Use of AI Assistants}
\label{app:ai-assistants}

We used an AI assistant solely for English proofreading and
stylistic polishing.  All technical content, experiments, and
conclusions were produced and verified by the authors.

\section{Prompts: Seed Schema and Architect Templates}
\label{app:prompts}

This appendix lists the three text artefacts that ground every
\sepo iteration: the cold-start seed schema $s_0$, the attribution
architect's prompt template, and the patch architect's prompt
template.  The schema parser at the end of the section is the
deterministic counterpart that converts the architect's JSON
output back into a two-layer schema.

\subsection{Seed Schema and Initialisation}
\label{app:seed-schema}

The cold-start seed schema $s_0$ is \emph{task-agnostic}: every
benchmark in our suite starts from the same generic three-step
backbone with no step-local rules.  This puts every optimiser on
an equal footing --- no task-specific priors are injected into the
seed --- and means the first iteration's $\textsc{newly\_broken}$
is empty by construction, so the loop enters the no-regression
fallback (Appendix~\ref{app:train-only-fallback}) on iteration~1.  The exact
seed used in all runs is shown below.

\begin{sepoexample}{Seed schema $s_0$ (task-agnostic)}
\footnotesize\ttfamily
1. Understand the problem — read carefully, parse all entities, and decompose into sub-questions if needed.\\[1pt]
2. Reason step by step.\\[1pt]
3. Output the final answer on the last line as: Answer: (X).
\end{sepoexample}

\noindent
The seed has no step-local rules: all sub-rules in the converged
schema (e.g.\ the BBH Word Sorting example in \S\ref{sec:schema})
are introduced by the patch architect during optimisation; no rule
is copied from $\mathcal{D}_{\mathrm{tr}}$ or any
benchmark-specific prompt template.

\subsection{Attribution Prompt Template}
\label{app:prompt-attribution}

\sepo's first architect call returns a typed target $\tau$ and a
failure-pattern rationale; it does not rewrite the schema.  The
prompt is identical across benchmarks --- only the schema text,
the evidence packet, and the breadcrumb vary per call.  Below is
the full block structure of the attribution prompt;
\{\}-braced fields are populated at render time.

\begin{sepoexample}{Attribution architect prompt}
\footnotesize\ttfamily
Your task: diagnose why the current schema is failing, by reading parent's behavior on the train cases shown below.  Decide which single schema component to fix, and explain why in your own words.\\[3pt]
The schema is the worker's system prompt across MANY cases of this task family --- not just the few cases shown here.  Your fix should generalize to unseen cases of this task family.  Different evidence cases may fail for materially different reasons; if so, say that explicitly in your rationale instead of forcing them into a single narrative.\\[6pt]
The schema has been refined across prior optimization rounds via discrete, traceable patches.  The block below shows what the LAST patch did, so you can analyse its observed example-level effects.\\
\{lineage breadcrumb $\ell$ --- op-specific before-state diff\}\\[6pt]
=== CURRENT SCHEMA ===\\
(Numbered lines '1.', '2.', '3.' are TOP-LEVEL STEPS.  Indented '\,-' lines are SUB-RULES belonging to the most recent numbered step above them.)\\
\{$p$\}\\
=== END SCHEMA ===\\[3pt]
=== Parent on train ===\\
\{focus + anchor + train cases, each rendered as Train Case \#i with transition tag, Input, Gold, Worker output\}\\[6pt]
=== ANALYSIS GUIDANCE ===\\
1.~Read each evidence case yourself --- input, gold, worker output \ldots\\
2.~If evidence cases span materially different failure subtypes, say so explicitly and pick the SINGLE dominant subtype to target.\\[3pt]
=== TARGET PREFERENCE ===\\
Choose the smallest target that explains the dominant failure:\\
1.~Use "step <N>" when the failure can be repaired inside step N.\\
2.~Use explicit structure targets only when the top-level numbered step structure itself is the problem:\\
\hspace*{1em}- "ADD step at <pos>" \quad - "SPLIT step <N>"\\
\hspace*{1em}- "DROP step <N>"\\
3.~Use "backbone" only when the whole 1./2./3.\ macro structure is wrong.\\[3pt]
=== OUTPUT ===\\
Return ONLY a JSON object:\\
\{"target": "step <N>" | "ADD step at <pos>" | "SPLIT step <N>" | "DROP step <N>" | "backbone",\\
~"rationale": "1--2 sentences, MAX 300 chars",\\
~"supporting\_observations": [2--4 short observations]\}
\end{sepoexample}

\subsection{Patch Prompt Template}
\label{app:prompt-patch}

The second architect call consumes the same evidence packet plus
the attribution output $(\tau, \text{rationale})$ and emits the
full revised schema $\widetilde{s}$.  We list the full block
structure of the patch prompt below; the style block uses the
balanced variant in every run of our launch scripts.

\begin{sepoexample}{Patch architect prompt}
\footnotesize\ttfamily
Your task: produce ONE revised version of the schema below so the worker stops failing on the diagnosed pattern.\\[3pt]
The schema is the worker's system prompt across MANY cases of this task family; the cases below are a small sample it will NOT see again.  Your patch must generalize.\\[6pt]
A diagnosis has already been performed (see THIS ROUND'S DIAGNOSIS below).  Use it together with the train evidence and the lineage breadcrumb to decide what to change.\\
\{lineage breadcrumb $\ell$\}\\[6pt]
=== THIS ROUND'S DIAGNOSIS ===\\
~~target:~~~~~\{$\tau$\}\\
~~rationale:~\{rationale from attribution\}\\[3pt]
=== SCOPE RULE ===\\
For target='step <N>', target is the step at the locus of the dominant failure.  Choose the smallest applicable operation below:\\
~~- edit an existing sub-rule under the target step\\
~~- drop a sub-rule under the target step\\
~~- merge two overlapping sub-rules under the target step into one\\
~~- add a new sub-rule under the target step\\
~~- rewrite the target step's numbered top-level text\\[3pt]
If target is explicitly structural, execute that structure edit:\\
~~- target='ADD step at <pos>': insert a new top-level numbered step at position <pos>.\\
~~- target='SPLIT step <N>': split step N into two consecutive top-level steps.\\
~~- target='DROP step <N>': remove step N as a top-level step.\\
~~- target='backbone': rewrite the global top-level structure.\\[3pt]
Sub-rules under non-target steps are preserved unless they become irrelevant after the change.\\[6pt]
=== CURRENT SCHEMA ===\\
\{$p$\}\\
=== END SCHEMA ===\\[3pt]
=== Parent on train ===\\
\{focus + anchor + train cases with transition tags\}\\[6pt]
=== STYLE GUIDANCE ===\\
From the worker responses and gold answers above, identify the niche, task-specific procedures, constraints, and edge cases the worker is getting wrong, and encode each as a directly executable rule with a worked example.  Avoid jargon; every rule must be directly executable.\\[3pt]
=== OUTPUT SCHEMA FORMAT ===\\
Schema: numbered points (1./2./3./...) with "\,-" indented elaborations.  Step text can be a short imperative OR a longer paragraph with persona/role context.\\[3pt]
Example output for target='step 2' (adding two task-specific sub-rules under step 2):\\
~~1. Understand the problem.\\
~~2. Reason step by step.\\
~~~~- When the option uses positive words to describe an objectively bad situation, classify as sarcasm.\\
~~~~- Compare literal meaning against real-world facts; if mismatched, the option is sarcastic.\\
~~3. Output the answer on the last line as: Answer: (X).\\[3pt]
In this example the backbone (1./2./3.) is unchanged; only elaborations were added under step 2.\\[3pt]
=== OUTPUT FORMAT ===\\
Return ONLY a JSON object:\\
\{"schema": "<complete outline-tree markdown>",\\
~"rationale": "one or two sentences explaining what changed and why"\}
\end{sepoexample}

\subsection{Schema Parser}
\label{app:parser}

A deterministic parser extracts the outline step list and per-step
rule lists from the patch architect's JSON output.  Across the
attribution and patch stages, our runs made $2{,}520$ architect calls
in total.  Of these, 58 patch calls ($2.3\%$ of all architect calls)
returned missing JSON or schema text that did not match the
``\texttt{<N>.}'' regex.  The corresponding children were discarded
with the audit tuple recorded as \texttt{parse\_failure}, and all
consumed tokens were included in the reported optimisation cost.

\section{Algorithm Details}
\label{app:algorithm}

We give two views of \sepo's per-iteration loop.
Algorithm~\ref{alg:sepo} is the slim, audit-tuple-first description
of one iteration, with explicit cross-references to
attribution (Eq.~\ref{eq:attribution}), patch (Eq.~\ref{eq:patch}),
operator inference (Eq.~\ref{eq:operator}), and example-level
admission (Eq.~\ref{eq:admission}).
Algorithm~\ref{alg:sepo-full} is the engineering view that spells
out the helpers (parent selection, evidence-packet assembly,
breadcrumb rendering, subsample precheck, archive capacity).  The
two algorithms describe the same loop.

\paragraph{Metric-call accounting.}
The focus and anchor slots provide cached parent outputs as architect
evidence; Gate~1 evaluates the new child only on the $n_{tr}$ examples
in $M_{\mathrm{train}}$.  A candidate rejected at Gate~1 therefore
incurs $n_{tr}$ new metric calls.  For a candidate that passes, these
$n_{tr}$ child results are reused and only the remaining
$|\mathcal{D}_{\mathrm{tr}}|-n_{tr}$ examples are evaluated, for
$|\mathcal{D}_{\mathrm{tr}}|$ new metric calls in total.
Below, \textsc{CompleteEval} evaluates only the uncached training
examples and then assembles the full score vector with the cached
Gate~1 results.

\begin{algorithm}[t]
\small
\DontPrintSemicolon
\SetKwInOut{Input}{Input}\SetKwInOut{Output}{Output}
\caption{\sepo: one iteration of lineage-aware structural editing.}
\label{alg:sepo}
\Input{seed schema $s_0$; training set $\mathcal{D}_{\mathrm{tr}}$; worker $W$; architects $A_{\mathrm{attr}},
       A_{\mathrm{patch}}$; probe size $n_{tr}$; metric-call budget $B$.}
\Output{archive $\mathcal{A}$.}
\BlankLine
$\mathbf{c}(s_0) \leftarrow \textsc{Eval}(W, s_0, \mathcal{D}_{\mathrm{tr}})$;\quad
$\mathcal{A} \leftarrow \{s_0\}$;\quad
$u \leftarrow |\mathcal{D}_{\mathrm{tr}}|$\;
\While{$u+|\mathcal{D}_{\mathrm{tr}}| \le B$}{
  $p \leftarrow \textsc{SelectParent}(\mathcal{A})$;\quad
  $g \leftarrow \pi(p)$
     \tcp*{Lexicase w.p.~$0.8$; uniform w.p.~$0.2$}
  \eIf{$g=\emptyset$}{
    $\textsc{nb},\textsc{nf} \leftarrow \emptyset,\emptyset$
       \tcp*{cold start}
  }{
    $\textsc{nb} \leftarrow \{i : \mathbf{c}(g)_i{=}1,\, \mathbf{c}(p)_i{=}0\}$;\quad
    $\textsc{nf} \leftarrow \{i : \mathbf{c}(g)_i{=}0,\, \mathbf{c}(p)_i{=}1\}$\;
  }
  $M \leftarrow \bigl(M_{\mathrm{focus}}, M_{\mathrm{anchor}}, M_{\mathrm{train}}\bigr)$
     \tcp*{sampled from $(\textsc{nb}, \textsc{nf}, \mathcal{D}_{\mathrm{tr}})$;
           fallback if $\textsc{nb}{=}\emptyset$}
  $\ell \leftarrow \textsc{Breadcrumb}\bigl(o(p),\tau(p),
       +|\textsc{nf}|,-|\textsc{nb}|\bigr)$\;
  $(\tau,r) \leftarrow A_{\mathrm{attr}}\!\bigl(p,M,\ell\bigr)$
     \tcp*{Eq.~\ref{eq:attribution}}
  $\widetilde{s} \leftarrow A_{\mathrm{patch}}\!\bigl(p,\tau,r,M,\ell\bigr)$
     \tcp*{Eq.~\ref{eq:patch}}
  $c \leftarrow \textsc{ParseSchema}(\widetilde{s})$;\quad
  \lIf{$c=\bot$}{\textbf{continue}
     \tcp*[h]{parse failure}}
  $\pi(c) \leftarrow p$;\quad $\tau(c) \leftarrow \tau$;\quad
  $o(c) \leftarrow \textsc{InferOp}\!\bigl(\textsc{Diff}(p, c),\tau(c)\bigr)$
     \tcp*{Eq.~\ref{eq:operator}}
  $\mathbf{c}_{M_{\mathrm{train}}}(c) \leftarrow
     \textsc{Eval}(W,c,M_{\mathrm{train}})$\;
  \lIf{$\sum_{i \in M_{\mathrm{train}}}\!\mathbf{c}_{M_{\mathrm{train}}}(c)_i \,\le\,
         \sum_{i \in M_{\mathrm{train}}}\!\mathbf{c}(p)_i$}
        {$u \mathrel{+}{=} n_{tr}$;\, \textbf{continue}
     \tcp*[h]{Gate 1: precheck}}
  $\mathbf{c}(c) \leftarrow
     \textsc{CompleteEval}(W,c,\mathcal{D}_{\mathrm{tr}};
       \mathbf{c}_{M_{\mathrm{train}}}(c))$
     \tcp*{reuse Gate~1 results}
  $u \mathrel{+}{=} |\mathcal{D}_{\mathrm{tr}}|$\;
  \lIf{\textnormal{Eq.~\ref{eq:admission} holds}}{$\mathcal{A} \leftarrow \mathcal{A} \cup \{c\}$
     \tcp*[h]{Gate 2: admit}}
}
\Return $\mathcal{A}$
\end{algorithm}

\subsection{Expanded Engineering View}
\label{app:algorithm-full}

\begin{algorithm}[!b]
\small
\caption{\sepo expanded loop with helper subroutines.}
\label{alg:sepo-full}
\KwIn{seed $s_0$;\; training set $\mathcal{D}_{\mathrm{tr}}$;\; $W,\, A$;\;
      budget $B$;\; iter cap $T$;\; archive cap $C$;\;
      $(n_f, n_a, n_{tr})$.}
\KwOut{archive $\mathcal{A}$.}
$\mathbf{c}(s_0) \!\leftarrow\! \textsc{Eval}(W, s_0, \mathcal{D}_{\mathrm{tr}})$;\;
$\mathcal{A} \!\leftarrow\! \{s_0\}$;\;
$u \!\leftarrow\! |\mathcal{D}_{\mathrm{tr}}|$\;
\For{$t = 1, \ldots, T$}{
  \lIf{$u + |\mathcal{D}_{\mathrm{tr}}| > B$}{\textbf{break}}
  $p \leftarrow \textsc{LexicaseOrUniform}(\mathcal{A})$
     \tcp*{Lexicase w.p.~$0.8$; uniform w.p.~$0.2$}
  $M \leftarrow \textsc{BuildMinibatch}(p, \mathcal{A},
      \mathcal{D}_{\mathrm{tr}})$
     \tcp*{fallback when $\textsc{nb}(p){=}\emptyset$}
  $\ell \leftarrow \textsc{RenderBreadcrumb}(p, \pi(p))$\;
  $(\tau,r) \leftarrow A_{\mathrm{attr}}\bigl(p,\, M,\, \ell\bigr)$
     \tcp*{Eq.~\ref{eq:attribution}}
  $\widetilde{s} \leftarrow A_{\mathrm{patch}}\bigl(p,\, \tau,\, r,\,
      M,\, \ell\bigr)$
     \tcp*{Eq.~\ref{eq:patch}}
  \lIf{$\textsc{ParseSchema}(\widetilde{s})$ fails}{\textbf{continue}}
  $c \leftarrow$ schema with $\pi(c){=}p,\, \tau(c){=}\tau$\;
  $o(c) \leftarrow \textsc{InferOp}(\textsc{Diff}(p, c),\tau(c))$
     \tcp*{Eq.~\ref{eq:operator}}
  \lIf{\textsc{SubsamplePrecheck}$(p, c, M_{\mathrm{train}})$ rejects}
      {$u \mathrel{+}{=} n_{tr}$;\, \textbf{continue}}
  $\mathbf{c}(c) \leftarrow
     \textsc{CompleteEval}(W,c,\mathcal{D}_{\mathrm{tr}})$
     \tcp*{reuse Gate~1 results}
  $u \mathrel{+}{=} |\mathcal{D}_{\mathrm{tr}}|$\;
  \textsc{Admit}$(c, p, \mathcal{A})$
     \tcp*{Gate 2, Eq.~\ref{eq:admission}}
  \lIf{$|\mathcal{A}| > C$}
      {\textsc{ManageCapacity}$(\mathcal{A};\, C)$}
}
\Return $\mathcal{A}$
\end{algorithm}

\paragraph{Where each module appears.}
The five-module decomposition of \S\ref{sec:method} maps onto
Algorithm~\ref{alg:sepo-full} as follows.  Parent selection
(\S\ref{sec:loop}) is the \textsc{LexicaseOrUniform} call; the
counterfactual minibatch (\S\ref{sec:editing} Step~1) is
\textsc{BuildMinibatch}, which contains the no-regression fallback
when $\textsc{newly\_broken}(p)$ is empty; the lineage breadcrumb
(\S\ref{sec:editing} Step~2) is \textsc{RenderBreadcrumb}; the attribution
and patch architects (\S\ref{sec:editing} Steps~3--4) are the
$A_{\mathrm{attr}}$ and $A_{\mathrm{patch}}$ calls; the two-layer
schema (\S\ref{sec:schema}) is materialised by \textsc{ParseSchema}
and the assignment of $\pi(c), \tau(c)$ on the next line;
operator inference (\S\ref{sec:schema}) is the $\textsc{InferOp}$
call; the subsample precheck and example-level admission
(\S\ref{sec:loop}) are \textsc{SubsamplePrecheck} and
\textsc{Admit}; archive capacity is \textsc{ManageCapacity}.

\subsection{Operator Label Space}
\label{app:operator-space}

The schema of Eq.~\ref{eq:schema} is two-layer (a top-level
\emph{outline} plus step-local \emph{rules}), so the
operator vocabulary $\mathcal{O}$ in Eq.~\ref{eq:operator} is
naturally indexed by which layer it edits.
Table~\ref{tab:operator-vocab} lists the eight sub-operations
in $\mathcal{O} = \mathcal{O}_{\mathrm{rule}} \cup
\mathcal{O}_{\mathrm{outline}}$ together with the architect's
target form $\tau$ that elicits each one (the architect chooses
``the smallest applicable operation'' from these alternatives;
see Appendix~\ref{app:prompt-patch}).
$\mathcal{O}_{\mathrm{rule}}$ contains four rule-layer sub-ops
that act on the step-local rule list, leaving the top-level
outline unchanged; $\mathcal{O}_{\mathrm{outline}}$ contains
four outline-layer sub-ops that touch the top-level outline.
Each sub-op is its own $o(c)$ label, so the post-hoc operator
takes eight distinct values; these are the eight labels counted
in Table~\ref{tab:edit-distribution} of the main paper.

\begin{table}[!ht]
\centering
\scriptsize
\setlength{\tabcolsep}{3pt}
\renewcommand{\arraystretch}{1.15}
\arrayrulecolor{tableRule}
\resizebox{\columnwidth}{!}{%
\begin{tabular}{@{}lll@{}}
\toprule
\textbf{Sub-op} & \textbf{Target $\tau$} & \textbf{Effect on schema} \\
\midrule
\multicolumn{3}{@{}l}{\textit{$\mathcal{O}_{\mathrm{rule}}$ (rule-layer): 4 sub-ops, all under \texttt{step <N>}}} \\
\textsc{AddRule}    & \texttt{step <N>} & insert sub-rule in step $N$ \\
\textsc{DropRule}   & \texttt{step <N>} & remove sub-rule from step $N$ \\
\textsc{RefineRule} & \texttt{step <N>} & rewrite one sub-rule in step $N$ \\
\textsc{MergeRules} & \texttt{step <N>} & merge two sub-rules in step $N$ \\
\midrule
\multicolumn{3}{@{}l}{\textit{$\mathcal{O}_{\mathrm{outline}}$ (outline-layer): 4 sub-ops, one $o(c)$ label each}} \\
\textsc{RewriteStep} & \texttt{step <N>}          & rewrite step $N$'s top-level text \\
\textsc{AddStep}     & \texttt{ADD step at <pos>} & insert new top-level step at \texttt{<pos>} \\
\textsc{SplitStep}   & \texttt{SPLIT step <N>}    & split step $N$ into two steps \\
\textsc{Backbone}    & \texttt{backbone}          & rewrite entire $1./2./3.$ structure \\
\bottomrule
\end{tabular}}
\arrayrulecolor{black}
\caption{Operator vocabulary $\mathcal{O}$ decomposed by schema
layer.  Each row is a sub-operation together with the architect's
target form $\tau$ that elicits it; each row also maps to its own
post-hoc $o(c)$ label.  All four rule-layer sub-ops live inside
one step's rule list.  All four outline-layer sub-ops change the
top-level outline structure or text and are distinguished post-hoc
from the diff together with the typed target $\tau$
(Alg.~\ref{alg:diff-schemas}); the architect never declares the
operator label itself.  The architect's \texttt{DROP step <N>}
target form is supported in the prompt template
(Appendix~\ref{app:prompt-attribution}) but yielded no admitted
children in our 14-task $\times$ 3-seed runs, so it does not
appear as a separate $o(c)$ label here.  The verbatim
patch-prompt phrase eliciting each sub-op is given in
Appendix~\ref{app:prompt-patch} (\texttt{SCOPE RULE} block).}
\label{tab:operator-vocab}
\end{table}

The architect's \emph{target} $\tau$ has a slightly richer
surface vocabulary
($\texttt{step <N>}$, $\texttt{ADD step at <pos>}$,
$\texttt{SPLIT step <N>}$, $\texttt{DROP step <N>}$,
$\texttt{backbone}$; see Appendix~\ref{app:prompt-attribution})
than $o(c)$ because $\tau$ pre-locates the edit before the patch
is written.  The realised $o(c)$ is recovered post-hoc from the
structural diff (Alg.~\ref{alg:diff-schemas}); the
\texttt{DROP step <N>} target was never admitted in our runs
and therefore has no corresponding $o(c)$ label in
Table~\ref{tab:operator-vocab}.

\paragraph{Why $\textsc{Diff}$ is non-trivial.}
The patch architect returns a free-form schema text $\widetilde{s}$;
it does not annotate which child rule corresponds to which parent
rule, nor does it self-declare an edit operator.  Recovering
$o(c) \in \mathcal{O}$ therefore reduces to first computing a
\emph{structural} parent--child alignment, after which the operator
label follows by counting non-empty buckets.  Because rules are
short natural-language strings, we align by sequence similarity
rather than by edit distance over tokens, and we use two thresholds
to separate three cases that the patch prompt actually distinguishes
(Appendix~\ref{app:prompt-patch}, \texttt{SCOPE RULE}): a refinement
of one parent rule, a merge of two parent rules into one, and the
introduction of a new rule.

Algorithm~\ref{alg:diff-schemas} computes the diff at two levels.
It first aligns the outline steps in order, recording added, dropped,
refined, and copied step locations.  Within each aligned step pair,
each child rule is greedily matched against the top-two unused parent
rules from that same step under the similarity $\sigma$.  The three
branches (\textsc{refined}, \textsc{merged}, \textsc{added}) are gated
by the two thresholds $\theta_R, \theta_M$ together with an exact-match
shortcut (\textsc{copied}) and a merge-evidence predicate
$\textsc{LooksLikeMerge}$ defined below.  Any parent rule to which no
child rule was assigned is reported as \textsc{dropped}.  The resulting
hierarchical diff record, together with the typed target $\tau$, is the
input to $\textsc{InferOp}$; its per-step outline locations are also
used for breadcrumb rendering (Appendix~\ref{app:breadcrumb-rendering}).
Here $\textsc{AlignOutline}$ computes a maximum-similarity monotone
one-to-one alignment of outline steps under $\sigma$ (ties are resolved
by the lower parent index); unmatched child and parent steps are marked
as added and dropped, while matched steps are copied if their stripped
texts are identical and refined otherwise.

\begin{algorithm}[t]
\small
\DontPrintSemicolon
\SetKwInOut{Input}{Input}\SetKwInOut{Output}{Output}
\caption{$\textsc{Diff}(p, c)$ --- deterministic structural diff
between a parent and child schema.  Used by $\textsc{InferOp}$
(Eq.~\ref{eq:operator}) and the breadcrumb renderer
(Appendix~\ref{app:breadcrumb-rendering}).}
\label{alg:diff-schemas}
\Input{parent schema
       $p{=}(\alpha_p,\{\rho_{p,i}\}_{i=1}^{m_p})$ and child schema
       $c{=}(\alpha_c,\{\rho_{c,j}\}_{j=1}^{m_c})$, where
       $R_{p,k}{=}\{\rho_{p,k,i}\}$ and
       $R_{c,k'}{=}\{\rho_{c,k',j}\}$ denote the rule lists attached
       to parsed steps $k$ and $k'$;
       similarity $\sigma(s_1, s_2)$ (Python
       \texttt{difflib.SequenceMatcher} ratio, $\in[0,1]$);
       thresholds $\theta_R{=}0.55$ (refine), $\theta_M{=}0.40$
       (merge).}
\Output{hierarchical diff
        $\mathcal{D}=(\mathcal{D}_{\mathrm{outline}},
        \mathcal{D}_{\mathrm{rule}})$.}
\BlankLine
$H,\mathcal{D}_{\mathrm{outline}}
  \leftarrow \textsc{AlignOutline}(\alpha_p,\alpha_c;\sigma)$
  \tcp*{order-preserving step alignment}
$\textsc{added}, \textsc{dropped}, \textsc{refined},
 \textsc{merged}, \textsc{copied} \leftarrow \emptyset$\;
\ForEach{aligned step pair $(k,k') \in H$}{
  $\textsc{used}[k,i] \leftarrow \textsc{False}$ \textbf{for}
    $i=1,\ldots,|R_{p,k}|$\;
  \For{$j = 1, \ldots, |R_{c,k'}|$}{
    $S \leftarrow$ list of
       $(i,\sigma(\rho_{p,k,i},\rho_{c,k',j}))$ for
       $\neg\textsc{used}[k,i]$, sorted by $\sigma$ descending\;
    \lIf{$S = \emptyset$ \textbf{or} $S_1.\sigma < \theta_M$}{
      $\textsc{added} \mathrel{+}{=} \{(k',j)\}$;\, \textbf{continue}}
    $(i^\star,s^\star) \leftarrow S_1$\;
    \lIf{$\textsc{strip}(\rho_{p,k,i^\star})=
            \textsc{strip}(\rho_{c,k',j})$}{
      $\textsc{copied} \mathrel{+}{=} \{(k,i^\star,k',j)\}$;\,
      $\textsc{used}[k,i^\star] \leftarrow \textsc{True}$;\,
      \textbf{continue}}
    \uIf{$|S|\geq2 \,\land\, S_2.\sigma\geq\theta_M
          \,\land\, s^\star<0.95
          \,\land\,\textsc{LooksLikeMerge}(\rho_{p,k,i^\star},
                     \rho_{p,k,S_2.i},\rho_{c,k',j})$}{
      $\textsc{merged} \mathrel{+}{=}
        \{(k,i^\star,S_2.i,k',j)\}$;\;
      $\textsc{used}[k,i^\star],\textsc{used}[k,S_2.i]
        \leftarrow \textsc{True},\textsc{True}$\;
    }
    \uElseIf{$s^\star\geq\theta_R$}{
      $\textsc{refined} \mathrel{+}{=} \{(k,i^\star,k',j)\}$;\,
      $\textsc{used}[k,i^\star] \leftarrow \textsc{True}$\;
    }
    \lElse{$\textsc{added} \mathrel{+}{=} \{(k',j)\}$}
  }
  $\textsc{dropped} \mathrel{+}{=}
    \{(k,i):\neg\textsc{used}[k,i]\}$\;
}
$\begin{aligned}
\mathcal{D}_{\mathrm{rule}} \leftarrow
  (&\textsc{added},\textsc{dropped},\\[-1pt]
   &\textsc{refined},\textsc{merged},\textsc{copied})
\end{aligned}$\;
$\mathcal{D} \leftarrow
  (\mathcal{D}_{\mathrm{outline}},\mathcal{D}_{\mathrm{rule}})$\;
\Return $\mathcal{D}$\;
\end{algorithm}

\paragraph{Merge-evidence predicate.}
$\textsc{LooksLikeMerge}(a, b, c)$ checks that the child string $c$
inherits distinctive vocabulary from \emph{both} candidate parents,
not just from one.  Let $W(s)$ denote the set of lower-cased
whitespace-separated tokens of length at least four in string $s$
(the length filter removes function words), and define the
``unique-from-$x$ tokens that appear in $c$'' shorthand
\[
U_{x \mid y}(c) \;\triangleq\; \bigl(W(x) \setminus W(y)\bigr) \cap W(c).
\]
$\textsc{LooksLikeMerge}$ is then
\begin{align*}
&\textsc{LooksLikeMerge}(a, b, c) \;\triangleq\; \\
&\quad
  \bigl[U_{a \mid b}(c) \neq \emptyset\bigr]
  \,\land\,
  \bigl[U_{b \mid a}(c) \neq \emptyset\bigr].
\end{align*}
The companion guard $s^\star < 0.95$ on the best similarity inside
Alg.~\ref{alg:diff-schemas} prevents the merge branch from firing
when one parent rule already covers nearly the entire child --- in
that case the edit is more naturally a refinement of that parent
with auxiliary phrasing borrowed from a sibling.

\paragraph{InferOp classification rule.}
With the hierarchical diff record $\mathcal{D}$ and typed target
$\tau$ in hand, $\textsc{InferOp}$ emits one label per changed
location, so $o(c)$
is a multiset over $\mathcal{O}$ rather than a single dominant
label.  This is why a single edit in
Table~\ref{tab:edit-distribution} can carry several labels (e.g.\
\textsc{RefineRule}$+$\textsc{AddRule} when the patch refines one
sub-rule and adds another) and the per-op shares sum above
$100\%$.  $\textsc{InferOp}$ classifies each location independently
by the following rules:
\begin{enumerate}[leftmargin=*,itemsep=0pt,topsep=2pt]
\item Each \emph{merged rule alignment}---two parent rules replaced by
one child rule that shares distinctive vocabulary with both---
contributes a \textsc{MergeRules} label.
\item Each remaining added rule contributes an \textsc{AddRule}
label; each refined rule contributes a \textsc{RefineRule} label;
each dropped rule contributes a \textsc{DropRule} label.
\item Outline-layer changes are classified jointly with the
typed target $\tau$, each contributing one label:
$\tau{=}\texttt{ADD step at <pos>}$ with
$|\alpha_c|{=}|\alpha_p|{+}1$ $\to$ \textsc{AddStep};
$\tau{=}\texttt{SPLIT step <N>}$ aligning one parent step to two
adjacent child steps $\to$ \textsc{SplitStep};
$\tau{=}\texttt{step <N>}$ with exactly one step's top-level text
replaced and step count preserved $\to$ \textsc{RewriteStep};
otherwise (e.g.\ $\tau{=}\texttt{backbone}$ or the diff spans
multiple steps) $\to$ \textsc{Backbone}.
\end{enumerate}
The procedure is deterministic given the parent--child schema pair
and typed target $\tau$; the architect never self-declares any of
these operator labels.

\section{Evidence Packet Construction}
\label{app:evidence-packet}

This appendix complements \S\ref{sec:editing} (\emph{Balanced
evidence packet}) by specifying the slot sizes, weights, and
breadcrumb-rendering template.

\subsection{Slot Sizes and Weights}
\label{app:hyperparams}

The minibatch $M = (M_{\mathrm{focus}}, M_{\mathrm{anchor}},
M_{\mathrm{train}})$ uses three slots with sizes
$n_f{=}2, n_a{=}1, n_{tr}{=}3$, totalling up to six examples; in
fallback iterations without newly broken examples, $M_{\mathrm{focus}}$
is empty.
$M_{\mathrm{focus}}$ samples without replacement from
$\textsc{newly\_broken}(p)$ with weights
$w_i \propto a_i$~\citep{efraimidis2006weighted}, where
\[
a_i \;\triangleq\; |\mathcal{A}|^{-1} \!\sum_{s \in \mathcal{A}}\! \mathbf{c}(s)_i \in [0,1]
\]
is the archive agreement on example $i$.  $M_{\mathrm{anchor}}$
samples the parent's \textsc{newly\_fixed} examples, weighted by
$1-a_i$ toward rare cases few archive members solve; when no
\textsc{newly\_fixed} case exists it falls back to specialist solves
(correct in the parent but in none of its ancestors), then to any
other case in $C(p)$, and
$M_{\mathrm{train}}$ samples uniformly from training-set examples
not already selected into $M_{\mathrm{focus}}$ or
$M_{\mathrm{anchor}}$.
None of the three slots is recoverable from
$\sum_i \mathbf{c}(p)_i$ alone, which motivates explicit
edit-effect lineage conditioning rather than scalar-reward conditioning.

\subsection{Fallback without Regressions}
\label{app:train-only-fallback}

When $\textsc{newly\_broken}(p)$ is empty (cold start
$g{=}\emptyset$, or the parent's last edit caused no regression on
$\mathcal{D}_{\mathrm{tr}}$), $M_{\mathrm{focus}}$ is empty and we fall back to
$M = (\emptyset, M_{\mathrm{anchor}}, M_{\mathrm{train}})$; the
anchor and train slot sizes are unchanged ($n_a{=}1$, $n_{tr}{=}3$).
The architect prompt is told there is no regression focus so that
diagnoses target train probes rather than hallucinating a
non-existent training regression.

\subsection{Breadcrumb Rendering}
\label{app:breadcrumb-rendering}

The breadcrumb $\ell$ is rendered as a short multi-line
block titled \texttt{=== PARENT'S LAST PATCH ===}.  It is
\emph{not} a single per-operator template: instead the renderer
iterates over the grandparent\,$\to$\,parent structural diff and
emits one bullet section per non-empty diff bucket.  A single
breadcrumb can therefore carry multiple operator types if the last
patch made several changes at once (e.g.\ an \textsc{AddRule}
together with a \textsc{RefineRule}).  Each change is anchored to
the parent schema by a content \emph{snippet} (the first
$\sim$$25$ chars of the after-text, or a \texttt{**Bold Title**}
if present), so the architect can locate the change by substring
search rather than by a numeric rule id, which keeps the
\texttt{CURRENT SCHEMA} block free of inline markers.

\paragraph{Section structure.}
The renderer composes the block from the following pieces, in
order:
\begin{enumerate}[leftmargin=*,itemsep=1pt,topsep=2pt]
\item Title line \texttt{=== PARENT'S LAST PATCH ===}.
\item Optional \texttt{Last rationale\,$\to$\,\textellipsis} line
  (truncated to $160$ chars) carrying the previous round's
  attribution rationale.
\item Outline-layer sections (only if the outline diff is
  non-empty): \texttt{Rewritten outline step(s)}, \texttt{New
  outline step(s)}, or a fallback \texttt{Backbone text changed
  (no per-step anchors).}
\item Rule-layer sections, one per non-empty diff bucket:
  \texttt{Refined sub-rule(s)}, \texttt{New sub-rule(s)},
  \texttt{Merged-into sub-rule(s)}, and a tally line
  \texttt{Dropped <K> sub-rule(s) (content removed)}.
\item A final summary line
  \texttt{Effect on train: +<f> newly\_fixed, -<b> newly\_broken}.
\end{enumerate}

\paragraph{Cold-start branch.}
When the parent is the seed schema (no prior patch to learn
from), the renderer emits the fixed message
\texttt{(parent is the seed schema; no prior patch to learn from)}
under the title and skips all diff sections.

\begin{sepoexample}{Breadcrumb example (\textsc{RefineRule} + \textsc{AddRule} in the same edit)}
\footnotesize\ttfamily
=== PARENT'S LAST PATCH ===\\
~~Last rationale $\to$ Worker conflated sarcastic and literal options on step~2; adding an explicit sentiment-mismatch check.\\
~~Refined sub-rule(s):\\
~~~~- in step 2 $\to$ When the option uses positive\ldots\\
~~New sub-rule(s):\\
~~~~- in step 2 $\to$ Compare literal meaning agai\ldots\\
~~Effect on train: +2 newly\_fixed, -1 newly\_broken
\end{sepoexample}

\noindent
Both architect calls (attribution and patch) consume this same
rendered block as the \texttt{\{lineage\_block\}} placeholder in
Appendix~\ref{app:prompt-attribution} and~\ref{app:prompt-patch}.
The \textsc{$-$Breadcrumb} ablation
(\S\ref{sec:q2-mechanism}) replaces the entire block with a single
line that strips everything except the \texttt{Effect on train}
summary; the \textsc{$-$Lineage} ablation removes the block
entirely and uniformises $M_{\mathrm{focus}}$ over the parent's
current errors.

\section{Qualitative Analysis}
\label{app:qualitative}

This appendix illustrates \sepo's schemas across tasks
(\S\ref{app:schema-gallery}) and then grounds the main paper's
auditability claims in the actual optimisation log of one run
(\S\ref{app:worked-trace}--\S\ref{app:failure-modes}; BBH Word
Sorting, seed 45, Llama-3.1-8B-Instruct worker).  All numbers and
text are reconstructed from the single \texttt{result.json} each run
writes by default; no extra instrumentation is required.

\subsection{Cross-Task Schema Examples}
\label{app:schema-gallery}

The schemas \sepo converges to look different across tasks because
their failure modes differ.  Together with the BBH Word Sorting
schema in \S\ref{sec:schema}, the two examples below show that the
two-layer schema architecture supports task-tailored rule
distributions without imposing one: on MBPP+, rules concentrate on
extraction (function-name fidelity, type parsing, constraint
mapping) and implementation; on HotpotQA Medium, rules concentrate
on grounding and entity verification.  Rule text is taken verbatim
from \sepo's final prompts (light line-wrapping only).

\begin{sepoexample}{Example schema (MBPP+, seed 45, test acc.\ 63\%)}
\footnotesize
\textbf{Step 1.} Understand the problem and extract constraints.
\begin{itemize}[label={-},leftmargin=14pt,topsep=1pt,itemsep=0pt]
  \item Identify the required function name from the problem
        description or \texttt{assert} statement (e.g.,
        \texttt{assert func\_name(\ldots)}) and ensure your solution
        defines a function with this exact name.
  \item Parse the input/output types and specific logic rules
        (e.g., ``remove characters at odd indices'', ``sum of digits
        equal to length'') before planning.
  \item Explicitly map the problem's specific logical constraints
        to a concrete algorithm before coding (e.g., for a magic
        square: verify all rows, columns, and diagonals sum to the
        same value; for character replacement: iterate through the
        string and swap specific target characters).
\end{itemize}
\textbf{Step 2.} Plan and implement the solution in code.
\begin{itemize}[label={-},leftmargin=14pt,topsep=1pt,itemsep=0pt]
  \item Write a complete Python function that implements the parsed
        logic exactly as described, ignoring any unrelated patterns
        from training data.
  \item Include the function definition, the logic body, and a test
        call using the provided \texttt{assert} statement to verify
        correctness.
\end{itemize}
\textbf{Step 3.} Output the final answer.
\begin{itemize}[label={-},leftmargin=14pt,topsep=1pt,itemsep=0pt]
  \item If the task asks for a function, output the full Python
        code block containing the function and the test.
  \item Output the final result on the last line as
        ``\texttt{Answer: (X)}'', where X is the return value of
        the function or the code itself if requested.
\end{itemize}
\end{sepoexample}

\begin{sepoexample}{Example schema (HotpotQA Medium, seed 45, test acc.\ 68\%)}
\footnotesize
\textbf{Step 1.} Parse the question and locate the relevant context.
\begin{itemize}[label={-},leftmargin=14pt,topsep=1pt,itemsep=0pt]
  \item Identify the key entities and constraints in the question
        (e.g., specific dates, titles, roles, locations), then scan
        the provided passages for sentences mentioning those exact
        entities.
\end{itemize}
\textbf{Step 2.} Verify, link, and answer.
\begin{itemize}[label={-},leftmargin=14pt,topsep=1pt,itemsep=0pt]
  \item Strictly ground your answer in the provided text; do not use
        outside knowledge or hallucinate details not present in the
        context.
  \item Verify entity matches before extracting attributes: if the
        question asks about Entity A, do not extract facts about
        Entity B even if they appear in the same passage or share a
        category.
  \item For multi-hop questions, explicitly trace the chain across
        passages (e.g., Entity A in passage 1 to Property B in
        passage 2); once the answer is reached, stop reasoning
        immediately and do not repeat the same logical loop.
\end{itemize}
\textbf{Step 3.} Output the final answer.
\begin{itemize}[label={-},leftmargin=14pt,topsep=1pt,itemsep=0pt]
  \item The final line must be exactly ``\texttt{Answer: (X)}''
        where X is the concise value with no introductory phrases,
        explanations, or repetitions of the question.
\end{itemize}
\end{sepoexample}

\subsection{Worked Iteration Trace: Discovering an Anti-Verification Rule}
\label{app:worked-trace}

The log records a two-iteration sequence in which \sepo
\emph{adds} a self-verification rule, observes it triggering worker
repetition loops, and \emph{removes} it---replacing it with the
opposite instruction.  The final schema is the winner shown in
Figure~\ref{fig:edit-tree}.

\paragraph{Iteration 26 --- adding self-verification.}
The breadcrumb from the parent (\texttt{fa5513}, $31/50$ correct on
$\mathcal{D}_{\mathrm{tr}}$) reports a \textsc{RewriteStep}$+$\textsc{AddRule} edit on
\texttt{step\,3} with example-level effect $(+17,{-}6)$.  The
evidence packet's focus slot draws two newly broken training
examples on which the worker \emph{has} the right sorting logic but
emits a typo: \textbf{Ex \#33} sorts \texttt{skat} before
\texttt{scandalous} in its reasoning yet writes the reversed order
on the answer line; \textbf{Ex \#32} shows the same on
\texttt{castrate}/\texttt{cancelled}.  The attribution architect
returns $\tau{=}\texttt{step\,2}$ with the reason ``the worker
ignores its own verification rule, leading to sorting typos despite
correct logic.''  The patch architect adds one CRITICAL bullet under
step\,2:

\begin{quote}\small\itshape
Before finalizing, perform a strict side-by-side verification:
compare your sorted list against the original input word-for-word;
if the order of any two adjacent words differs from their
alphabetical order, immediately re-sort.
\end{quote}

\noindent The post-hoc operator is \textsc{AddRule}; the child
(\texttt{aebcc2}, $32/50$) is admitted by the score-above-parent
rule of Eq.~\ref{eq:admission}.

\paragraph{Iteration 32 --- removing self-verification.}
The breadcrumb from \texttt{aebcc2} reports the iter-26
\textsc{AddRule}.  The focus slot now draws two examples that the
iter-26 child \emph{newly broke}: \textbf{Ex \#40} enters an
infinite repetition loop, hallucinating the word \texttt{anaheim}
hundreds of times in place of the sort, and \textbf{Ex \#7} outputs
an incorrectly sorted list (\texttt{wiley} before \texttt{whimper})
while falsely claiming the verification passed.  The attribution
architect again localises to $\tau{=}\texttt{step\,2}$ with the
reason ``the CRITICAL verification sub-rule is ineffective and
triggering loops; the model cannot execute self-verification without
hallucinating.''  The patch architect \emph{removes} both the
side-by-side verification and the confirmation
bullets---both are forms of post-sort re-processing---and adds an
anti-re-processing rule:

\begin{quote}\small\itshape
CRITICAL: after sorting, do not re-process the list (no re-reading or
side-by-side checks)---both trigger hallucination loops on this worker;
output immediately.
\end{quote}

\noindent Crucially the edit is \emph{not} a narrow patch on one signal:
its evidence packet draws \emph{both} the focus regressions
(\texttt{Ex\,\#40,\#7}, fixed on $\mathcal{D}_{\mathrm{tr}}$) and the train probe
(\texttt{Tr\,\#44,\#49,\#35}, the $M_{\mathrm{train}}$ slot), so the same
``stop re-processing'' principle addresses the parent's regression and
the train errors together.  The aggregate
training score stays at $32/50$, but the child (\texttt{7b0643})
solves a training example no other archive member solves; it is
admitted by the \textbf{unique-contribution} branch (a) of
Eq.~\ref{eq:admission}---a typed, Pareto-style admission a scalar reward
cannot express.  This rule (``do \emph{not} re-process'') contradicts
standard chain-of-thought~\citep{wei2022chain} practice and is
reproducibly discoverable \emph{only} because the per-example
$(\textsc{newly\_broken},\textsc{newly\_fixed})$ transitions in the
evidence packet expose it; whole-prompt rewriters with reward-only
feedback have no comparable mechanism.

\clearpage
\twocolumn[%
  {\centering\includegraphics[width=\textwidth]{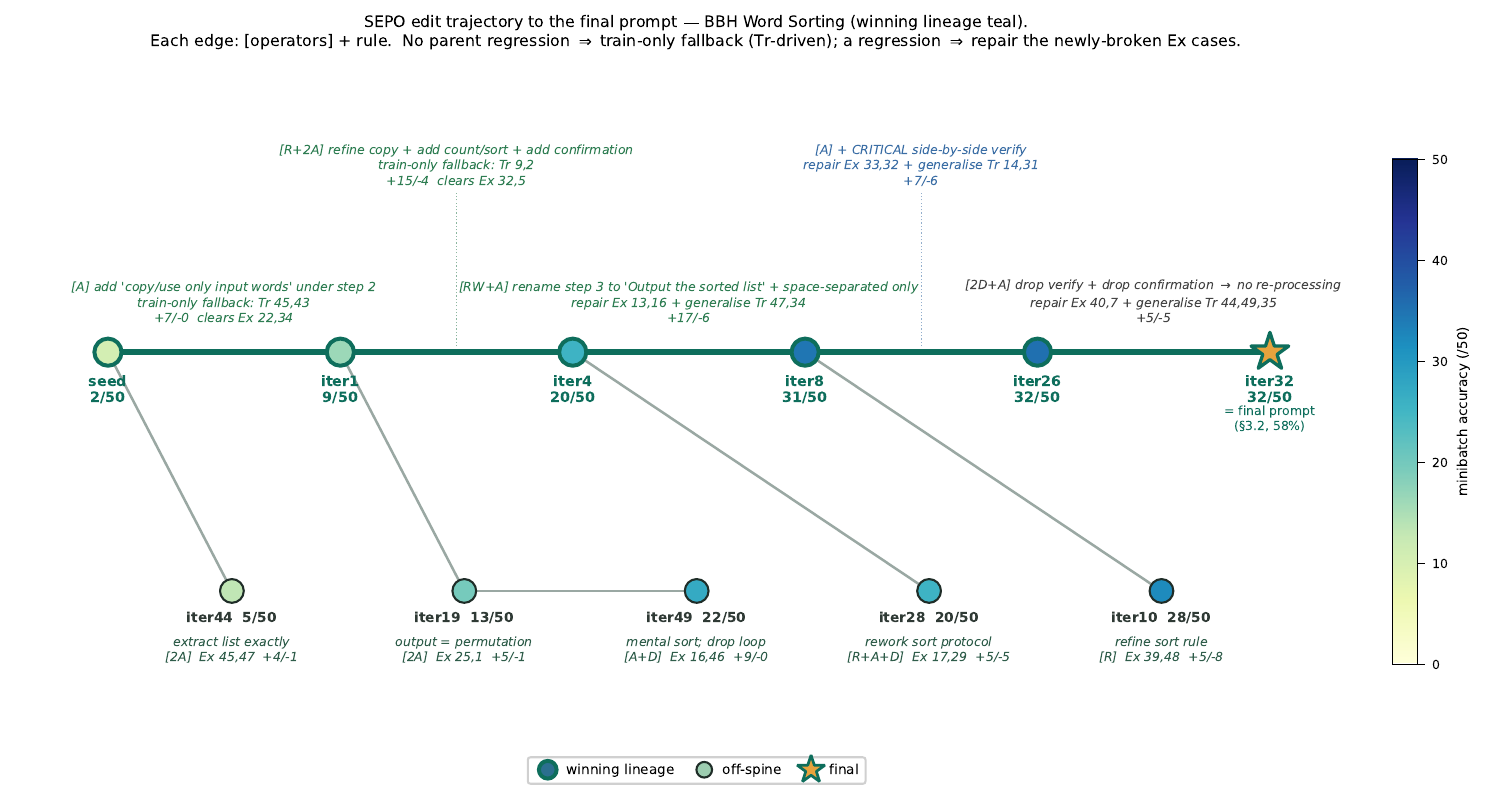}\par}
  \vspace{0.5em}
  {\small\refstepcounter{figure}\label{fig:edit-tree}\noindent
  \textbf{Figure~\thefigure: Edit trajectory to the final
  prompt} (BBH Word Sorting, seed 45).  The \textcolor{teal}{teal spine}
  is the winning lineage to the \S\ref{sec:schema} prompt.  Each edge
  shows its operator(s) (\textsc{A}=Add, \textsc{R}=Refine,
  \textsc{D}=Drop, \textsc{RW}=RewriteStep) and the rule changed; the
  accumulated rules at iter\,32 are exactly the \S\ref{sec:schema}
  prompt.  The train probe (\texttt{Tr}, the $M_{\mathrm{train}}$ slot)
  is in \emph{every} packet, giving two real modes: when the parent left
  \emph{no} regression the edit is a \emph{no-regression fallback} driven by
  \texttt{Tr} (and still lifts $\mathcal{D}_{\mathrm{tr}}$); when the parent
  \emph{broke} cases the edit \emph{repairs} those newly-broken
  training cases (\texttt{Ex}) \emph{and} lets \texttt{Tr} steer
  generalisation.  Lower branches are off-spine accepted schemas.  Iter\,4$\to$8's
  \textsc{RW} renames step\,3 to \S\ref{sec:schema}'s task-specific
  ``Output the sorted list'' heading; the modes, the train probe, every
  other operator, and the \texttt{Ex} repairs are grounded in the log.  Node
  colour $=$ accuracy; $+$fixed$/{-}$broken on $\mathcal{D}_{\mathrm{tr}}$.  Per-edit
  detail is in the text.\par}
  \vspace{0.9em}
]
\subsection{Edit Tree from Seed to Final Winner (BBH Word Sorting)}
\label{app:case-study-wordsort}

Figure~\ref{fig:edit-tree} shows the path to \sepo's
final Word Sorting prompt (\S\ref{sec:schema}).  We walk through each edit
below: its operation, the failing cases that motivated it, and---from
the per-example scores---whether it repaired the parent edit's
regression or a train-probe failure.

\paragraph{Per-edit walkthrough (winning lineage).}
For each accepted edit we give the operator(s) (\textsc{A}=AddRule,
\textsc{R}=RefineRule, \textsc{D}=DropRule, \textsc{RW}=RewriteStep) and
the rule changed; the rules accumulated at iter\,32 are exactly the
\S\ref{sec:schema} prompt.  The train probe (\texttt{Tr}, the
$M_{\mathrm{train}}$ slot) is in \emph{every} packet (\S\ref{sec:editing}),
giving two modes: when the parent left \emph{no} $\mathcal{D}_{\mathrm{tr}}$
regression the edit is a \emph{no-regression fallback} driven by
\texttt{Tr}; when the parent \emph{broke} cases the edit \emph{repairs}
those newly-broken \texttt{Ex} cases \emph{and} is steered by
\texttt{Tr} to generalise.  \texttt{Ex} repairs are verified on
$\mathcal{D}_{\mathrm{tr}}$, and the modes and train probe are grounded in the log;
iter\,4$\to$8's \textsc{RW} renames step\,3 to \S\ref{sec:schema}'s
task-specific ``Output the sorted list'' heading.
\begin{itemize}\setlength{\itemsep}{2pt}
\item \textbf{seed$\,\to\,$iter\,1} ($2\!\to\!9$),
  \textbf{\textsc{A}}:
  add ``copy/use only the input words'' rule under step\,2.
  \emph{Train-only fallback} (seed broke nothing): driven by the train
  probe \texttt{Tr\,\#45,\#43}; the rule generalises to clear the
  persistent hallucination failures \texttt{Ex\,\#22,\#34}.  \emph{Effect:}
  $+7/{-}0$.
\item \textbf{iter\,1$\,\to\,$iter\,4} ($9\!\to\!20$),
  \textbf{\textsc{R}$+$2\textsc{A}}: refine the copy rule (``do not import
  from memory''), add ``count the input words once, then sort in strict
  A--Z order in a single pass,'' and add a confirmation rule
  (``re-state the full sorted list to check that no word is dropped'').
  \emph{Train-only fallback} (iter\,1 broke nothing): \texttt{Tr\,\#9,\#2}
  drives it; the confirmation catches the hallucinated / dropped-word
  failures \texttt{Ex\,\#32,\#5}.  \emph{Effect:} $+15/{-}4$.
\item \textbf{iter\,4$\,\to\,$iter\,8} ($20\!\to\!31$),
  \textbf{\textsc{RW}$+$\textsc{A}} on step\,3: rename step\,3 heading
  to ``Output the sorted list'' and add ``output one space-separated
  line, no commas/newlines'' under step\,3.  \emph{Repairs} iter\,4's
  regression \texttt{Ex\,\#16,\#13} (correct sort, wrong format);
  anchor \texttt{Ex\,\#17} (already correct sort + correct format) is
  preserved by the rule being additive rather than overwriting, with
  \texttt{Tr\,\#47,\#34} steering generalisation.  \emph{Effect:}
  $+17/{-}6$.
\item \textbf{iter\,8$\,\to\,$iter\,26} ($31\!\to\!32$),
  \textbf{\textsc{A}}: add ``\textsc{critical}: side-by-side verify''
  (an \emph{attempt} to fix sort typos).  \emph{Repairs} iter\,8's
  regression \texttt{Ex\,\#33,\#32} (reversed output), with
  \texttt{Tr\,\#14,\#31} steering generalisation.  \emph{Effect:}
  $+7/{-}6$.
\item \textbf{iter\,26$\,\to\,$iter\,32} ($32\!\to\!32$, final),
  \textbf{2\textsc{D}$+$\textsc{A}}: drop both the side-by-side
  verification rule and the confirmation rule (both are forms of
  post-sort re-processing) and add ``\textsc{critical}: do not
  re-process; trust the single-pass sort.''  \emph{Repairs} iter\,26's
  own regression \texttt{Ex\,\#40,\#7} (the verify rule's loop /
  false-verify damage; the confirmation rule also re-reads the list
  and is subsumed by the same anti-re-process rule), with
  \texttt{Tr\,\#44,\#49,\#35} steering generalisation.
  \emph{Effect:} $+5/{-}5$; removes the loop pathology (final test
  $58\%$).
\end{itemize}
So \sepo{} never stalls: with a regression it surgically repairs the
newly-broken \texttt{Ex} cases while the train probe steers
generalisation; with none it keeps improving in no-regression fallback---a
dual mode whole-prompt rewriters lack.  The operators span
\textsc{RW}/\textsc{R}/\textsc{A}/\textsc{D}, and the accumulated rules
at iter\,32 reproduce the \S\ref{sec:schema} prompt exactly; the modes,
the train probe, the \texttt{Ex} repairs, and the
\textsc{RW}/\textsc{R}/\textsc{A}/\textsc{D} operators are grounded in the
log, with iter\,4$\to$8's \textsc{RW} renaming step\,3 to the
task-specific ``Output the sorted list'' heading.

\subsection{Failure-Mode Analysis}
\label{app:failure-modes}

\sepo's Word Sorting schema repairs the dominant failure modes
(entity hallucination, sort-method loops, output-format collapse)
but its effect saturates with input length.
Word Sorting uses the same $50$-train / held-out-test convention as
the other BBH tasks, with a disjoint $200$-example held-out test
split.  Table~\ref{tab:wordsort-length} bins a $100$-example
diagnostic subset of that held-out test split by the number of words
to be sorted: short lists ($\leq 5$ words) reach $79\%$ while long
lists ($\geq 16$ words) collapse to $31\%$.  The degradation is
monotone and held across every schema variant we examined.  The cause
is that the schema's discipline is
\emph{instructional}, not \emph{algorithmic}: rules such as
``copy the words exactly'' and ``trust the initial sort'' raise the
probability of a correct single pass but do not give the worker an
external scratchpad, so the per-token error of an $n$-word sort still
compounds with $n$.  This delineates the scope of prompt-only
structural editing: it repairs systematic, length-independent
failure modes (format, hallucination, verification loops) but cannot
substitute for an algorithmic execution mechanism on long inputs.

\begin{table}[!ht]
\centering
\small
\setlength{\tabcolsep}{6pt}
\renewcommand{\arraystretch}{1.10}
\arrayrulecolor{tableRule}
\begin{tabular}{@{}lrr@{}}
\toprule
List length (words) & $n$ & Diagnostic acc. \\
\midrule
$\leq 5$    & 24 & $0.79$ \\
$6$--$10$   & 30 & $0.70$ \\
$11$--$15$  & 20 & $0.40$ \\
$\geq 16$   & 26 & $0.31$ \\
\midrule
\rowcolor{tableMean}
Diagnostic subset & 100 & $0.56$ \\
\bottomrule
\end{tabular}
\arrayrulecolor{black}
\caption{Held-out accuracy of \sepo's BBH Word Sorting schema
(\S\ref{sec:schema}) binned by sort length on a
$100$-example diagnostic subset drawn from the $200$-example held-out
test split with the Llama-3.1-8B-Instruct worker.
Accuracy degrades monotonically with list length, the signature of
instructional (not algorithmic) repair.}
\label{tab:wordsort-length}
\end{table}

\section{Baseline Configurations}
\label{app:baselines}

For each (method, benchmark) cell in the main comparison
experiments, we run three independent optimisation seeds.  All
baselines share the worker LLM, dataset splits, three-seed
schedule, and the $\texttt{max\_metric\_calls}{=}2000$ metric-call
budget.  Architect choices, hyperparameters, and
budget-normalisation details follow below.

\paragraph{\textsc{Seed} (unoptimised).}
The cold-start seed schema $s_0$ of Appendix~\ref{app:seed-schema},
deployed without any optimisation.  Reported for reference.

\paragraph{\textsc{OPRO}.}
Architect: Qwen3.5-397B-A17B.  Architect temperature $1.0$,
candidates per step $5$, optimiser steps $40$, optimisation
trajectory window $20$.  Architect prompts follow the public
reference implementation.

\paragraph{\textsc{APSF}.}
Architect: Qwen3.5-397B-A17B.  Number of factors $5$, candidates per
factor $3$, factor selection greedy on the training slice.  We
use the public reference implementation with no other changes.

\paragraph{\textsc{MIPROv2}.}
DSPy v2.5; architect Qwen3.5-397B-A17B; \texttt{auto="light"}
configuration ($n_{\mathrm{init}}{=}8$ bootstrap demonstrations,
$n_{\mathrm{trials}}{=}20$).  Bayesian optimiser uses the default
prior.

\paragraph{\textsc{MPO}~\citep{sharma2026modular}.}
Architect: Qwen3.5-397B-A17B.  We use the released hyperparameters
($T{=}40$ optimisation iterations, $b{=}5$ candidates per iteration,
trajectory window $10$).

\paragraph{\textsc{GEPA}.}
Architect: Qwen3.5-397B-A17B.  Pareto-front archive with capacity
$32$; reflection prompt and reflective sample size $3$ follow the
released configuration.  We rerun the released code with the same
metric-call budget as \sepo.

\section{Extended Experimental Results}
\label{app:extended-results}

This appendix supplements \S\ref{sec:experiments} with additional
Llama runs on high-variance tasks, a matched HotpotQA-Hard sensitivity
test, per-method and per-benchmark token-cost decompositions, and
budget-matched accuracy curves.

\subsection{Additional Runs}
\label{app:additional-runs}

We also run three additional paired Llama optimisations on five tasks
with high run-to-run variance. Across these $15$ added comparisons,
\sepo wins $12$, and its six-run mean remains higher on all five tasks
(Table~\ref{tab:sepo-additional-runs}).

\begin{strip}
\centering
\scriptsize
\setlength{\tabcolsep}{3.2pt}
\renewcommand{\arraystretch}{1.12}
\arrayrulecolor{tableRule}
\resizebox{\textwidth}{!}{%
\begin{tabular}{@{}lccccccc@{}}
\toprule
Task &
\shortstack{Reported 3 runs\\mean $\pm$ SD (S/G)} &
Seed 46 (S/G) & Seed 47 (S/G) & Seed 48 (S/G) &
\shortstack{All 6 runs\\mean $\pm$ SD (S/G)} &
6-run $\Delta$ & Added W/T/L \\
\midrule
Causal Judgement &
$61.1{\pm}4.6/59.9{\pm}3.8$ & $64.2/62.8$ & $55.5/56.9$ & $57.7/56.9$ &
$60.1{\pm}4.2/59.4{\pm}3.3$ & $+0.7$ & $2/0/1$ \\
Date Understanding &
$76.5{\pm}4.3/75.3{\pm}6.1$ & $82.0/80.0$ & $76.5/82.0$ & $77.5/74.0$ &
$77.6{\pm}3.5/77.0{\pm}5.0$ & $+0.6$ & $2/0/1$ \\
Disambiguation QA &
$68.8{\pm}4.6/64.5{\pm}6.6$ & $74.5/74.0$ & $68.5/63.5$ & $67.5/61.0$ &
$69.5{\pm}3.8/65.3{\pm}6.1$ & $+4.2$ & $3/0/0$ \\
Logical Deduction &
$47.3{\pm}2.1/47.0{\pm}3.9$ & $49.5/45.5$ & $44.0/51.0$ & $46.5/41.5$ &
$47.0{\pm}2.2/46.5{\pm}3.9$ & $+0.5$ & $2/0/1$ \\
Word Sorting &
$58.2{\pm}1.3/39.3{\pm}18.3$ & $59.5/56.0$ & $57.0/31.0$ & $58.0/31.5$ &
$58.2{\pm}1.1/39.4{\pm}14.7$ & $+18.8$ & $3/0/0$ \\
\bottomrule
\end{tabular}}
\arrayrulecolor{black}
\captionof{table}{Additional paired Llama runs on five high-variance tasks.
S/G denotes \sepo/\textsc{GEPA}; the W/T/L column counts only the
three added seeds. Accuracies and $\Delta$ are in percentage points.}
\label{tab:sepo-additional-runs}
\end{strip}

\subsection{Matched HotpotQA-Hard Sensitivity}
\label{app:hqa-hard-sensitivity}

To test whether the Llama HotpotQA split choice changes the method
ranking, we evaluate the compared methods below on the same $400$
held-out Hard examples.

\begin{table}[!ht]
\centering
\small
\setlength{\tabcolsep}{9pt}
\renewcommand{\arraystretch}{1.08}
\arrayrulecolor{tableRule}
\begin{tabular}{@{}lr@{}}
\toprule
Method & HQA-Hard EM (\%) \\
\midrule
Seed           & 8.00 \\
\textsc{OPRO}  & 24.00 \\
\textsc{APSF}  & 26.00 \\
\textsc{MPO}   & 28.25 \\
\textsc{GEPA}  & 28.75 \\
\rowcolor{llamaBlock}
\sepo           & \textbf{29.25} \\
\bottomrule
\end{tabular}
\arrayrulecolor{black}
\caption{Matched HotpotQA-Hard sensitivity test with the
Llama-3.1-8B-Instruct worker on $400$ held-out examples.}
\label{tab:hqa-hard-sensitivity}
\end{table}

Replacing HotpotQA-Medium with Hard in the Llama 14-task macro changes
\sepo from $61.9\%$ to $59.0\%$ and \textsc{GEPA} from $58.8\%$ to
$55.9\%$; \sepo therefore remains first, with the same $3.1$\,pp lead.
The compressed $24.0$--$29.25\%$ range reflects limited answer
accessibility: with the fixed seed prompt, pass@1 is $8.0\%$, oracle
pass@64 is $31.0\%$, and $69.0\%$ of examples remain incorrect in all
$64$ samples. We use Medium under Llama to retain a more discriminative
optimiser comparison.

\subsection{Per-Benchmark Token Cost}
\label{app:cost-extra}

\paragraph{List prices used in Table~\ref{tab:sepo-efficiency}.}
The \textbf{\$/cell} and $\rho_{\$}$ columns of
Table~\ref{tab:sepo-efficiency} are computed from public list prices
as of 2026-05-26.  The architect (Qwen3.5-397B-A17B) is priced at the
SiliconFlow public rate of \textyen$1.20$\,/\,M input and
\textyen$7.20$\,/\,M output tokens (0--128k context tier); the worker
(Llama-3.1-8B-Instruct) is priced at the OpenRouter listing of
\$0.02\,/\,M input and \$0.05\,/\,M output tokens.  CNY amounts are
converted at the 2026-05-26 spot rate of 1\,USD$=$6.79\,CNY, yielding
architect prices of \$0.177\,/\,M input and \$1.060\,/\,M output.
For method $m$, we compute
\begin{equation}
  C_m = \sum_{x\in\{A,W_{\mathrm{opt}},W_{\mathrm{dep}}\}}
  \left(p_x^{\mathrm{in}}T_{x,m}^{\mathrm{in}}+
  p_x^{\mathrm{out}}T_{x,m}^{\mathrm{out}}\right),
  \label{eq:dollar-cost}
\end{equation}
Here $x$ indexes the architect, worker-optimisation, and
worker-deployment channels; $p$ and $T$ denote the corresponding list
price and measured token count.  Each cell includes one held-out
deployment of the selected prompt; we average cell costs across seeds
and tasks.  Input/output separation matters because architect output
is approximately six times as expensive as input at this snapshot.
For \textsc{MPO}, totals are measured but the unavailable I/O split is
estimated from \textsc{GEPA}, so only its dollar estimate depends on
this approximation.  The token-based $\rho$ is provider-independent,
whereas $\rho_{\$}{=}\Delta\text{Acc}/C_m$ covers all three billed
channels; dollar results are therefore a dated price snapshot.

\begin{table*}[tp]
\centering
\footnotesize
\setlength{\tabcolsep}{4pt}
\renewcommand{\arraystretch}{1.3}
\arrayrulecolor{tableRule}
\begin{tabularx}{\textwidth}{@{}lRRRRRR@{}}
\toprule
 & \multicolumn{2}{c}{Architect (M tok)} & \multicolumn{2}{c}{Worker-Opt (M tok)} & \multicolumn{2}{c}{Worker-Dep (k tok)} \\
\cmidrule(lr){2-3}\cmidrule(lr){4-5}\cmidrule(lr){6-7}
Method & Input & Output & Input & Output & Input & Output \\
\midrule
\textsc{OPRO}    & 6.44 & 0.40 & 2.04 & 1.07 & 134.7 & 109.3 \\
\textsc{MIPROv2} & 3.32 & 0.20 & 1.64 & 0.96 & 105.0 & 103.4 \\
\textsc{APSF}    & 0.69 & 0.00 & 1.33 & 0.80 &  78.9 &  78.9 \\
\textsc{MPO}$^{\dagger}$ & 0.17 & 0.02 & 2.68 & 0.72 & 198.3 &  62.7 \\
\textsc{GEPA}    & 0.20 & 0.03 & 3.06 & 0.82 & 321.9 & 101.8 \\
\rowcolor{llamaBlock}
\sepo            & 0.32 & 0.02 & 1.61 & 0.98 & 108.8 &  94.4 \\
\bottomrule
\end{tabularx}
\arrayrulecolor{black}
\caption{Per-method input/output token decomposition (M / k tokens
per cell, mean over 3 seeds, Llama-3.1-8B-Instruct 14-task suite).  Row sums
match the corresponding columns of Table~\ref{tab:sepo-efficiency}.
$^{\dagger}$\textsc{MPO}'s I/O split is estimated from \textsc{GEPA};
only its row totals are measured.}
\label{tab:perbench-io-aggregate}
\end{table*}

\begin{table*}[tp]
\centering
\scriptsize
\setlength{\tabcolsep}{3pt}
\renewcommand{\arraystretch}{1.3}
\arrayrulecolor{tableRule}
\begin{tabularx}{\textwidth}{@{}l*{15}{R}@{}}
\toprule
\multicolumn{1}{@{}l}{\textbf{Method}} & \multicolumn{8}{c}{\textbf{BBH}} & \multicolumn{1}{c}{\textbf{Code}} & \multicolumn{3}{c}{\textbf{MMLU-Pro}} & \multicolumn{2}{c}{\textbf{Reasoning}} & \multicolumn{1}{c@{}}{\textbf{Mean}} \\
\cmidrule(lr){2-9}\cmidrule(lr){10-10}\cmidrule(lr){11-13}\cmidrule(lr){14-15}\cmidrule(l){16-16}
& Caus. & Date & Disamb. & Fall. & Deduct. & Obj. & Snarks & Word. & MBPP+ & CS & Econ. & Psych. & GSM-H & HQA & Macro \\
\midrule
\textsc{OPRO}    & 21.22 & 4.91 & 9.68 & 19.47 & 13.91 & 19.56 & 3.14 & 0.46 & 0.13 & 0.71 & 0.63 & 1.76 & 0.04 & 0.08 & 6.84 \\
\textsc{MIPROv2} & 12.05 & 5.71 & 5.23 & 1.07 & 1.44 & 5.38 & 3.02 & 0.53 & 1.34 & 5.46 & 2.92 & 4.61 & 0.18 & 0.28 & 3.52 \\
\textsc{APSF}    & 1.57 & 1.67 & 0.28 & 0.95 & 0.89 & 0.34 & 0.14 & 0.91 & 0.28 & 0.61 & 0.57 & 0.24 & 0.61 & 0.58 & 0.69 \\
\textsc{MPO}     & 0.32 & 0.30 & 0.25 & 0.27 & 0.23 & 0.30 & 0.22 & 0.21 & 0.09 & 0.10 & 0.13 & 0.10 & 0.04 & 0.07 & 0.19 \\
\textsc{GEPA}    & 0.24 & 0.23 & 0.15 & 0.54 & 0.52 & 0.18 & 0.13 & 0.37 & 0.68 & 0.07 & 0.07 & 0.04 & 0.03 & 0.00 & 0.23 \\
\rowcolor{llamaBlock}
\sepo            & 0.38 & 0.50 & 0.28 & 0.58 & 0.89 & 0.32 & 0.28 & 0.59 & 0.10 & 0.18 & 0.17 & 0.13 & 0.07 & 0.30 & 0.34 \\
\bottomrule
\end{tabularx}
\arrayrulecolor{black}
\caption{Per-benchmark \emph{architect} token cost (millions of
tokens per cell, mean over 3 seeds).  The \textsc{Mean} row matches
the \emph{Arch.} column of Table~\ref{tab:sepo-efficiency}.}
\label{tab:perbench-architect}
\end{table*}

\begin{table*}[tp]
\centering
\scriptsize
\setlength{\tabcolsep}{3pt}
\renewcommand{\arraystretch}{1.3}
\arrayrulecolor{tableRule}
\begin{tabularx}{\textwidth}{@{}l*{15}{R}@{}}
\toprule
\multicolumn{1}{@{}l}{\textbf{Method}} & \multicolumn{8}{c}{\textbf{BBH}} & \multicolumn{1}{c}{\textbf{Code}} & \multicolumn{3}{c}{\textbf{MMLU-Pro}} & \multicolumn{2}{c}{\textbf{Reasoning}} & \multicolumn{1}{c@{}}{\textbf{Mean}} \\
\cmidrule(lr){2-9}\cmidrule(lr){10-10}\cmidrule(lr){11-13}\cmidrule(lr){14-15}\cmidrule(l){16-16}
& Caus. & Date & Disamb. & Fall. & Deduct. & Obj. & Snarks & Word. & MBPP+ & CS & Econ. & Psych. & GSM-H & HQA & Macro \\
\midrule
\textsc{OPRO}    & 3.14 & 3.78 & 2.19 & 5.40 & 4.93 & 3.09 & 1.48 & 1.70 & 1.35 & 3.68 & 2.81 & 2.92 & 2.47 & 4.65 & 3.11 \\
\textsc{MIPROv2} & 3.10 & 3.11 & 1.42 & 4.37 & 4.96 & 2.23 & 1.20 & 1.17 & 1.14 & 2.96 & 2.36 & 2.08 & 2.05 & 4.28 & 2.60 \\
\textsc{APSF}    & 1.08 & 2.02 & 0.49 & 3.76 & 2.48 & 1.08 & 0.30 & 4.56 & 1.16 & 2.68 & 2.20 & 2.05 & 1.25 & 4.69 & 2.13 \\
\textsc{MPO}     & 3.98 & 3.76 & 2.56 & 5.20 & 5.71 & 2.94 & 2.13 & 4.33 & 1.52 & 2.60 & 3.16 & 2.39 & 2.28 & 4.99 & 3.40 \\
\textsc{GEPA}    & 5.62 & 4.11 & 4.28 & 6.95 & 5.59 & 3.94 & 3.08 & 6.06 & 1.64 & 3.80 & 3.17 & 2.54 & 2.16 & 1.32 & 3.88 \\
\rowcolor{llamaBlock}
\sepo            & 3.03 & 2.84 & 1.50 & 3.76 & 3.44 & 1.93 & 1.48 & 2.64 & 0.95 & 2.76 & 2.24 & 2.14 & 3.26 & 4.26 & 2.59 \\
\bottomrule
\end{tabularx}
\arrayrulecolor{black}
\caption{Per-benchmark \emph{worker-optimisation} token cost
(millions of tokens per cell, mean over 3 seeds). The
\textsc{Mean} row matches the \emph{Work.} column of
Table~\ref{tab:sepo-efficiency}.}
\label{tab:perbench-worker-opt}
\end{table*}

\begin{table*}[tp]
\centering
\scriptsize
\setlength{\tabcolsep}{3pt}
\renewcommand{\arraystretch}{1.3}
\arrayrulecolor{tableRule}
\begin{tabularx}{\textwidth}{@{}l*{15}{R}@{}}
\toprule
\multicolumn{1}{@{}l}{\textbf{Method}} & \multicolumn{8}{c}{\textbf{BBH}} & \multicolumn{1}{c}{\textbf{Code}} & \multicolumn{3}{c}{\textbf{MMLU-Pro}} & \multicolumn{2}{c}{\textbf{Reasoning}} & \multicolumn{1}{c@{}}{\textbf{Mean}} \\
\cmidrule(lr){2-9}\cmidrule(lr){10-10}\cmidrule(lr){11-13}\cmidrule(lr){14-15}\cmidrule(l){16-16}
& Caus. & Date & Disamb. & Fall. & Deduct. & Obj. & Snarks & Word. & MBPP+ & CS & Econ. & Psych. & GSM-H & HQA & Macro \\
\midrule
\textsc{OPRO}    & 84.7 & 268.0 & 180.8 & 226.4 & 319.0 & 209.2 & 86.0 & 65.7 & 117.0 & 287.6 & 234.5 & 241.7 & 380.6 & 714.7 & 244.0 \\
\textsc{MIPROv2} & 83.5 & 233.4 & 123.6 & 186.3 & 324.0 & 182.2 & 67.7 & 46.7 & 98.9 & 225.6 & 195.3 & 175.9 & 316.1 & 659.0 & 208.4 \\
\textsc{APSF}    & 36.0 & 137.9 & 36.9 & 144.4 & 164.8 & 77.6 & 14.9 & 177.7 & 87.3 & 183.5 & 157.8 & 154.2 & 175.8 & 660.6 & 157.8 \\
\textsc{MPO}     & 108.1 & 277.2 & 223.1 & 216.4 & 359.1 & 251.9 & 123.5 & 168.1 & 132.4 & 206.8 & 264.7 & 203.7 & 350.9 & 767.7 & 261.0 \\
\textsc{GEPA}    & 208.0 & 413.4 & 535.7 & 337.9 & 447.8 & 468.3 & 236.9 & 295.6 & 153.1 & 506.8 & 424.1 & 500.8 & 741.6 & 661.3 & 423.7 \\
\rowcolor{llamaBlock}
\sepo            & 92.7 & 207.0 & 128.4 & 161.4 & 260.7 & 160.6 & 83.0 & 117.6 & 81.3 & 219.8 & 185.9 & 181.4 & 263.7 & 701.0 & 203.2 \\
\bottomrule
\end{tabularx}
\arrayrulecolor{black}
\caption{Per-benchmark \emph{worker-deployment} token cost
(thousands of tokens per cell, mean over 3
seeds). Deployment cost is the average tokens per worker
call multiplied by the test-set size. HotpotQA's seed cost is
dominated by retrieval context length, which inflates all methods
uniformly. The \textsc{Mean} row matches the \emph{Dep.} column of
Table~\ref{tab:sepo-efficiency}.}
\label{tab:perbench-worker-deploy}
\end{table*}

Table~\ref{tab:perbench-io-aggregate} reports the aggregate
input/output split of each cost axis per method (one row per method,
no per-benchmark breakdown).
Tables~\ref{tab:perbench-architect}--\ref{tab:perbench-worker-deploy}
then decompose each axis benchmark by benchmark.  Together the four
tables let any downstream dollar-cost computation be done at the
level of (method, axis, I/O channel), and let reviewers verify that
the per-method aggregates match the main-paper
Table~\ref{tab:sepo-efficiency} both in total (cross-checked in the
\textsc{Mean} rows of the per-benchmark tables) and in I/O
composition.  Worker-optimisation and deployment tokens are reported
per evaluation run; architect tokens are reported as measured.

\subsection{Per-Benchmark Budget-Matched Curves}
\label{app:metric-call-curves}

\begin{figure*}[tp]
\centering
\includegraphics[width=\textwidth]{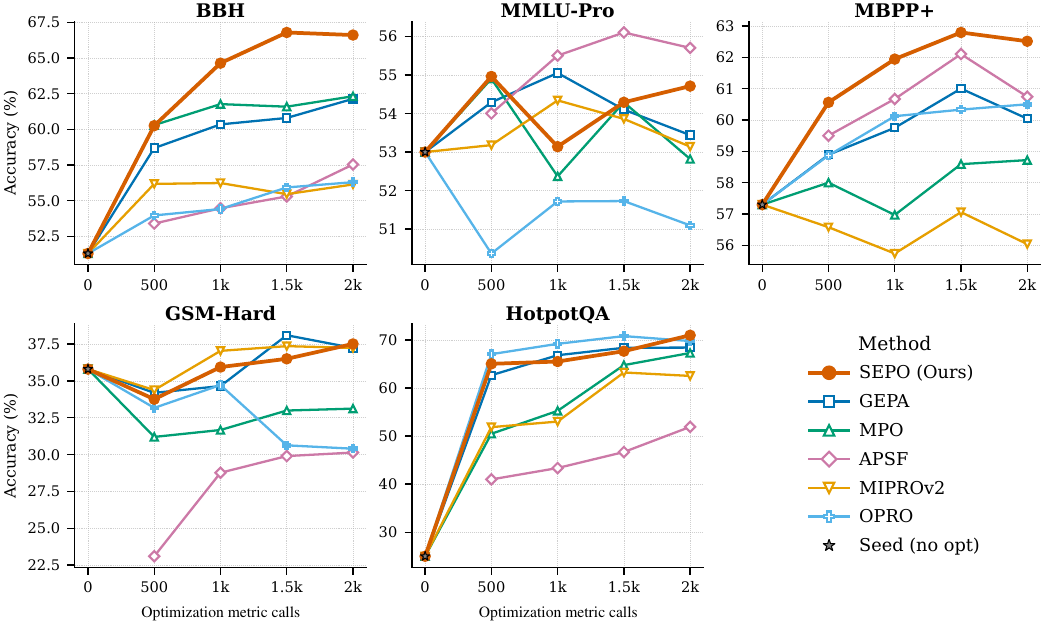}
\caption{\textbf{Per-benchmark budget-matched accuracy curves}
(Llama-3.1-8B-Instruct, held-out).  For each benchmark family --- BBH and
MMLU-Pro shown as the mean over their sub-tasks, and MBPP+,
GSM-Hard, and HotpotQA shown individually --- we plot the held-out
accuracy of every method's stored best-so-far prompt at metric-call
checkpoints $0$ (seed), $500$, $1000$, $1500$, and $2000$, under a
shared evaluation protocol.  \sepo (highlighted) leads consistently on
BBH and MBPP+, remains competitive on MMLU-Pro, GSM-Hard, and HotpotQA,
and yields the strongest macro-level trajectory.}
\label{fig:rollout-perdomain}
\end{figure*}

Figure~\ref{fig:rollout-perdomain} extends the Q1 macro headline of
the main paper to the budget axis: for every benchmark family we
re-evaluate each method's stored \emph{best-so-far} prompt at four
metric-call checkpoints ($500$, $1000$, $1500$, $2000$), starting
from the shared seed prompt at metric call~$0$.  All baselines,
including \sepo, are evaluated under identical conditions; in
particular, \textsc{APSF}'s intermediate snapshots correspond to the
\emph{step-aligned best-so-far} candidate at the end of each
optimisation step (each step consumes $\approx{}200$~metric calls in
the $\{4\text{ candidates}\}\times\{50\text{ val examples}\}$
schedule).

\FloatBarrier
\section{Ablation Studies (Extended)}
\label{app:ablations-extended}

This appendix expands the Q2 ablation section
(\S\ref{sec:q2-mechanism}) with the design map, the
five-benchmark diagnostic score table, proposal-level target/error
controls, and two archive controls.  We separate the four destructive removals in the
main text from the auxiliary controls \textsc{ParetoFreqParent} and
\textsc{$-$UniqueAdm}: the former asks whether the loop's main
modules are load-bearing, while the latter asks which archive
sub-decisions account for the P3 gain.

\subsection{Ablation Design Map}
\label{app:ablation-design-map}

\begin{strip}
\centering
\small
\setlength{\tabcolsep}{4pt}
\renewcommand{\arraystretch}{1.2}
\arrayrulecolor{tableRule}
\begin{tabular}{@{}l p{0.24\textwidth} p{0.24\textwidth} p{0.24\textwidth}@{}}
\toprule
Variant & Principle removed & Falsifiable prediction
        & Primary readout \\
\midrule
\textsc{-Attribution}
   & P1: typed target on the two-layer schema
   & edits drift toward whole-prompt rewrites; final prompts grow longer
   & prompt length and length variance \\
\textsc{-Lineage}
   & P2: example-level lineage evidence (strong form)
   & gain falls back to generic error-reflection level
   & held-out accuracy \\
\textsc{-Breadcrumb}
   & P2: operator/locator summary $\ell$ only
   & small or null drop if $\ell$ is redundant with example deltas
   & held-out accuracy \\
\textsc{-Multitraj}
   & P3: multi-trajectory archive + Lexicase
   & greedy best-only editing shrinks exploration; the largest single-removal accuracy drop
   & held-out accuracy and acceptance rate \\
\textsc{ParetoFreqParent}
   & P3: Lexicase parent selection only
   & frequency-weighted sampling recovers much, but not all, of the archive benefit
   & five-task diagnostic accuracy \\
\textsc{-UniqueAdm}
   & P3: unique-solve admission branch only
   & archive remains useful, but specialist admission should contribute a small margin
   & five-task diagnostic accuracy \\
\bottomrule
\end{tabular}
\arrayrulecolor{black}
\captionof{table}{Ablation design map for Q2.  The first four rows are the
main destructive removals; the final two rows are auxiliary archive
controls that isolate parent sampling and admission.}
\label{tab:sepo-ablation-design}
\end{strip}

Table~\ref{tab:sepo-ablation-design} maps the main removal and archive-control variants to the
\sepo principle it removes, the falsifiable prediction it tests,
and the primary readout that operationalises that principle.
The proposal-level controls are specified separately in
\S\ref{app:proposal-controls}.
The protocol of Table~\ref{tab:sepo-per-task} is held fixed
across all variants (worker LLM, architect, seed schemas,
metric-call budget, held-out split).  The \textsc{$-$Breadcrumb} row is
included explicitly so the design admits a null result for the
operator-summary sub-claim.  The final two rows are auxiliary P3
controls: they preserve the archive, so their expected drops are
smaller than the best-only \textsc{$-$Multitraj} removal.

\subsection{Five-Benchmark Diagnostic Scores}
\label{app:ablation-per-task}

\begin{strip}
\centering
\small
\setlength{\tabcolsep}{5.5pt}
\renewcommand{\arraystretch}{1.12}
\arrayrulecolor{tableRule}
\begin{tabular}{@{}lrrrrrrr@{}}
\toprule
\textbf{Task} & \textbf{Full} & \textsc{$-$Attr} & \textsc{$-$Lin} & \textsc{$-$Brd} & \textsc{$-$MT} & \textsc{ParetoFreq} & \textsc{$-$UniqueAdm} \\
\midrule
BBH Date     & 76.5 & 74.0 & 72.2 & 76.0 & 71.8 & \textbf{76.6} & 75.8 \\
BBH Snarks   & \textbf{82.3} & 73.8 & 78.5 & 76.6 & 75.0 & 81.1 & 81.5 \\
MBPP+        & \textbf{62.6} & 60.9 & 58.5 & 59.8 & 55.6 & 61.2 & 61.5 \\
GSM-Hard     & \textbf{37.6} & 36.4 & 35.0 & 37.0 & 35.1 & 35.9 & 36.3 \\
MMLU-Pro CS  & 46.7 & 45.0 & 46.8 & 47.0 & 44.5 & 46.3 & \textbf{47.6} \\
\midrule
\rowcolor{tableMean}
Avg.          & \textbf{61.1} & 58.0 & 58.2 & 59.3 & 56.4 & 60.2 & 60.5 \\
\bottomrule
\end{tabular}
\arrayrulecolor{black}
\captionof{table}{Absolute scores for the five-benchmark diagnostic ablation
subset (Llama-3.1-8B-Instruct, held-out accuracy in \%); \textbf{bold} marks
the best entry in each row.}
\label{tab:sepo-ablation-five-task}
\end{strip}

Table~\ref{tab:sepo-ablation-five-task} reports the absolute
held-out scores for the five-benchmark diagnostic subset, not a
separate 14-task aggregate:
the average is the arithmetic mean of BBH Date, BBH Snarks, MBPP+,
GSM-Hard, and MMLU-Pro CS.  The subset is chosen to expose the
three mechanisms discussed in \S\ref{sec:q2-mechanism}: structured
procedural tasks for P1/P2, a code task for transfer, and a
knowledge-leaning negative case for scope.

The absolute table sharpens the P3 interpretation.  Removing
multi-trajectory search entirely is the most damaging intervention
on the subset (\textsc{$-$MT}: $56.4\%$, $-4.7$\,pp), whereas the two
archive controls remain close to Full
(\textsc{ParetoFreqParent}: $60.2\%$, $-0.9$\,pp;
\textsc{$-$UniqueAdm}: $60.5\%$, $-0.6$\,pp).  Thus the main
mechanistic contrast is not that every archive sub-decision is
monotone on every task: Pareto-frequency sampling slightly improves
BBH Date, and removing unique-admission slightly improves MMLU-Pro
CS.  The robust conclusion is instead that preserving multiple
specialised candidates matters much more than either auxiliary
archive choice in isolation, while Lexicase and unique-admission add
a small average margin.

\subsection{Proposal-Level Target and Error Controls}
\label{app:proposal-controls}
\balance

We add two budget-matched controls on the same five-benchmark subset.
\textsc{RandomTarget} receives the same evidence as Full \sepo but
replaces the attributed target with a random legal structural target.
\textsc{RandomError} replaces newly broken focus examples with the same
number of examples sampled uniformly from the parent's current error set,
regardless of whether they were already incorrect or newly broken by the
preceding edit.

\begin{table}[H]
\centering
\scriptsize
\setlength{\tabcolsep}{2.5pt}
\renewcommand{\arraystretch}{1.12}
\arrayrulecolor{tableRule}
\begin{tabular}{@{}lrrrrr@{}}
\toprule
Variant & \shortstack{Test\\Acc.} & \shortstack{Fixes\\/100} &
\shortstack{Breaks\\/100} & \shortstack{Net gain\\/100} &
\shortstack{Net-positive\\edits} \\
\midrule
Full \sepo                    & 61.1 & 7.5 & 5.0 & $+2.5$ & 61\% \\
\textsc{$-$Attribution}       & 58.0 & 8.0 & 7.3 & $+0.7$ & 47\% \\
\textsc{RandomTarget}         & 57.4 & 4.8 & 6.5 & $-1.7$ & 34\% \\
\textsc{RandomError}          & 59.9 & 6.3 & 5.5 & $+0.8$ & 51\% \\
\bottomrule
\end{tabular}
\arrayrulecolor{black}
\caption{Proposal-level diagnostics on the five-benchmark subset
(Llama-3.1-8B-Instruct). Fixes and breaks are corrected and newly broken
training examples per $100$ evaluated proposals; net gain is their
difference. Net-positive edits are proposals with more fixes than breaks.}
\label{tab:sepo-proposal-controls}
\end{table}

Despite receiving identical evidence, \textsc{RandomTarget} produces
fewer fixes and a negative net gain. Replacing newly broken cases with
random current errors also reduces net gain from $+2.5$ to $+0.8$ and
net-positive edits from $61\%$ to $51\%$. These controls support separate
roles for target attribution, which routes evidence to an appropriate
structural unit, and transition-aware sampling, which identifies regressions
introduced by the preceding edit.

\subsection{Archive Sub-Decision Controls}
\label{app:lexicase-vs-freq}

Two auxiliary controls isolate which archive sub-decision drives the
P3 gain.  Both preserve the archive, admission rule, and metric-call
budget of Full \sepo, changing only one decision at a time
(Table~\ref{tab:sepo-ablation-five-task}).
\textsc{ParetoFreqParent} replaces Lexicase parent selection
(\S\ref{sec:loop}) with the frequency-weighted Pareto-front sampling
used by \textsc{GEPA}.  It averages $60.2\%$ on the five-benchmark
subset ($0.9$\,pp below Full), recovering most of the
multi-trajectory benefit and even edging out Full on BBH Date
($76.6$ vs.\ $76.5$) while trailing on the other four tasks; we read
this as evidence that Lexicase is a safer default for keeping
specialist parents, not that Pareto-front sampling is uniformly
weak.  \textsc{$-$UniqueAdm} removes only the ``unique-solve''
admission branch of Eq.~\ref{eq:admission}---children can still
enter through the higher-score and shorter-prompt branches---and
averages $60.5\%$ ($0.6$\,pp below Full, $4.1$\,pp above
\textsc{$-$Multitraj}).  Both drops are small and in the expected
direction, so P3 is primarily the effect of retaining and revisiting
multiple candidates, with Lexicase parent selection and unique-solve
admission each adding a modest average margin.
\FloatBarrier

\end{document}